\pdfoutput=1

\documentclass[11pt]{article}

\usepackage[final]{acl}

\usepackage{times}
\usepackage{latexsym}

\usepackage[T1]{fontenc}

\usepackage[utf8]{inputenc}

\usepackage{microtype}

\usepackage{inconsolata}

\usepackage{graphicx}

\usepackage{forest} 
\usepackage{tikz}
\usetikzlibrary{arrows.meta, shapes, positioning, shadows, trees}
\usepackage{pifont} 
\usepackage{booktabs} 
\usepackage{enumitem} 
\usepackage{url} 
\usepackage{hyperref} 
\usepackage{multirow} 
\usepackage{subcaption} 

\usepackage{tcolorbox}

\newtcolorbox{querybox}{
  colback=gray!10,
  colframe=black,
  boxrule=0.8pt,
  arc=2mm,
  auto outer arc,
  left=2mm, right=2mm, top=2mm, bottom=2mm,
  after skip=6pt,
  before skip=6pt,
  fontupper=\small\ttfamily\linespread{1.2}\selectfont
}

\usepackage{xurl}

\usepackage{makecell}   
\usepackage{colortbl}   
\usepackage[table]{xcolor} 

\newcommand{\clickablecheckmark}[1]{\href{#1}{\ding{51}}}

\title{\raisebox{-4pt}{\includegraphics[width=24pt]{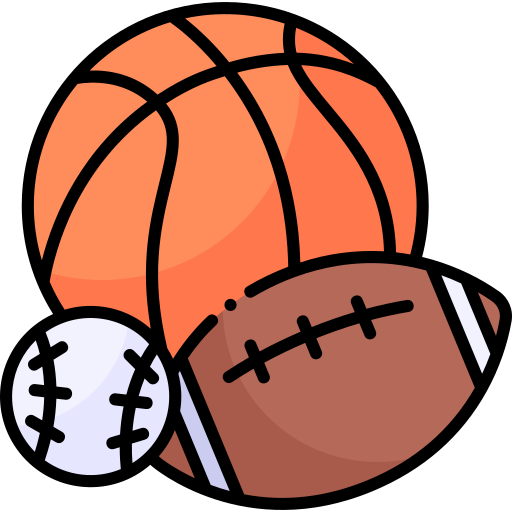}}~~A Survey of Large Models in Sports}

\author{
Yichen Xu$^{1}$\thanks{These authors contributed equally.} ~~
Jianzhe Ma$^{1}$\footnotemark[1] ~~
Chuhan Wang$^{2}$ ~~
Zhonghao Cao$^{3}$\\
\textbf{Liangyu Chen}$^{1}$ ~~
\textbf{Wenxuan Wang}$^{1}$\thanks{Qin Jin and Wenxuan Wang are corresponding authors.} ~~
\textbf{Qin Jin}$^{1}$\footnotemark[2] \\
$^1$Renmin University of China \quad
$^2$Sichuan University \\
$^3$Beijing University of Posts and Telecommunications \\
\texttt{\{xu\_yichen, majianzhe, liangyuchen, wangwenxuan, qjin\}@ruc.edu.cn} \\
\texttt{wangchuhan51@stu.scu.edu.cn} \quad
\texttt{caozhonghao@bupt.edu.cn}
}

\begin{document}
\maketitle
\begin{abstract}
Sports have witnessed growing global enthusiasm in recent years, serving as a vital force for physical health, cultural exchange, social connection, and economic growth. The rapid advancement of large models, particularly (multimodal) large language models (M)LLMs, has demonstrated transformative potential to reshape sports understanding, analysis, and interaction across diverse domains. This paper presents a comprehensive survey of large models in sports, including (i) an overview of tasks and applications across different participant groups; (ii) a detailed analysis of sports-related datasets and benchmarks; and (iii) a critical discussion of current challenges and future directions. Our goal is to establish a foundation for advancing research and practical development of large-model-driven sports intelligence. 
An open-source GitHub repository is maintained at:
\url{https://github.com/Road2Redemption/Awesome_Large_Models_In_Sports1}.
\end{abstract}

\section{Introduction}
\label{sec:intro}

In recent years, the global enthusiasm for sports has continued to rise, with more and more people actively participating in it, and the sports industry has also flourished. To further drive this development, modern sports increasingly rely on massive data support~\citep{hutchins2016tales}, while the introduction of Artificial Intelligence (AI) has greatly accelerated this trend~\citep{zhou2025artificial}.
A pivotal pillar of this transformation is the ability to process and generate sports-related language, which serves as a vital bridge translating raw athletic data into actionable insights for participants and fans alike.

\begin{figure*}[htbp]
    \centering 
  \includegraphics[width=1\textwidth]{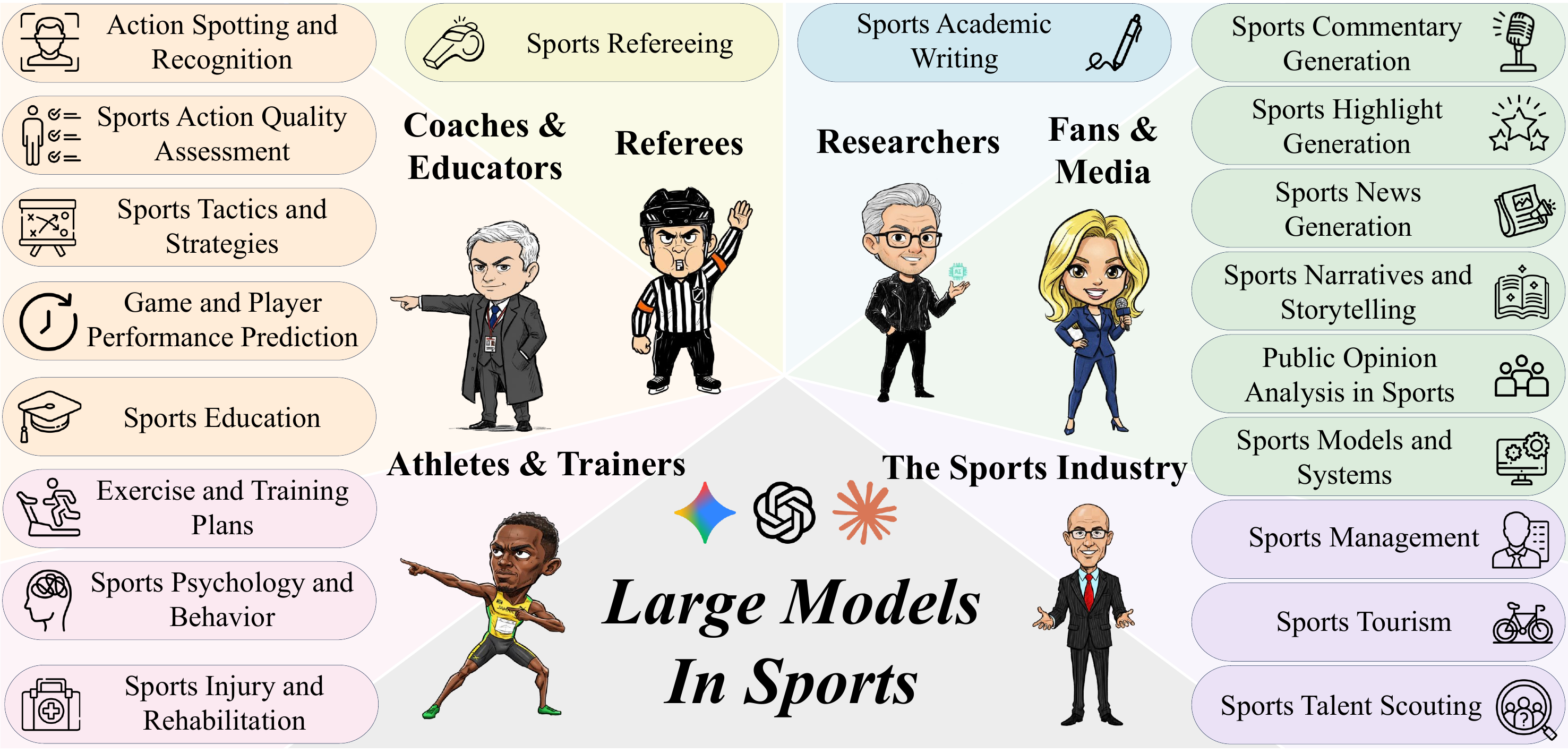}
  \caption{
    Large models have shown powerful applications across 6 sports stakeholder groups: athletes and trainers, coaches and educators, referees, researchers, fans and social media, and the sports industry, enabling diverse tasks.
  }
  \vspace{-0.5cm}
  \label{fig:fig1}
\end{figure*}

Early interdisciplinary research in sports and AI focused on natural language processing and computer vision, with applications in tasks such as sports data processing~\citep{cossich2023technological} and video analysis~\citep{naik2022comprehensive}.
As shown in Figure~\ref{fig:fig1}, the transition to the era of large models—underpinned by the rapid evolution of Large Language Models (LLMs) and Multimodal Large Language Models (MLLMs) like GPT-4~\citep{achiam2023gpt4} and Gemini~\citep{team2023gemini}---has brought new opportunities and challenges to the sports domain.
With linguistic intelligence at their core, these models not only generate language effectively but also process multiple data modalities, enabling broader applications in sports.
Tasks that were previously difficult---such as designing athlete training plans~\citep{skerik2018automated}, developing coaching strategies~\citep{bunker2022application}, and generating sports game summarization~\citep{huang2020generating}---have been greatly enhanced by large models. Moreover, leveraging their vast knowledge bases, these models can generate more comprehensive and personalized content~\citep{app4fans_com_lin2024personalized}.
The number of papers on large models in sports has grown rapidly, from just 1 in
2020 to 78 in 2024, and continues to increase in 2025 (see Figure~\ref{fig:year} in the Appendix).

A growing body of review literature has examined the use of AI and deep learning in sports~\citep{zhou2025artificial, zhao2025survey}. The most relevant work on large models includes studies on their applications to exercise recommendations
~\citep{lai2025using}, sports science and medicine~\citep{connor2023large, naughton2024challenges}, and the sports industry~\citep{wang2024impact}, along with surveys of datasets for language and multimodal models~\citep{xia2024language}. However, these studies are still limited in scope, lacking comprehensive coverage of the diverse sports-related tasks and datasets where large models can be applied.

To ensure a comprehensive and rigorous survey, we adopted a systematic snowballing methodology~\citep{snowballing}, adhering to the PRISMA statement~\citep{page2021prisma}. 
Starting from the aforementioned review papers, we performed iterative forward and backward searches to capture the latest advancements in the era of large models (Jan 2020--July 2025). 
This process resulted in a final collection of \textbf{241} core academic papers addressing large models in sports. 
The detailed selection methodology is provided in the Appendix~\ref{app_sec:appendix_prisma}.

We first systematically categorize and summarize existing applications of large models in sports across 6 key groups (§\ref{sec:app}). 
Then, we review and conduct an in-depth analysis of the relevant datasets and benchmarks in sports (§\ref{sec:data}).
Subsequently, we discuss the current challenges in this field and, finally, outline the potential future directions (§\ref{dis}).


\definecolor{myA}{RGB}{253, 244, 248} 
\definecolor{myB}{RGB}{255, 248, 240} 
\definecolor{myC}{RGB}{250, 250, 232} 
\definecolor{myD}{RGB}{240, 247, 240} 
\definecolor{myE}{RGB}{243, 249, 251} 
\definecolor{myF}{RGB}{248, 245, 252} 

\tikzset{
    basic/.style  = {draw,rectangle, font=\scriptsize},
    root/.style   = {basic, rounded corners=3pt, thin, 
    fill=white, text width=1.5cm,font=\scriptsize\bfseries},
    onode/.style = {basic, thin, rounded corners=3pt, align=left, fill=white, text width=1.7cm},
    onodeA/.style = {basic, thin, rounded corners=3pt, align=left, fill=myA, text width=1.7cm},
    onodeB/.style = {basic, thin, rounded corners=3pt, align=left, fill=myB, text width=1.7cm},
    onodeC/.style = {basic, thin, rounded corners=3pt, align=left, fill=myC, text width=1.7cm},
    onodeD/.style = {basic, thin, rounded corners=3pt, align=left, fill=myD, text width=1.7cm},
    onodeE/.style = {basic, thin, rounded corners=3pt, align=left, fill=myE, text width=1.7cm},
    onodeF/.style = {basic, thin, rounded corners=3pt, align=left, fill=myF, text width=1.7cm},
    tnode/.style = {basic, thin, rounded corners=3pt, align=left, fill=white, text width=3.5cm},
    tnodeA/.style = {basic, thin, rounded corners=3pt, align=left, fill=myA, text width=3.5cm},
    tnodeB/.style = {basic, thin, rounded corners=3pt, align=left, fill=myB, text width=3.5cm},
    tnodeC/.style = {basic, thin, rounded corners=3pt, align=left, fill=myC, text width=3.5cm},
    tnodeD/.style = {basic, thin, rounded corners=3pt, align=left, fill=myD, text width=3.5cm},
    tnodeE/.style = {basic, thin, rounded corners=3pt, align=left, fill=myE, text width=3.5cm},
    tnodeF/.style = {basic, thin, rounded corners=3pt, align=left, fill=myF, text width=3.5cm},
    xnode/.style = {basic, thin, rounded corners=3pt, align=left, fill=blue!5, text width=6.9cm},
    xnodeA/.style = {basic, thin, rounded corners=3pt, align=left, fill=myA, text width=6.9cm},
    xnodeB/.style = {basic, thin, rounded corners=3pt, align=left, fill=myB, text width=6.9cm},
    xnodeC/.style = {basic, thin, rounded corners=3pt, align=left, fill=myC, text width=6.9cm},
    xnodeD/.style = {basic, thin, rounded corners=3pt, align=left, fill=myD, text width=6.9cm},
    xnodeE/.style = {basic, thin, rounded corners=3pt, align=left, fill=myE, text width=6.9cm},
    xnodeF/.style = {basic, thin, rounded corners=3pt, align=left, fill=myF, text width=6.9cm},
}
\begin{figure*}[t] 
    \centering 
    \begin{forest} for tree={
      grow=east,
      anchor=west,
      s sep+=-5pt,
      edge={-},
      edge path={
        \noexpand\path [draw, ->, \forestoption{edge}] (!u.parent anchor) -- +(5pt,0) |- (.child anchor)\forestoption{edge label};
      },
      parent anchor=east,
      child anchor=west,
    }
    [Applications and Tasks(§\ref{sec:app}), root
        [The Sports \\Industry~(§\ref{subsec:ind}), onodeF
            [Sports Tourism, tnodeF
                [{E.g., AI-Powered ChatGPT~\citeyearpar{app4ind_tou_memon2025ai}, Esports Tourism~\citeyearpar{app4ind_tou_yenisoy2025investigating}}, xnodeF
                ]
            ]
            [Sports Talent Scouting, tnodeF
                [{E.g., LLM+RAG\citeyearpar{app4ind_tal_martire2025leveraging}, Footyintel\citeyearpar{app4ind_tal_raskar2025footyintel}, TwelveGPT Scout\citeyearpar{app4ind_tal_mateus2024empowering}} , xnodeF
                ]
            ]
            [Sports Management, tnodeF
                [{E.g., Finance~\citeyearpar{app4ind_man_haghparast2024financial}, Database~\citeyearpar{app4ind_man_merilehto2024pdfs},  Facilities Site~\citeyearpar{app4ind_man_salimi2025comprehensive}} , xnodeF
                ]
            ]
        ]
        [Researchers(§\ref{subsec:res}), onodeE
            [Sports Academic Writing, tnodeE
                [{E.g., Manuscript Generation~\citeyearpar{app4res_anderson2023ai}, Sample-Size Calculation\citeyearpar{app4res_methnani2023chatgpt}} , xnodeE
                ]
            ]
        ]
        [Fans and \\Media~(§\ref{subsec:fan}), onodeD
            [Sports Models and Systems, tnodeD
                [{E.g., Megan~\citeyearpar{app4fans_mod_priya2024megan}, BleacherBot~\citeyearpar{app4fans_mod_kim2025bleacherbot}, FineQuest~\citeyearpar{chen2025finequest}} , xnodeD
                ]
            ]
            [Public Opinion Analysis in Sports, tnodeD
                [{E.g., Aspect-Based Analysis~\citeyearpar{app4fans_opi_qian2025experience}, Few-Shot Pipeline~\citeyearpar{app4fans_opi_rauchegger2024onelove}} , xnodeD
                ]
            ]
            [Sports Narratives and Storytelling, tnodeD
                [{E.g., Sportify~\citeyearpar{app4fans_nar_lee2024sportify}, SoccerSum~\citeyearpar{app4fans_nar_sarkhoosh2024multimodal}, SportsBuddy~\citeyearpar{app4fans_nar_lin2025sportsbuddy}} , xnodeD
                ]
            ]
            [Sports News Generation, tnodeD
                [{E.g., KES~\citeyearpar{app4fans_new_wang2022knowledge}, SNIL~\citeyearpar{app4fans_new_cheng2024snil}, BADGE~\citep{app4fans_new_chiang2024badge}} , xnodeD
                ]
            ]
            [Sports Highlight Generation, tnodeD
                [{E.g., DIAMOND~\citeyearpar{app4fans_hig_kang2025diamond}, SportSummarizer~\citep{app4fans_hig_davids2025sportsummarizer}} , xnodeD
                ]
            ]
            [Sports Commentary Generation, tnodeD
                [{E.g., MatchVoice~\citeyearpar{app4fans_com_rao2024matchtime}, AiCommentator~\citeyearpar{app4fans_com_andrews2024aicommentator}, LiveCC~\citeyearpar{app4fans_com_chen2025livecc}} , xnodeD
                ]
            ]
        ]
        [Referees(§\ref{subsec:ref}), onodeC
            [Sports Refereeing, tnodeC
                [{E.g., X-VARS~\citep{app4ref_ref_app4held2024x, app4ref_ref_held2025enhancing}} , xnodeC
                ]
            ]
        ]
        [Coaches and \\Educators~(§\ref{subsec:coa}), onodeB
            [Sports Education, tnodeB
                [{E.g., Sports Management~\citeyearpar{app4coa_edu_keiper2023artificial}, Sports Science~\citeyearpar{app4coa_edu_fazackerley2025harnessing}, PE~\citeyearpar{app4coa_edu_gencc2023artificial}} , xnodeB
                ]
            ]
            [Game and Player Performance Prediction, tnodeB
                [{E.g., RallyTemPose~\citep{app4coa_pre_ibh2024stroke}, AI4Handball~\citep{app4coa_pre_felice2024ai}} , xnodeB
                ]
            ]
            [Sports Tactics and Strategies, tnodeB
                [{E.g., TacticalGPT~\citeyearpar{app4coa_tac_caron2023tacticalgpt}, Smartboard~\citeyearpar{app4coa_tac_liu2024smartboard}, TacticExpert~\citeyearpar{app4coa_tac_lingrui2025tacticexpert}} , xnodeB
                ]
            ]
            [Sports Action Quality Assessment, tnodeB
                [{E.g., FitnessAgent~\citeyearpar{app4coa_aqa_tang2025fitnessagent}, LLM-FMS~\citep{app4coa_aqa_xing2025llm}} , xnodeB
                ]
            ]
            [Action Spotting and Recognition, tnodeB
                [{E.g., SV3.3B~\citeyearpar{app4coa_ana_kodathala2025sv3}, F-16~\citeyearpar{app4coa_ana_li2025improving}, ActionAtlas\citeyearpar{app4coa_ana_salehi2024actionatlas}, F³Set\citeyearpar{app4coa_ana_liu2025f} } , xnodeB
                ]
            ]
        ]
        [Athletes and \\Trainers~(§\ref{subsec:ath}), onodeA
            [Sports Psychology and Behavior, tnodeA
                [{E.g., cMABxLLM~\citeyearpar{app4ath_psy_song2025investigating}, Multisport YODA~\citeyearpar{app4ath_psy_zuccolotto11}, PHIA~\citeyearpar{app4ath_psy_merrill2024transforming}}
                    , xnodeA
                ]
            ]
            [Sports Injury and Rehabilitation, tnodeA
                [{E.g., PanelGPT~\citeyearpar{app4ath_inj_mcbee2023interdisciplinary, app4ath_inj_mcbee2024assessing}, GaLore~\citep{app4ath_inj_zhu2025full}}
                    , xnodeA
                ]
            ]
            [Exercise and Training Plans, tnodeA
                [{E.g., GPTCoach~\citeyearpar{app4ath_pla_jorke2025gptcoach}, ExpertAF~\citeyearpar{app4ath_pla_ashutosh2025expertaf}, MAAIG~\citeyearpar{app4ath_pla_yeh2023maaig}}
                    , xnodeA
                ]
            ]
        ]
    ]
    \end{forest}
    \caption{Taxonomy of applications, tasks, and approaches of large models in sports.}
    \vspace{-0.5cm}
    \label{app_and_task_figure}
\end{figure*}
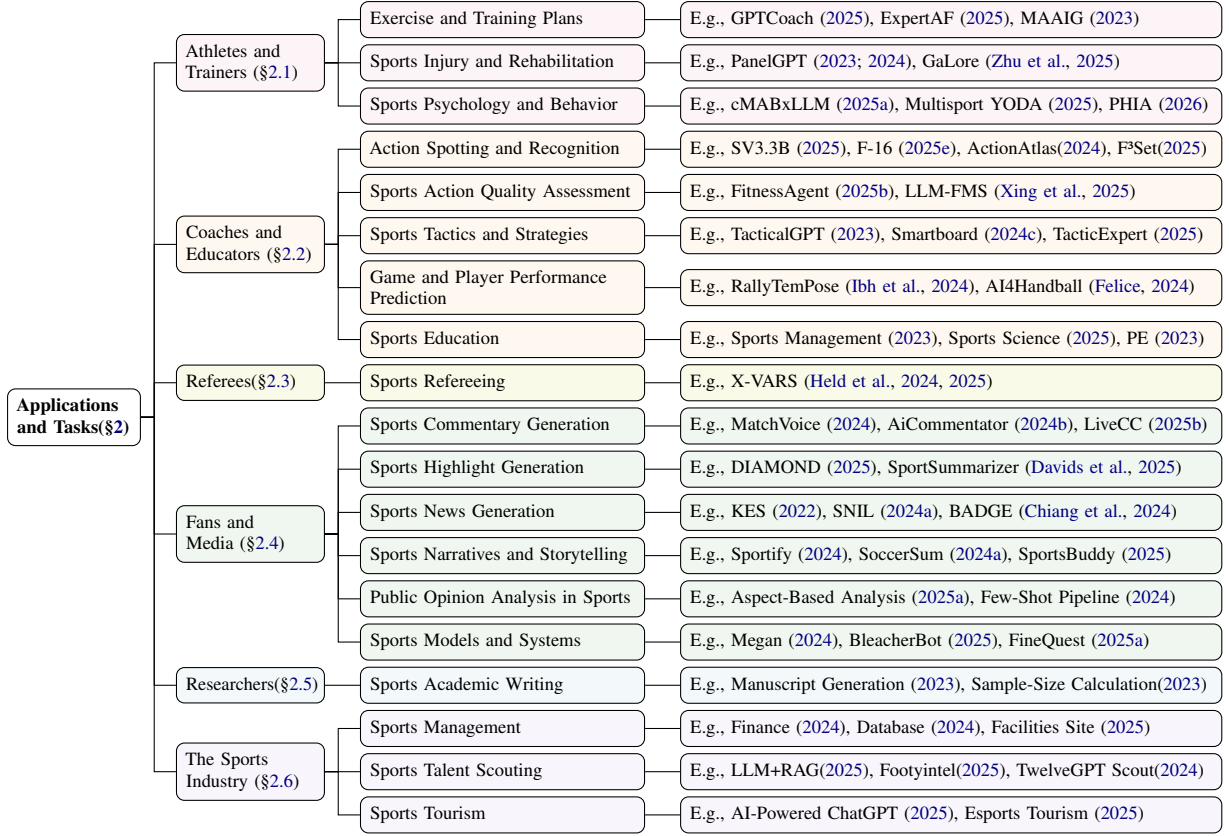

\section{Large Model Applications in Sports}
\label{sec:app}

The fast growth of large models has brought big chances for their use in sports.
As shown in Figure~\ref{app_and_task_figure}, we categorize these applications into a taxonomy with \textbf{6 stakeholder groups} and \textbf{19 specific tasks}. 
In this section, beyond merely listing existing literature, we conduct \textbf{a detailed analysis} of the impact of large models on each task, focusing on defining the task, analyzing technical paradigms, and summarizing common evaluation metrics.
For comprehensive reviews of specific works associated with each task, see Appendix~\ref{app_sec:appendix_app}.

\subsection{Applications for Athletes and Trainers}
\label{subsec:ath}

\noindent\textbf{Exercise and Training Plans.}
Large models help athletes and trainers create exercise prescriptions, translating sports science into practice to improve performance~\citep{app4ath_pla_phillips2012exercise, app4ath_pla_wackerhage2021personalized}. Recent AI coaches powered by LLMs significantly facilitate the generation of personalized training plans across a wide spectrum of health conditions and fitness goals, ranging from general weight management~\citep{app4ath_pla_saracc2025evaluating} to chronic disease guidance~\citep{app4ath_pla_onan2025examining}.
As shown in Table~\ref{tab:task_performance_comparison}, in the YourSkatingCoach dataset~\citep{chen2024yourskatingcoach}, a fine-tuned T5 model~\citep{raffel2020T5} achieves a BLEU-4 score of \textit{0.27}, while a vanilla Transformer~\citep{vaswani2017attention_transformer} trained from scratch only reaches 0.04~\citep{app4ath_pla_yeh2023maaig}. This highlights that LLMs leverage pre-trained knowledge to address sports data scarcity and excel at open-ended generation, outperforming traditional rule-based or smaller deep learning models. Additionally, strategies like Retrieval-Augmented Generation (RAG)~\citep{app4ath_pla_zhang2025rag} and agentic paradigms~\citep{app4ath_pla_vahdati2025multi} have been explored to enhance reliability and personalization. Common evaluation metrics include BLEU-4, METEOR, and ROUGE-L.



\noindent\textbf{Sports Injury and Rehabilitation.}
Large models assist athletes and trainers throughout the entire lifecycle of sports injury management, spanning prevention, diagnosis, and rehabilitation, with applications expanding from providing preventive advice~\citep{app4ath_inj_zhu2025full} to aiding clinical decision-making for surgical treatments~\citep{app4ath_inj_saglam2025comparative}.
While LLMs possess the interdisciplinary knowledge required for orthopedics and rehabilitation~\citep{app4ath_inj_mcbee2024assessing}, most current applications rely on the direct deployment of pre-trained large models for Question Answering (QA) and classification. Deep technical integration remains limited, with only early exploration of efficient fine-tuning methods like GaLore~\citep{zhao2024galore} to tailor models for sports medicine~\cite{app4ath_inj_zhu2025full}. This indicates the field is in its infancy, lacking unified evaluation metrics.



\noindent\textbf{Sports Psychology and Behavior.}
Sports psychology enhances athletes' training performance and mental well-being through behavioral interventions. Recent LLM applications range from general cognitive assessment~\citep{app4ath_psy_zuccolotto11} to targeted interventions for specific behavioral issues~\citep{app4ath_psy_masur2025assessment}. Recent advances move beyond text generation by integrating multimodal physiological data from wearable sensors---such as heart rate and IMU signals---to deliver personalized interventions~\citep{app4ath_psy_imran2024llasa, app4ath_psy_merrill2024transforming}. However, task definitions remain ambiguous, and the area lacks standardized benchmarks, requiring further exploration.

\begin{table*}[t]
\centering
\resizebox{1.0\textwidth}{!}{%
\begin{tabular}{l c c c c c}
\toprule
\textbf{Dataset} & \textbf{Model} & \textbf{Architecture} & \textbf{Paradigm} & \textbf{Metric} & \textbf{Performance} \\
\midrule
\multicolumn{6}{c}{\cellcolor{gray!10}\textit{\textbf{Exercise and Training Plans}}} \\
\midrule
\multirow{2}{*}{\textbf{YourSkatingCoach}~\citeyearpar{chen2024yourskatingcoach}}
 & MAAIG~\citeyearpar{app4ath_pla_yeh2023maaig} & T5 (pretrained)~\citeyearpar{raffel2020T5} & fine-tuning & \multirow{2}{*}{BLEU-4} & \textbf{0.27} \\
 & Transformer~\citeyearpar{vaswani2017attention_transformer} & vanilla Transformer & train from scratch &  & 0.04 \\
\midrule
\multicolumn{6}{c}{\cellcolor{gray!10}\textit{\textbf{Action Spotting and Recognition}}} \\
\midrule
\multirow{3}{*}{\textbf{SoccerNet-v2}~\citeyearpar{deliege2021soccernetV2}} 
 & Soccer-CLIP~\citeyearpar{app4coa_ana_shin2025soccer} & CLIP~\citeyearpar{radford2021clip} & fine-tuning & \multirow{3}{*}{t-AmAP} & \textbf{75.7} \\
 & COMEDIAN~\citeyearpar{denize2024comedian} & spatiotemporal Transformer & train from scratch &  & 73.1 \\
 & Llama 3.1-8B~\citeyearpar{dubey2024llama3} & LLM w/ textual commentary~\citeyearpar{app4coa_ana_chakraborty2025we} & few-shot &  & 60.8 \\
\midrule
\multicolumn{6}{c}{\cellcolor{gray!10}\textit{\textbf{Sports Action Quality Assessment}}} \\
\midrule
\multirow{3}{*}{\textbf{FineFS}~\citeyearpar{ji2023finefs}} 
 & Beats-to-Scores~\citeyearpar{app4coa_aqa_wang2025beats} & Video-Audio (V-A) fusion Transformer & train from scratch & \multirow{3}{*}{Spearman's $\rho$} & \textbf{0.88} \\
 & InternVL2~\citeyearpar{chen2024internvl_2} & InternViT + MLP + InternLM2 & fine-tuning &  & 0.86 \\
 & Qwen2-VL~\citeyearpar{wang2024qwen2vl} & ViT + MLP + Qwen2 & fine-tuning &  & 0.75 \\
\midrule
\multicolumn{6}{c}{\cellcolor{gray!10}\textit{\textbf{Sports Commentary Generation}}} \\
\midrule
\multirow{4}{*}{\textbf{SoccerNet-Caption}~\citeyearpar{app4fans_com_mkhallati2023soccernet}} 
 & MatchVoice~\citeyearpar{app4fans_com_rao2024matchtime} & ViT + Aggregator \& MLP + Llama 3 & fine-tuning & \multirow{4}{*}{CIDEr} & \textbf{38.42} \\
 & SoccerComment~\citeyearpar{app4fans_com_li2025multi} & MLLM + memory unit & fine-tuning &  & 36.58 \\
 & SN-Caption~\citeyearpar{app4fans_com_mkhallati2023soccernet} & encoder-decoder Transformer & train from scratch &  & 23.74 \\
 & Video-LLaMA~\citeyearpar{zhang2023videollama} & V-A encoder + Q-Former + LLaMA & zero-shot &  & 3.44 \\
\bottomrule
\end{tabular}
}

\caption{Quantitative comparison of modeling paradigms across 4 representative sports tasks.}
\label{tab:task_performance_comparison}
\vspace{-0.5cm}
\end{table*}

\subsection{Applications for Coaches and Educators}
\label{subsec:coa}



\noindent\textbf{Action Spotting and Recognition.}
Action spotting and recognition in sports involves temporally localizing and classifying fine-grained player movements or events to provide reliable match facts for downstream analytics~\citep{zhao2025survey}. 
Traditional methods relied on specific deep learning architectures trained from scratch, whereas recent MLLMs leverage pre-training alignment for enhanced semantic understanding. As shown in Table~\ref{tab:task_performance_comparison}, on the SoccerNet-v2 action spotting benchmark~\citep{deliege2021soccernetV2}, fine-tuned Soccer-CLIP achieves a state-of-the-art \textit{75.7\%} t-AmAP~\citep{app4coa_ana_shin2025soccer}, slightly surpassing specialized Transformers (73.1\%)~\citep{denize2024comedian} and highlighting the importance of pre-training methods.
In contrast, relying solely on language-centric LLMs through textual commentary prompts yields significantly lower results (60.8\%)~\citep{app4coa_ana_chakraborty2025we}, underscoring the necessity of fine-grained visual alignment rather than merely injecting textualized visual information.
Common metrics include mAP, top-1 accuracy, and F1 score.



\noindent\textbf{Sports Action Quality Assessment.}
Sports Action Quality Assessment (AQA) quantifies the execution of athletic movements for coaching and officiating~\citep{zhou2024comprehensive, zhao2025survey}. 
Methodologies have evolved from simple regression to fine-tuning MLLMs for personalized evaluation~\citep{app4coa_aqa_dibenedetto2025fine} and developing unified agents~\citep{app4coa_aqa_tang2025fitnessagent}.
On the FineFS benchmark~\citep{ji2023finefs} (see Table~\ref{tab:task_performance_comparison}), specialized small-scale Transformers currently outperform general MLLMs (\textit{0.88} versus 0.86 Spearman’s $\rho$) by explicitly aligning audio-visual features~\citep{app4coa_aqa_wang2025beats}. This indicates that high-precision scoring still depends on domain-specific traditional structural designs.
Moreover, architectural choices within MLLMs remain pivotal, as seen in models like InternVL2~\citep{chen2024internvl_2} and Qwen2-VL~\cite{wang2024qwen2vl}; model design and training details can lead to noticeably different performance on this task.
Common metrics include Spearman’s rank correlation, mean square error, and accuracy.



\noindent\textbf{Sports Tactics and Strategies.}
Sports tactics and strategy analysis models on-field interactions to extract actionable strategic patterns using large models~\citep{app4coa_tac_caron2023tacticalgpt}.
Current methodologies employ large model-based frameworks for tactical analysis and visualization (processing structured and unstructured data)~\citep{app4coa_tac_janssens2024large, app4coa_tac_michielssen2024using}, and tactical exploration and design~\citep{app4coa_tac_liu2024smartboard}.
Technically, current research mainly relies on prompt engineering with pre-trained large models, rather than extensive post-training, due to the scarcity of high-quality tactical datasets. This limits the depth of tactical discovery to the capabilities of the frozen base model, indicating that the field is still nascent and requires future exploration to address these data and methodological constraints.


\noindent\textbf{Game and Player Performance Prediction.}
Game and player performance prediction utilizes historical, contextual, and multimodal data to forecast match outcomes and individual behaviors, thereby providing valuable insights for strategic planning and preparation~\citep{xia2024language}. Methodologies have advanced from BERT-based specific action forecasting~\citep{app4coa_pre_ibh2024stroke} to LLM-driven approaches that integrate diverse data sources for more holistic and interpretable predictions~\citep{app4coa_pre_bhatnagar2025analyzing}. Common metrics include accuracy and F1 score.


\noindent\textbf{Sports Education.}
Recent applications of large models in sports education have demonstrated their versatility for educators and teachers. Current research primarily uses general-purpose LLMs to generate and analyze pedagogical data and content~\citep{app4coa_edu_zhang2024using, app4coa_edu_gao2025motion}. However, a gap exists in high-level applications. Professional athlete guidance, in particular, demands deep domain-specific expertise that general models often lack, presenting a promising direction for exploration.

\subsection{Applications for Referees}
\label{subsec:ref}


\noindent\textbf{Sports Refereeing.}
Large models improve sports refereeing by supporting decision-making and enhancing fairness and transparency. Key tasks include QA, captioning, and action recognition. For example, X-VARS~\citep{app4ref_ref_app4held2024x} uses QLoRA~\citep{dettmers2023qlora} fine-tuning on MLLMs to accurately understand video content while following soccer rules, representing the first step toward explainable LLMs for refereeing.

\subsection{Applications for Fans and Social Media}
\label{subsec:fan}

\noindent\textbf{Sports Commentary Generation.}
Sports commentary generation creates natural-language narratives that integrate factual event descriptions, tactical analysis, and emotionally resonant insights, setting it apart from standard video captioning~\citep{app4fans_com_ge2024scbench}. Recent methods have advanced from end-to-end fine-tuning of MLLMs for better temporal alignment and coherence~\citep{app4fans_com_rao2024matchtime, app4fans_com_wang2024commentary} to agentic frameworks that dynamically adapt by prompting LLMs with key events and tracking data~\citep{app4fans_com_andrews2024aicommentator, app4fans_com_vijayakumar2025player}. 
As shown in Table~\ref{tab:task_performance_comparison}, in this semantic-rich task, adapted MLLM architectures like MatchVoice (CIDEr: \textit{38.42})~\citep{app4fans_com_rao2024matchtime} decisively outperform traditional encoder-decoder models (23.74)~\citep{app4fans_com_mkhallati2023soccernet} due to temporal aggregators that handle long-form narratives.
Conversely, the near-failure of zero-shot Video-LLaMA (3.44)~\citep{zhang2023videollama} confirms that practical utility requires domain-specific fine-tuning or RAG~\citep{app4fans_com_li2025multi} to bridge the linguistic gap between raw visual signals and professional terminology.
Key metrics include METEOR, ROUGE-L, and CIDEr.


\noindent\textbf{Sports Highlight Generation.}
Sports highlight generation aims to automatically identify and compile significant match moments into concise summaries for social media. Existing approaches typically employ hybrid frameworks combining computer vision and LLMs to perform sub-tasks like key frame extraction, event localization, video clipping, and captioning~\citep{lee2020highlight, app4fans_hig_midoglu2024ai}. However, the field faces ambiguous definitions and variance in practical applications, requiring future research to clarify task boundaries and standardize protocols.


\noindent\textbf{Sports News Generation.}
Sports news generation automatically produces factual match reports in a standard journalistic style, summarizing key outcomes, events, and statistics. Existing techniques primarily rely on comprehensive frameworks powered by large models. Recent advances include knowledge retrieval~\cite{app4fans_new_wang2022knowledge}, in-context learning~\cite{app4fans_new_cheng2024snil}, and Chain-of-Thought (CoT) prompting~\cite{app4fans_new_chiang2024badge} to enhance quality. Notably, specialized methods like Tree-of-Report address table-to-text challenges, ensuring accurate conversion of structured data into coherent narratives~\citep{app4fans_new_chiang2025tree}. Common metrics include ROUGE-L, F1 score, and LLM-based metrics.


\noindent\textbf{Sports Narratives and Storytelling.}
Sports narratives and storytelling aim to create long-form, multimodal stories that combine match events with contextual details to engage fans. Current work typically inputs keyframe information, commentator narration, and other relevant data into LLMs to generate engaging tactical analyses and personalized narratives for social media~\citep{app4fans_nar_sarkhoosh2024multimodal, app4fans_nar_lin2025sportsbuddy}. While LLMs excel at crafting narratives, they often struggle with the intricacies of specific sports domains. To address this, meticulous prompt engineering is crucial to prevent factual inaccuracies and enhance personalization and engagement. Additionally, the field urgently needs unified evaluation metrics to effectively assess narrative quality.


\noindent\textbf{Public Opinion Analysis in Sports.}
Public opinion and sentiment analysis in sports involves detecting, classifying, and measuring public attitudes toward sporting events or related issues.
However, achieving high accuracy in sports sentiment analysis is challenging due to the complexity of sport-specific contexts. Recent works have employed strategies such as fine-tuning on sports corpora~\citep{app4fans_opi_qian2025experience} and inference-time techniques like in-context learning and CoT prompting~\citep{app4fans_opi_rauchegger2024onelove}. Yet, current research is mostly limited to small-scale analyses. Scaling these approaches to larger datasets is a crucial future direction to establish the generalizability and practical significance of the findings. Common evaluation metrics include accuracy and F1 score.

\noindent\textbf{Sports Models and Systems.} 
Unlike task-specific research, work on sports models and systems targets general-purpose infrastructures. These include: \textit{(1) sports-related chatbots and models} that utilize dialogue state tracking~\cite{app4fans_mod_song2025korean}, domain-specific fine-tuning~\cite{app4fans_mod_rao2025towards}, and multi-agent frameworks to coordinate specialized reasoning~\cite{app4fans_mod_rao2025multi} and knowledge graph integration~\cite{chen2025finequest}; and \textit{(2) search engines and retrieval systems} that employ RAG architectures and offline query understanding to ground LLM outputs in verified sports facts~\citep{app4fans_mod_karat2025system, app4fans_mod_strand2024soccerrag} and facilitate fine-grained video retrieval~\citep{app4fans_mod_gupta2025play}.

\subsection{Applications for Researchers}
\label{subsec:res}



\noindent\textbf{Sports Academic Writing.}
Sports academic writing entails crafting and refining scholarly content in fields like sports science and medicine. The advent of LLMs like ChatGPT has revolutionized this area, shifting the writing process from traditional manual methods to AI-assisted collaboration. Current LLMs excel at generating structured text, such as research outlines and abstract summaries~\citep{app4res_latzel2024artificial}. However, their reliability for scientific accuracy is undermined by model hallucinations, which often compromise factual integrity and calculation precision~\citep{app4res_methnani2023chatgpt, app4res_dergaa2023human}. Thus, while these models can be efficient writing partners, their outputs need thorough human verification and cautious use.

\subsection{Applications for the Sports Industry}
\label{subsec:ind}


\noindent\textbf{Sports Management.}
Large models are gaining traction in sports management, covering areas like financial, database, and facility management. Unlike traditional tools, large models excel at processing both structured and unstructured data, such as PDF reports and long interview transcripts~\citep{app4ind_man_merilehto2024pdfs, app4ind_man_haghparast2025foresight}. However, empirical research in this domain is still limited and requires further exploration.


\noindent\textbf{Sports Talent Scouting.}
Sports talent scouting is vital for clubs to identify, evaluate, and predict player potential, thus building successful teams. Large models, capable of processing vast data, can make this process more objective and data-driven~\citep{app4ind_tal_mateus2024empowering}. Recent research has used RAG to search unstructured data, speeding up football talent scouting~\citep{app4ind_tal_raskar2025footyintel, app4ind_tal_martire2025leveraging}. Yet, there is still much room to define and expand this task to other sports.


\noindent\textbf{Sports Tourism.}
Sports tourism integrates travel services with athletic activities and major events, enriching the experiences of fans and participants. In this realm, large models play key roles, such as analyzing tourism trends and enhancing community engagement~\citep{app4ind_tou_yenisoy2025investigating}. This shift transforms traditional, static travel planning into a dynamic, real-time interactive experience. Nevertheless, current models encounter several challenges, including privacy and data security concerns, as well as managing fan expectations and trust~\citep{app4ind_tou_memon2025ai}. Tackling these issues will be essential for future advancements.

\section{Datasets for Large Models in Sports}
\label{sec:data}

\begin{figure*}[ht]
    \centering
    \begin{subfigure}[t]{0.329\textwidth}
        \centering
        \includegraphics[width=\textwidth]{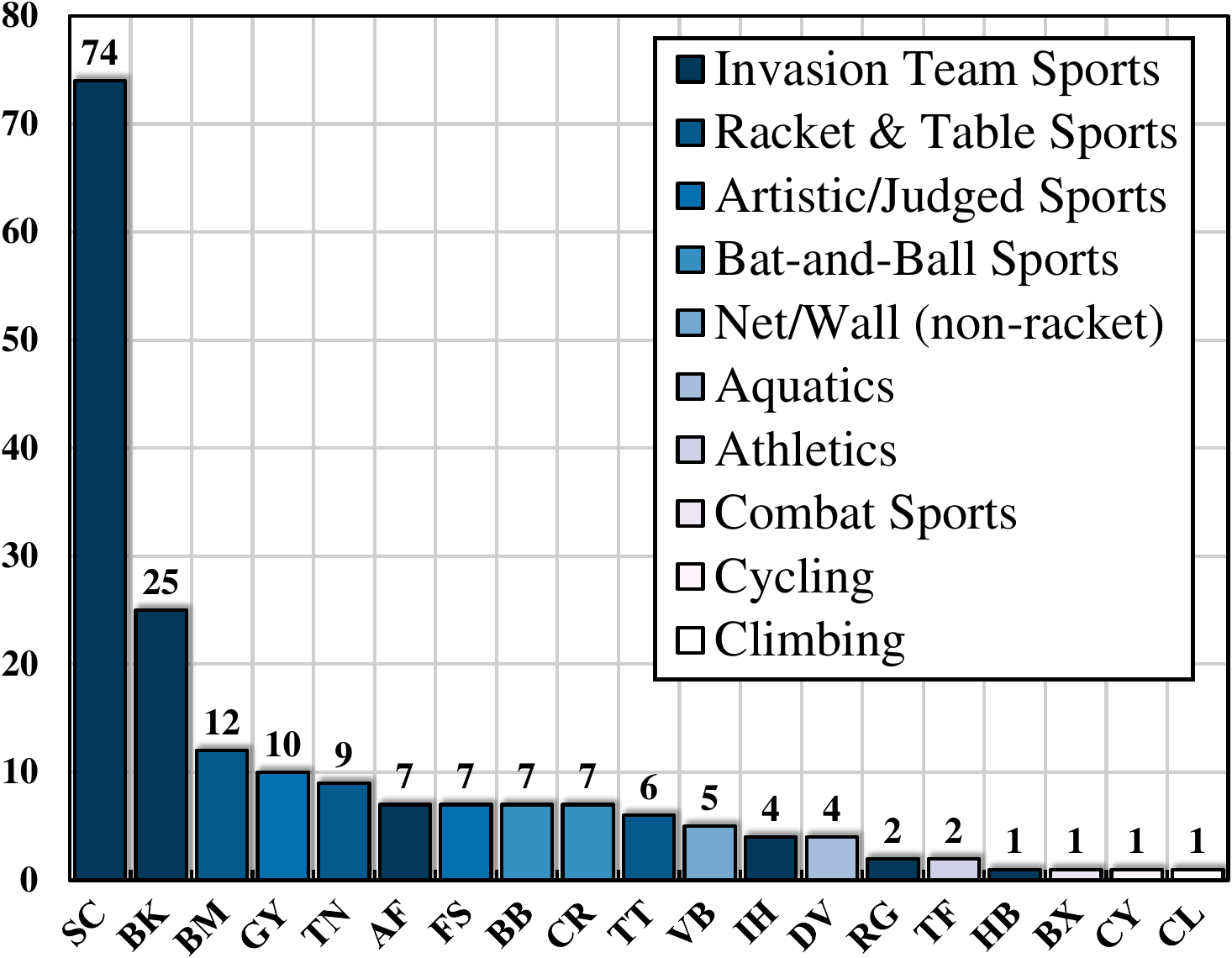}
        \caption{Analysis of sports types}
        \label{fig:fig1_a_ladar_performance}
        \vspace{-0.2cm}
    \end{subfigure}
    \begin{subfigure}[t]{0.329\textwidth}
        \centering
         \includegraphics[width=\textwidth]{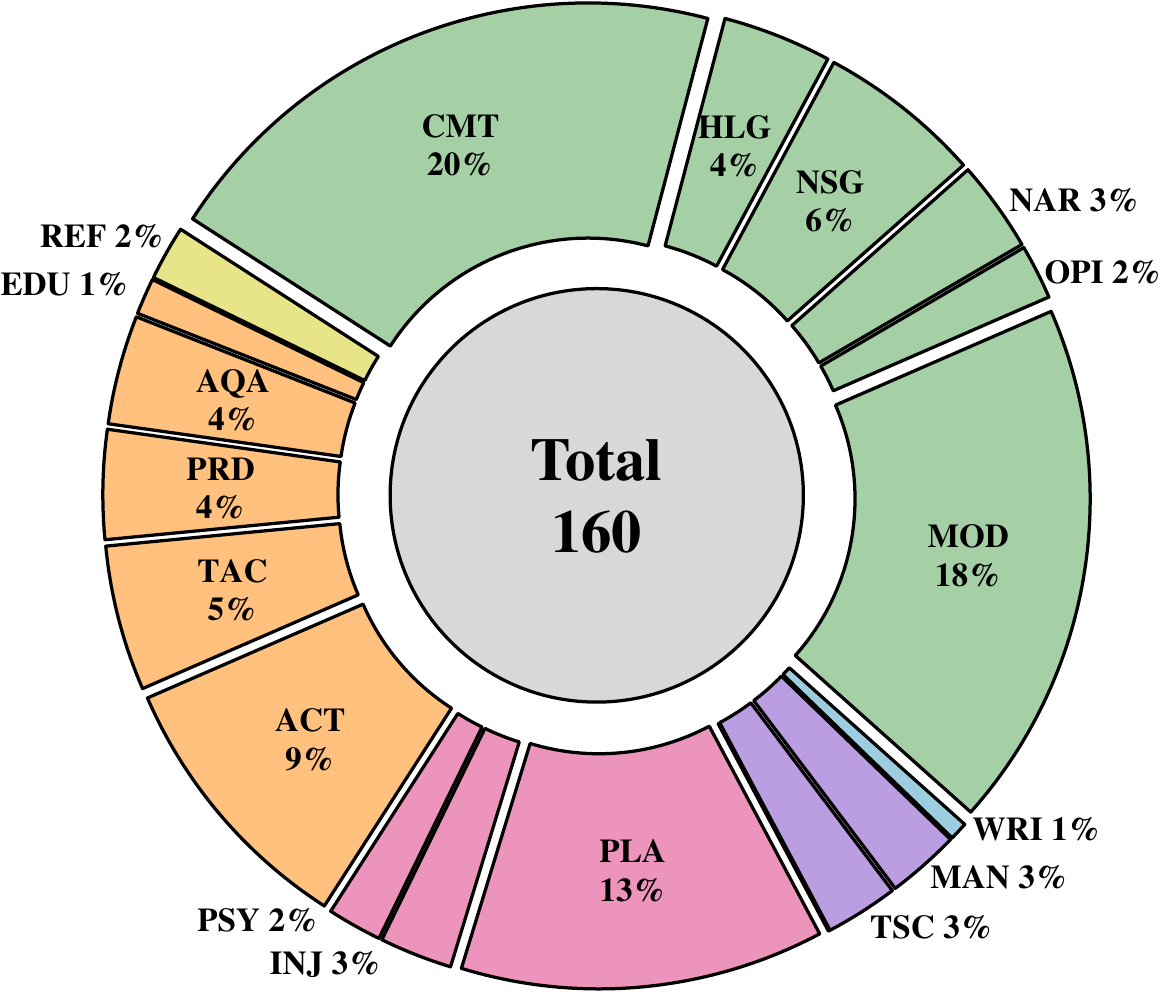}
        \caption{Analysis of application distribution}
        \label{fig:fig1_b_performance2}
        \vspace{-0.2cm}
    \end{subfigure}
    \begin{subfigure}[t]{0.329\textwidth}
        \centering
        \includegraphics[width=\textwidth]{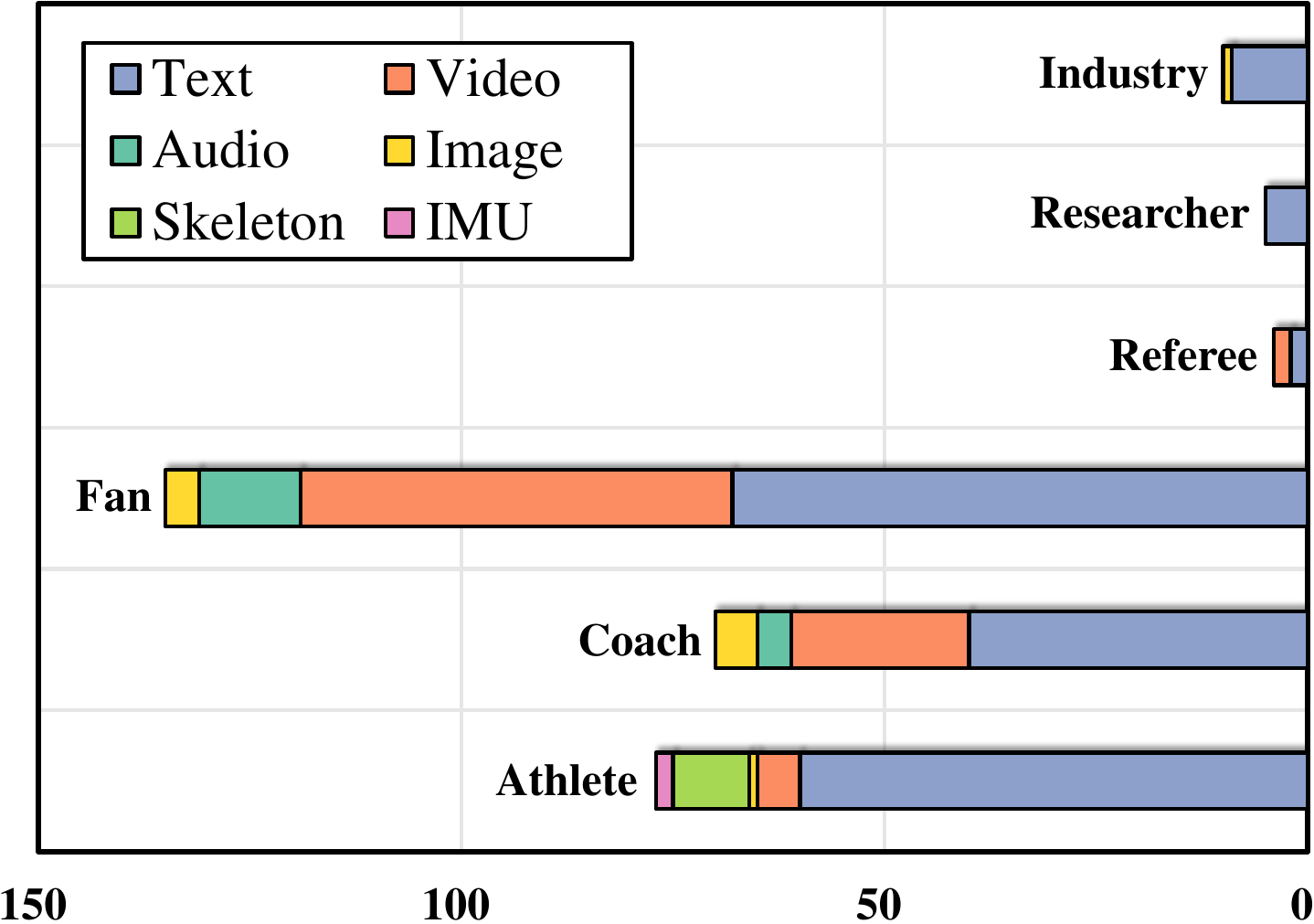}
        \caption{Analysis of modality composition}
        \label{fig:fig1_c_performance3}
        \vspace{-0.2cm}
    \end{subfigure}

    \caption{Data analysis from 3 different perspectives. Sports and tasks abbreviations are listed in Table~\ref{abbr_tab}.}
    \label{fig:fig2_stat}
    \vspace{-0.5cm}
\end{figure*}

In this section, we first categorize the landscape of sports datasets for large models into two main types based on their design objectives: \textbf{task-specific datasets} and \textbf{sports understanding datasets}. 
We then conduct a comprehensive multi-dimensional analysis of their \textbf{dataset distributions} across various facets to identify key trends and research gaps.
More details are provided in the Appendix~\ref{app_sec:appendix_data}.

\setlength{\tabcolsep}{2pt}
\begin{table}[ht]
\centering
\fontsize{8pt}{10pt}\selectfont

\begin{tabular}{lcccc}
\toprule
\textbf{Benchmark}          & \textbf{Sports}          & \textbf{Modal}                & \textbf{\# Video}         & \textbf{\# QA}     \\
\midrule
\textbf{BIG-bench-SU}~\citeyearpar{srivastava2023bigbench}     & SC, BK, etc.                           & text                                & -                                              & 986                               \\
\textbf{SportQA}~\citeyearpar{xia2024sportqa}       & TN, AF, etc.                                            & text                                & -                                             & 70592           \\
\textbf{SPORTU}~\citeyearpar{xiasportu}     & BB, IH, etc.      & video, text                         & 1701                                           & 12948                                 \\
\textbf{Sports-3K-QA}~\citeyearpar{app4fans_com_chen2025livecc} & 49 Sports & video, text & 412 & 1174 \\
\textbf{FSBench}~\citeyearpar{gao2025fsbench}     & FS                & video, text                         &     783                                   & 4000                             \\
\textbf{FBBench}~\citeyearpar{he2025finebadminton}          &     BM                                         & video, text                         &      2563                                  & 2563                                \\
\textbf{Gym-QA}~\citeyearpar{chen2025finequest}           & GY                                           & video, text                         & 6031                                        & 27469                                \\
\textbf{Diving-QA}~\citeyearpar{chen2025finequest}        & DV                                           & video, text                         & $\sim$100                                     & 1055                               \\
\textbf{Sports-QA}~\citeyearpar{li2024sports_qa}        & GY, VB, etc.                               & video, text                         & 5967                                      & 94073                            \\
\bottomrule
\end{tabular}
\caption{Overview of specialized sports understanding benchmarks related to large models. Sports abbreviations are listed in Table~\ref{abbr_tab}.}
\vspace{-0.5cm}
\label{tab:bench_table}
\end{table}

\subsection{Landscape and Categorization}
\label{subsec:data_categorization}

\noindent\textbf{Task-Specific Datasets.}
Task-specific datasets are created to support the practical applications of large models in various sports contexts, covering the 6 stakeholder groups and 19 tasks outlined in Section~\ref{sec:app}. These datasets offer detailed annotations and evaluation metrics, facilitating model training, fine-tuning, and performance assessment. They bridge the gap between the general capabilities of large models and real-world sports applications, enabling customized system development, reproducible evaluation, and advancing research on model deployment. Further details are provided in the Appendix~\ref{app:task_specific}, and Tables~\ref{tab:appendix_task_table1} and~\ref{tab:appendix_task_table2}.

\noindent\textbf{Sports Understanding Datasets.}
Sports are fast-paced, diverse, and strategically complex, presenting unique challenges for large models~\citep{xia2024sportqa}. To enhance models’ comprehension and reasoning in sports contexts, researchers have developed sports understanding datasets. These datasets fall into two main categories:
\textit{(1) datasets specifically for sports understanding}, with Table~\ref{tab:bench_table} providing an overview of relevant benchmarks; and
\textit{(2) general video understanding datasets containing sports content}, covering tasks like video captioning~\citep{wanginternvid}, multi-view understanding~\citep{grauman2024egoexo4d}, and fine-grained analysis~\citep{liu2024ETbench}. More details are in the Appendix~\ref{app:sports_understanding} and Table~\ref{tab:appendix_su_table}.

\subsection{Comprehensive Analysis}
\label{subsec:data_analysis}

\noindent\textbf{Dataset Distribution by Sport Type.} 
As illustrated in Figure~\ref{fig:fig1_a_ladar_performance}, the availability of datasets is notably skewed toward popular invasion team sports (e.g., soccer leading with 74 datasets), racket and table sports, and bat-and-ball sports. Conversely, individual disciplines such as cycling and boxing remain underrepresented with only a single dataset each, highlighting a significant coverage imbalance. Furthermore, the prevalence of fitness-related datasets (21) and the emergence of esports datasets (3) reflect the growing scholarly interest in these evolving domains.

\noindent\textbf{Dataset Distribution by Application.} 
Analysis of the 6 stakeholder groups (Figure~\ref{fig:fig1_b_performance2}) reveals that data representation is robust for athletes, coaches, and fans, aligning with the commercial popularity of these segments. Tasks such as sports commentary generation, exercise prescription, and action recognition have garnered substantial attention. In contrast, data for referees, sports researchers, and the sports industry remain scarce, with referee-relevant datasets accounting for a mere 2\%. This disparity underscores an urgent need to develop specialized datasets to bridge these application gaps.

\noindent\textbf{Dataset Distribution by Modality.} 
In the era of large-scale models, video and text remain the dominant modalities (Figure~\ref{fig:fig1_c_performance3}), while audio-centric studies are beginning to demonstrate their significance~\citep{xie2025maverix}. However, specialized modalities such as IMU sensor data and skeletal poses are relatively rare and primarily confined to athlete-focused motion analysis. Expanding the diversity of these less common modalities is essential to strengthening the cross-modal reasoning capabilities of MLLMs in complex sports scenarios.

\noindent\textbf{Dataset Distribution by Annotation Source.} 
Based on Table~\ref{tab:appendix_su_table}, we categorized the datasets by their annotation origin (automatic, manual, and expert). While manual annotation remains the standard to minimize noise, a critical deficit exists in expert-level labels. This is particularly evident in general datasets, where expert annotation accounts for only 6\%, compared to 41\% in specialized sports understanding benchmarks. This lack of high-quality, professional-grade labels poses a key bottleneck for the fine-tuning and reliability of models intended for elite-level sports analysis.


\noindent\textbf{Dataset Distribution by Modeling Paradigm.}
Sports datasets can be categorized based on their supported tasks, falling into two paradigms: discriminative and generative. \textit{Discriminative-oriented datasets} (21.5\%), such as action recognition and game prediction, benefit from large models' ability to capture spatio-temporal contexts---a notable improvement over traditional methods' handling of complex multimodal inputs. Conversely, \textit{generative tasks} dominate the landscape (78.5\%), spanning from descriptive applications like commentary generation to reasoning-intensive tactical synthesis. This transition underscores a paradigm shift in sports AI: moving beyond simple categorical labeling toward open-ended synthesis and logical reasoning facilitated by large models.

\section{Discussion}
\label{dis}
In the preceding sections, we have examined the landscape of large models in sports and their emerging capabilities across a range of stakeholders. Building upon these insights, this discussion distills the key barriers to practical deployment and outlines promising directions for future research.

\subsection{Challenges}
Despite rapid progress, existing large models in sports still encounter 4 fundamental challenges that hinder robust and trustworthy real-world adoption.

\noindent\textbf{Bias, Fairness, and Privacy.}  
Current sports datasets are heavily skewed toward a small set of popular sports such as soccer and basketball~\citep{deliege2021soccernetV2, app4fans_com_xi2025player}, leaving many less popular sports largely underrepresented. Moreover, existing research disproportionately focuses on data from elite, mainstream competitions, while settings such as the Paralympics, youth development programs, and school sports remain underexplored. Current datasets and models also predominantly focus on men’s sports, while women’s leagues and competitions remain substantially underrepresented~\citep{biester2025sports}. These biases may lead to unfair model behaviors and limit generalization across diverse sporting contexts~\citep{app4res_dergaa2023human, papini2025balancing}.


\noindent\textbf{Real-Time and High-FPS Understanding.}
Sports applications such as live officiating and broadcast commentary are highly latency-sensitive, yet current Video LLMs still incur substantial inference overhead~\citep{app4fans_com_you2025timesoccer}, limiting their use in time-critical workflows. Moreover, long-video understanding remains difficult: sports broadcasts often last for hours and demand sustained temporal reasoning over extended contexts~\citep{zou2024seconds}. Finally, most generic video pipelines are not designed for high-frame-rate inputs. In sports, decisive cues (e.g., ball contact, offside timing, foul initiation) can unfold within milliseconds; aggressive temporal downsampling removes these fine-grained dynamics, degrading event localization and rule-level judgments~\citep{app4coa_ana_li2025improving}.


\noindent\textbf{Hallucination and Interpretability.}
In practical sports workflows, stakeholders require explanations that directly support decisions, rather than descriptive summaries. For example, coaches and analysts need to identify actionable causes and detailed explanations, which remains challenging for current black-box models~\citep{mersha2024explainable}. Meanwhile, hallucinations in sports often invent key events, actors, or causal links, producing plausible narratives that can directly mislead downstream decisions. Such errors are especially harmful in high-stakes settings as they can quickly erode user trust~\citep{app4ath_pla_qiu2024impact, app4ref_ref_app4held2024x}.


\noindent\textbf{Practicability and Real-World Deployment.}
Recent research still falls short of real sports workflows in terms of ecological validity. Although models can perform well on curated clips, they often break down in the wild due to heavy occlusion, shifting camera viewpoints, and low-quality footage common in sports scenarios~\citep{niu2025ovobench}. At the same time, real-world deployment remains largely underexplored: practical setups such as deploying models on the sidelines or on edge devices like wearables and drone cameras remain challenging and are rarely validated in real-world settings~\citep{koh2021wilds, bandraupalli2025vlms}.

\subsection{Future Directions}
Building on the challenges discussed above, we outline 4 future directions to advance large models in sports toward robust real-world use.

\noindent\textbf{Trustworthy Sport AI.}
Future research should address these ethical and reliability challenges by integrating advanced technical safeguards into model development. Key directions include implementing rigorous data balancing and cleaning~\citep{bai2022constitutional} alongside sport-specific alignment via Reinforcement Learning from Human Feedback (RLHF) to minimize bias~\citep{yu2024rlhf, gallegos2024bias}. 


\noindent\textbf{Streaming and Long Video Mechanisms.}
Future work should better handle sports’ temporal demands. For low latency, explore streaming inference with more efficient KV-cache and attention mechanisms~\citep{chen2024videollm, ding2025streammind, xu2025streamingvlm}. For long matches, adapt long-video modeling via memory, parallelism, and token compression, or architectures like Mamba~\citep{ren2024timechat, app4fans_com_you2025timesoccer, chenlongvila, wang2024longllava}. For high-FPS events, develop vision backbones that support dense frames without losing fine-grained dynamics~\citep{app4coa_ana_li2025improving}.


\noindent\textbf{Knowledge Grounding and Tool Use.}
Future work should improve factual reliability via explicit retrieval from structured knowledge (e.g., RAG)~\citep{app4fans_mod_strand2024soccer,app4fans_mod_sepasdar2024soccer}, grounding claims in visual evidence~\citep{xia2025sportr}, and producing explicit rationales~\citep{app4ref_ref_app4held2024x}. Tool-enabled models that query live databases, rule engines, or match-tracking APIs can further make outputs verifiable and logically consistent~\citep{app4fans_mod_rao2025multi}.


\noindent\textbf{In-the-Wild Evaluation and Edge Deployment.}
Future work should prioritize practical deployment by developing ecologically valid, workflow- and latency-aware in-the-wild benchmarks that stress-test models under real sports conditions~\citep{wang2024tacticai, bandraupalli2025vlms}. In parallel, enabling on-device inference requires efficiency advances such as quantization~\citep{zhu2024survey, lin2024awq}, knowledge distillation~\citep{xu2024survey}, and mobile (M)LLMs~\citep{lu2024small, van2024survey} to support local analysis on wearables or mobile platforms.

\section{Conclusion}
\label{sec:concl}



This survey reviews the emerging landscape of large models in sports, establishing a structured taxonomy that spans 6 stakeholder groups. We provide a deep analysis of relevant datasets, and highlight fundamental challenges. By consolidating these disparate research efforts, we aim to establish a solid framework for future exploration. We hope this work serves as a foundation for advancing large-model-driven sports intelligence and provides a practical resource for research and development.

\section*{Limitations}


Although this survey strives to provide a comprehensive overview of large models in sports, several limitations remain. Firstly, given the rapid development of this field, our survey may not be able to timely reflect the latest progress before and after the survey.
Secondly, our literature selection primarily follows standard protocols focused on English-language publications. This may naturally limit the coverage of domestic research in other regions or studies published in other languages.
Thirdly, our analysis objectively reflects the current research imbalance in the field, which is heavily skewed toward a few dominant sports. Consequently, this leads to a lack of in-depth coverage for underrepresented or niche sporting scenarios in our survey.
Fourthly, as some studies span multiple application domains, minor overlaps are inevitable; we categorize each work based on its primary research focus while cross-referencing related sections when appropriate.
Finally, our analysis primarily centers on academic research, and the discussion of commercial systems or industrial applications remains limited.
Despite these limitations, this survey provides a valuable and timely overview of the field, offering a solid reference for subsequent research and development.

\section*{Acknowledgments}

We thank all reviewers for their insightful comments and suggestions. This work was partially supported by the Beijing Natural Science Foundation (No. L233008).


\bibliography{custom}

@article{team2023gemini,
  title={Gemini: a family of highly capable multimodal models},
  author={Team, Gemini and Anil, Rohan and Borgeaud, Sebastian and Alayrac, Jean-Baptiste and Yu, Jiahui and Soricut, Radu and Schalkwyk, Johan and Dai, Andrew M and Hauth, Anja and Millican, Katie and others},
  journal={arXiv preprint arXiv:2312.11805},
  year={2023}
}

@article{achiam2023gpt4,
  title={Gpt-4 technical report},
  author={Achiam, Josh and Adler, Steven and Agarwal, Sandhini and Ahmad, Lama and Akkaya, Ilge and Aleman, Florencia Leoni and Almeida, Diogo and Altenschmidt, Janko and Altman, Sam and Anadkat, Shyamal and others},
  journal={arXiv preprint arXiv:2303.08774},
  year={2023}
}

@article{hutchins2016tales,
  title={Tales of the digital sublime: Tracing the relationship between big data and professional sport},
  author={Hutchins, Brett},
  journal={Convergence},
  volume={22},
  number={5},
  pages={494--509},
  year={2016},
  publisher={SAGE Publications Sage UK: London, England}
}

@article{zhou2025artificial,
  title={Artificial intelligence in sport: A narrative review of applications, challenges and future trends},
  author={Zhou, Diwei and Keogh, Justin WL and Ma, Yingliang and Tong, Raymond KY and Khan, Abdul R and Jennings, Nicholas R},
  journal={Journal of Sports Sciences},
  pages={1--16},
  year={2025},
  publisher={Taylor \& Francis}
}

@article{cossich2023technological,
  title={Technological breakthroughs in sport: Current practice and future potential of artificial intelligence, virtual reality, augmented reality, and modern data visualization in performance analysis},
  author={Cossich, Victor RA and Carlgren, Dave and Holash, Robert John and Katz, Larry},
  journal={Applied Sciences},
  volume={13},
  number={23},
  pages={12965},
  year={2023},
  publisher={MDPI}
}

@article{naik2022comprehensive,
  title={A comprehensive review of computer vision in sports: Open issues, future trends and research directions},
  author={Naik, Banoth Thulasya and Hashmi, Mohammad Farukh and Bokde, Neeraj Dhanraj},
  journal={Applied Sciences},
  volume={12},
  number={9},
  pages={4429},
  year={2022},
  publisher={MDPI}
}

@article{lai2025using,
  title={Using Large Language Models to Enhance Exercise Recommendations and Physical Activity in Clinical and Healthy Populations: Scoping Review},
  author={Lai, Xiangxun and Chen, Jiacheng and Lai, Yue and Huang, Shengqi and Cai, Yongdong and Sun, Zhifeng and Wang, Xueding and Pan, Kaijiang and Gao, Qi and Huang, Caihua},
  journal={JMIR Medical Informatics},
  volume={13},
  number={1},
  pages={e59309},
  year={2025},
  publisher={JMIR Publications Inc., Toronto, Canada}
}

@article{connor2023large,
  title={Large language models in sport science \& medicine: Opportunities, risks and considerations},
  author={Connor, Mark and O'Neill, Michael},
  journal={arXiv preprint arXiv:2305.03851},
  year={2023}
}

@article{naughton2024challenges,
  title={Challenges and opportunities of artificial intelligence implementation within sports science and sports medicine teams},
  author={Naughton, Mitchell and Salmon, Paul M and Compton, Heidi R and McLean, Scott},
  journal={Frontiers in Sports and Active Living},
  volume={6},
  pages={1332427},
  year={2024},
  publisher={Frontiers Media SA}
}

@article{wang2024impact,
  title={Impact of ChatGPT technology on sports industry},
  author={Wang, Zefeng and Hu, Yueke and Liu, Jiaxi and Hu, Lianxin},
  journal={Journal of New Media and Economics},
  volume={1},
  number={4},
  pages={29--37},
  year={2024}
}

@article{zhao2025survey,
  title={A survey of deep learning in sports applications: Perception, comprehension, and decision},
  author={Zhao, Zhonghan and Chai, Wenhao and Hao, Shengyu and Hu, Wenhao and Wang, Guanhong and Cao, Shidong and Song, Mingli and Hwang, Jenq-Neng and Wang, Gaoang},
  journal={IEEE Transactions on Visualization and Computer Graphics},
  year={2025},
  publisher={IEEE}
}

@article{xia2024language,
  title={Language and multimodal models in sports: a survey of datasets and applications},
  author={Xia, Haotian and Yang, Zhengbang and Zhao, Yun and Wang, Yuqing and Li, Jingxi and Tracy, Rhys and Zhu, Zhuangdi and Wang, Yuan-fang and Chen, Hanjie and Shen, Weining},
  journal={arXiv preprint arXiv:2406.12252},
  year={2024}
}

@inproceedings{huang2020generating,
  title={Generating sports news from live commentary: A Chinese dataset for sports game summarization},
  author={Huang, Kuan-Hao and Li, Chen and Chang, Kai-Wei},
  booktitle={Proceedings of the 1st Conference of the Asia-Pacific Chapter of the Association for Computational Linguistics and the 10th International Joint Conference on Natural Language Processing},
  pages={609--615},
  year={2020}
}

@article{bunker2022application,
  title={The application of machine learning techniques for predicting match results in team sport: A review},
  author={Bunker, Rory and Susnjak, Teo},
  journal={Journal of Artificial Intelligence Research},
  volume={73},
  pages={1285--1322},
  year={2022}
}

@inproceedings{skerik2018automated,
  title={Automated training plan generation for athletes},
  author={Skerik, Tomas and Chrpa, Lukas and Faber, Wolfgang and Vallati, Mauro},
  booktitle={2018 IEEE international conference on systems, man, and cybernetics (SMC)},
  pages={3865--3870},
  year={2018},
  organization={IEEE}
}

@article{app4ath_pla_phillips2012exercise,
  title={The exercise prescription: a tool to improve physical activity},
  author={Phillips, Edward M and Kennedy, Mary A},
  journal={Pm\&r},
  volume={4},
  number={11},
  pages={818--825},
  year={2012},
  publisher={Elsevier}
}

@article{app4ath_pla_wackerhage2021personalized,
  title={Personalized, evidence-informed training plans and exercise prescriptions for performance, fitness and health},
  author={Wackerhage, Henning and Schoenfeld, Brad J},
  journal={Sports Medicine},
  volume={51},
  number={9},
  pages={1805--1813},
  year={2021},
  publisher={Springer}
}

@article{app4ath_pla_puce2025harnessing,
  title={Harnessing Generative Artificial Intelligence for Exercise and Training Prescription: Applications and Implications in Sports and Physical Activity—A Systematic Literature Review.},
  author={Puce, Luca and Bragazzi, Nicola Luigi and Curra, Antonio and Trompetto, Carlo},
  journal={Applied Sciences (2076-3417)},
  volume={15},
  number={7},
  year={2025}
}

@article{app4ath_pla_cavazzotto2024chatgpt,
  title={ChatGPT and exercise prescription: Human vs. machine or human plus machine?},
  author={Cavazzotto, Timothy Gustavo and Dantas, Diego Bessa and Queiroga, Marcos Roberto},
  journal={Journal of Sport and Health Science},
  volume={13},
  number={5},
  pages={661--662},
  year={2024},
  publisher={Elsevier}
}

@article{app4ath_pla_washif2024artificial,
  title={Artificial intelligence in sport: Exploring the potential of using ChatGPT in resistance training prescription},
  author={Washif, Jad and Pagaduan, Jeffrey and James, Carl and Dergaa, Ismail and Beaven, Christopher},
  journal={Biology of sport},
  volume={41},
  number={2},
  pages={209--220},
  year={2024},
  publisher={Termedia}
}

@inproceedings{app4ath_pla_li2025visualizing,
  title={Visualizing Exercise Data from Combat Exergame for Exploring the Insight from Personal Informatics with Large Language Models},
  author={Li, Zong-Ying and Lai, Yen-Hua and Lin, Chi-Yu and Chou, Chien-Hsing and Han, Ping-Hsuan},
  booktitle={Proceedings of the Extended Abstracts of the CHI Conference on Human Factors in Computing Systems},
  pages={1--8},
  year={2025}
}

@article{app4ath_pla_dergaa2024using,
  title={Using artificial intelligence for exercise prescription in personalised health promotion: A critical evaluation of OpenAI’s GPT-4 model},
  author={Dergaa, Ismail and Saad, Helmi Ben and El Omri, Abdelfatteh and Glenn, Jordan and Clark, Cain and Washif, Jad and Guelmami, Noomen and Hammouda, Omar and Al-Horani, Ramzi and Reynoso-S{\'a}nchez, Luis and others},
  journal={Biology of Sport},
  volume={41},
  number={2},
  pages={221--241},
  year={2024},
  publisher={Termedia}
}

@article{app4ath_pla_havers2025reproducibility,
  title={Reproducibility and quality of hypertrophy-related training plans generated by GPT-4 and Google Gemini as evaluated by coaching experts},
  author={Havers, Tim and Masur, Lukas and Isenmann, Eduard and Geisler, Stephan and Zinner, Christoph and Sperlich, Billy and D{\"u}king, Peter},
  journal={Biology of Sport},
  volume={42},
  number={2},
  pages={289--329},
  year={2025},
  publisher={Termedia}
}

@article{app4ath_pla_saracc2025evaluating,
  title={Evaluating the potential role of AI chatbots in designing personalized exercise programs for weight management},
  author={Sara{\c{c}}, Hakan and Ulusoy, {\.I}smet Tar{\i}k and Alpay, Janset and {\"O}demi{\c{s}}, Hasan and S{\"o}{\u{g}}{\"u}t, Mustafa},
  journal={International Journal of Human--Computer Interaction},
  pages={1--8},
  year={2025},
  publisher={Taylor \& Francis}
}

@article{app4ath_pla_wachholz2025acceptance,
  title={Acceptance and trust in AI-generated exercise plans among recreational athletes and quality evaluation by experienced coaches: a pilot study},
  author={Wachholz, Felix and Manno, Stefano and Schlachter, Daniel and Gamper, Nicole and Schnitzer, Martin},
  journal={BMC Research Notes},
  volume={18},
  number={1},
  pages={112},
  year={2025},
  publisher={Springer}
}

@article{app4ath_pla_akrimi2025chatgpt,
  title={ChatGPT-4o-Generated Exercise Plans for Patients with Type 2 Diabetes Mellitus—Assessment of Their Safety and Other Quality Criteria by Coaching Experts},
  author={Akrimi, Samir and Schwensfeier, Leon and D{\"u}king, Peter and Kreutz, Thorsten and Brinkmann, Christian},
  journal={Sports},
  volume={13},
  number={4},
  pages={92},
  year={2025},
  publisher={MDPI}
}

@inproceedings{app4ath_pla_jorke2025gptcoach,
  title={GPTCoach: Towards LLM-Based Physical Activity Coaching},
  author={J{\"o}rke, Matthew and Sapkota, Shardul and Warkenthien, Lyndsea and Vainio, Niklas and Schmiedmayer, Paul and Brunskill, Emma and Landay, James A},
  booktitle={Proceedings of the 2025 CHI Conference on Human Factors in Computing Systems},
  pages={1--46},
  year={2025}
}

@article{app4ath_pla_canuzakov2025digital,
  title={Digital coaches: an alternative to expert coaches for men's fitness goals},
  author={CANUZAKOV, Kanat and ABDIRAHMANOVA, Cipark{\"u}l and DEM{\.I}RHAN, Bilal and others},
  year={2025},
  publisher={PPHU Projack, Poland}
}

@inproceedings{app4ath_pla_ko2025legolas,
  title={LEGOLAS: Learning \& Enhancing Golf Skills through LLM-Augmented System},
  author={Ko, Kangbeen and Oh, Minwoo and Seong, Minwoo and Kim, SeungJun},
  booktitle={Proceedings of the Extended Abstracts of the CHI Conference on Human Factors in Computing Systems},
  pages={1--10},
  year={2025}
}

@inproceedings{app4ath_pla_vahdati2025multi,
  title={A Multi-Agent Digital Twin Framework for AI-Driven Fitness Coaching},
  author={Vahdati, Monica and Gholizadeh HamlAbadi, Kamran and Laamarti, Fedwa and El Saddik, Abdulmotaleb},
  booktitle={Proceedings of the 2025 ACM International Conference on Interactive Media Experiences},
  pages={380--385},
  year={2025}
}

@article{app4ath_pla_li2025gpt,
  title={GPT-4 as a virtual fitness coach: a case study assessing its effectiveness in providing weight loss and fitness guidance},
  author={Li, Guanchong and Li, Hansen and Su, Yuqin and Li, Yun and Jiang, Sijia and Zhang, Guodong},
  journal={BMC Public Health},
  volume={25},
  number={1},
  pages={2466},
  year={2025},
  publisher={Springer}
}

@inproceedings{app4ath_pla_ma2025t3set,
  title={T3Set: A Multimodal Dataset with Targeted Suggestions for LLM-based Virtual Coach in Table Tennis Training},
  author={Ma, Ji and Wu, Jiale and Wang, Haoyu and Zhang, Yanze and Xie, Xiao and Zhou, Zheng and Zhang, Hui and Wang, Jiachen and Wu, Yingcai},
  booktitle={Proceedings of the 31st ACM SIGKDD Conference on Knowledge Discovery and Data Mining V. 2},
  pages={5686--5697},
  year={2025}
}

@article{app4ath_pla_ma2025table,
  title={Table tennis coaching system based on a multimodal large language model with a table tennis knowledge base},
  author={Ma, Wenlong and Liu, Yang and Yi, Qing and Liu, Xutao and Xing, Wei and Zhao, Rongji and Liu, Huan and Li, Rongzhi},
  journal={PloS one},
  volume={20},
  number={2},
  pages={e0317839},
  year={2025},
  publisher={Public Library of Science San Francisco, CA USA}
}

@inproceedings{app4ath_pla_yeh2023maaig,
  title={MAAIG: Motion Analysis And Instruction Generation},
  author={Yeh, Wei-Hsin and Lin, Pei Hsin and Su, Yu-An and Cheng, Wen Hsiang and Ku, Lun-Wei},
  booktitle={Proceedings of the 5th ACM International Conference on Multimedia in Asia Workshops},
  pages={1--5},
  year={2023}
}

@inproceedings{app4ath_pla_yeh2025coachme,
  title={CoachMe: Decoding Sport Elements with a Reference-Based Coaching Instruction Generation Model},
  author={Yeh, Wei-Hsin and Su, Yu-An and Chen, Chih-Ning and Lin, Yi-Hsueh and Ku, Calvin and Chiu, Wenhsin and Hu, Min-Chun and Ku, Lun-Wei},
  booktitle={Proceedings of the 63rd Annual Meeting of the Association for Computational Linguistics (Volume 1: Long Papers)},
  pages={29126--29151},
  year={2025}
}

@article{app4ath_pla_zhu2024could,
  title={Who could and should give exercise prescription: Physicians, exercise and health scientists, fitness trainers, or ChatGPT?},
  author={Zhu, Weimo and Geng, Wenguang and Huang, Lingling and Qin, Xiong and Chen, Zezhao and Yan, Hai},
  journal={Journal of Sport and Health Science},
  volume={13},
  number={3},
  pages={368--372},
  year={2024},
  publisher={Elsevier}
}

@article{app4ath_pla_erol2024does,
  title={Does ChatGPT provide comprehensive and accurate information regarding the effects, types and programming of core exercises?},
  author={Erol, Erkan and Ar{\i}kan, Halime},
  journal={Turkish Journal of Kinesiology},
  volume={10},
  number={3},
  pages={178--182},
  year={2024},
  publisher={Nurtekin ERKMEN}
}

@article{app4ath_pla_duking2024chatgpt,
  title={ChatGPT generated training plans for runners are not rated optimal by coaching experts, but increase in quality with additional input information},
  author={D{\"u}king, Peter and Sperlich, Billy and Voigt, Laura and Van Hooren, Bas and Zanini, Michele and Zinner, Christoph},
  journal={Journal of sports science \& medicine},
  volume={23},
  number={1},
  pages={56},
  year={2024}
}

@inproceedings{app4ath_pla_zhang2025rag,
  title={RAG-LLM based evaluation pathway and technological exploration for the scientific validity of mass fitness},
  author={Zhang, Kexin and Qin, Yulin and Qin, Baosheng},
  booktitle={Of Papers Presented at 2025 6th Asia Sport Science Conference (ASSC)},
  year={2025}
}

@article{app4ath_pla_puce2024optimizing,
  title={Optimizing athletic performance through advanced nutrition strategies: can AI and digital platforms have a role in ultraendurance sports?},
  author={Puce, Luca and Ceylan, Halil {\.I}brahim and Trompetto, Carlo and Cotellessa, Filippo and Schenone, Cristina and Marinelli, Lucio and Zmijewski, Piotr and Bragazzi, Nicola and Mori, Laura},
  journal={Biology of Sport},
  volume={41},
  number={4},
  pages={305--313},
  year={2024},
  publisher={Termedia}
}

@article{app4ath_pla_solomon2025sports,
  title={The sports nutrition knowledge of large language model (LLM) artificial intelligence (AI) chatbots: An assessment of accuracy, completeness, clarity, quality of evidence, and test-retest reliability},
  author={Solomon, Thomas PJ and Laye, Matthew J},
  journal={PloS one},
  volume={20},
  number={6},
  pages={e0325982},
  year={2025},
  publisher={Public Library of Science San Francisco, CA USA}
}

@article{app4ath_pla_rocha2025can,
  title={Can people with epilepsy trust AI chatbots for information on physical exercise?},
  author={Rocha-Silva, Rizia and de Lima, Braulio Evangelista and Costa, Thalles Guilarducci and Morais, Naiane Silva and Jose, Geovana and Cordeiro, Douglas Farias and de Almeida, Alexandre Aparecido and Lopim, Glauber Menezes and Viana, Ricardo Borges and Sousa, Bolivar Saldanha and others},
  journal={Epilepsy \& Behavior},
  volume={163},
  pages={110193},
  year={2025},
  publisher={Elsevier}
}

@article{app4ath_pla_hegde2024infusing,
  title={Infusing behavior science into large language models for activity coaching},
  author={Hegde, Narayan and Vardhan, Madhurima and Nathani, Deepak and Rosenzweig, Emily and Speed, Cathy and Karthikesalingam, Alan and Seneviratne, Martin},
  journal={PLOS Digital Health},
  volume={3},
  number={4},
  pages={e0000431},
  year={2024},
  publisher={Public Library of Science San Francisco, CA USA}
}

@article{app4ath_pla_dindorf2025characteristics,
  title={Characteristics and perceived suitability of artificial intelligence-driven sports coaches: a pilot study on psychological and perceptual factors},
  author={Dindorf, Carlo and Dully, Jonas and Bartaguiz, Eva and Menges, Tessa and Reidick, Claudia and Seibert, Johann-Nikolaus and Fr{\"o}hlich, Michael},
  journal={Frontiers in Sports and Active Living},
  volume={7},
  pages={1548980},
  year={2025},
  publisher={Frontiers Media SA}
}

@article{app4ath_pla_han2025intent,
  title={Intent-aware personalized feedback generation from coach-athlete dialogues in sports training},
  author={Han, Yuxin},
  journal={Journal of King Saud University Computer and Information Sciences},
  volume={37},
  number={6},
  pages={1--17},
  year={2025},
  publisher={Springer}
}

@article{app4ath_pla_qiu2024impact,
  title={The impact of LLM hallucinations on motor skill learning: A case study in badminton},
  author={Qiu, Yepeng},
  journal={IEEE Access},
  year={2024},
  publisher={IEEE}
}

@inproceedings{app4ath_pla_bullard2025enhancing,
  title={Enhancing Athletic Performance Through AI: An Iterative Prompt Engineering Approach for LLM-Based Coaching Feedback},
  author={Bullard, Enya and Khan, Nibraas and Sarkar, Nilanjan},
  booktitle={International Conference on Human-Computer Interaction},
  pages={251--255},
  year={2025},
  organization={Springer}
}

@inproceedings{app4ath_pla_ashutosh2025expertaf,
  title={ExpertAF: Expert actionable feedback from video},
  author={Ashutosh, Kumar and Nagarajan, Tushar and Pavlakos, Georgios and Kitani, Kris and Grauman, Kristen},
  booktitle={Proceedings of the Computer Vision and Pattern Recognition Conference},
  pages={13582--13594},
  year={2025}
}

@article{app4ath_pla_seino2025expert,
  title={Expert Comment Generation Considering Sports Skill Level Using a Large Multimodal Model with Video and Spatial-Temporal Motion Features},
  author={Seino, Tatsuki and Saito, Naoki and Ogawa, Takahiro and Asamizu, Satoshi and Haseyama, Miki},
  journal={Sensors},
  volume={25},
  number={2},
  pages={447},
  year={2025},
  publisher={MDPI}
}

@article{app4ath_pla_papini2025balancing,
  title={Balancing Act: Generative AI Tools and Scope of Practice in Health Coaching},
  author={Papini, Natalie M and Meyer, Megan and Squires, Nikole D and Clifford, Dawn},
  journal={American Journal of Health Promotion},
  pages={08901171251340383},
  year={2025},
  publisher={SAGE Publications Sage CA: Los Angeles, CA}
}

@inproceedings{app4ath_pla_cosentinotowards,
  title={Towards a Personal Health Large Language Model},
  author={Cosentino, Justin and Belyaeva, Anastasiya and Liu, Xin and Yang, Zhun and Liu, Yun and Tailor, Shyam A and Althoff, Tim and Hernandez, John B and Matias, Yossi and Corrado, Greg and others},
  booktitle={Advancements In Medical Foundation Models: Explainability, Robustness, Security, and Beyond},
  year={2024}
}

@article{app4ath_pla_rocha2024potential,
  title={The potential of large language model chatbots for application to epilepsy: let’s talk about physical exercise},
  author={Rocha-Silva, Rizia and de Lima, Br{\'a}ulio Evangelista and Jos{\'e}, Geovana and Cordeiro, Douglas Farias and Viana, Ricardo Borges and Andrade, Mar{\'\i}lia Santos and Vancini, Rodrigo Luiz and Rosemann, Thomas and Weiss, Katja and Knechtle, Beat and others},
  journal={Epilepsy \& Behavior Reports},
  volume={27},
  pages={100692},
  year={2024},
  publisher={Elsevier}
}

@article{app4ath_pla_onan2025examining,
  title={Examining the ability of artificial intelligence with ChatGPT-4.0 to create an exercise program: Case scenario examples" lumbar disc herniation, chronic migraine, and urge urinary incontinence"},
  author={Onan, Dilara and Ar{\i}kan, Halime and Can, {\.I}rem and G{\"u}ven, {\c{S}}ahan and I{\c{s}}{\i}kay, Levent and Ozge, Aynur},
  journal={Turkish Journal of Kinesiology},
  volume={11},
  number={1},
  pages={28--44},
  year={2025},
  publisher={Nurtekin ERKMEN}
}

@article{app4ath_inj_zhu2025full,
  title={Full-Parameter Fine-Tuning Method of LLMs for Sports Injury Prevention and Treatment},
  author={Zhu, Xinli and Gao, Zhiqiang and Wang, Xu An},
  journal={International Journal of Mobile Computing and Multimedia Communications (IJMCMC)},
  volume={16},
  number={1},
  pages={1--14},
  year={2025},
  publisher={IGI Global Scientific Publishing}
}

@article{app4ath_inj_hasnain2023role,
  title={The role of ChatGPT in sports trauma: A mini review on strengths and limits of open AI application},
  author={Hasnain, Muhammad and Mehboob, Bilal and Imran, Shahid},
  journal={Discover Artificial Intelligence},
  volume={3},
  number={1},
  pages={40},
  year={2023},
  publisher={Springer}
}

@article{app4ath_inj_mcbee2023interdisciplinary,
  title={Interdisciplinary inquiry via PanelGPT: application to explore chatbot application in sports rehabilitation},
  author={McBee, Joseph C and Han, Daniel Y and Liu, Li and Ma, Leah and Adjeroh, Donald A and Xu, Dong and Hu, Gangqing},
  journal={medRxiv},
  year={2023}
}

@article{app4ath_inj_cheng2023artificial,
  title={Artificial intelligence in sports medicine: could GPT-4 make human doctors obsolete?},
  author={Cheng, Kunming and Guo, Qiang and He, Yongbin and Lu, Yanqiu and Xie, Ruijie and Li, Cheng and Wu, Haiyang},
  journal={Annals of Biomedical Engineering},
  volume={51},
  number={8},
  pages={1658--1662},
  year={2023},
  publisher={Springer}
}

@article{app4ath_inj_fayed2023artificial,
  title={Artificial intelligence and ChatGPT in Orthopaedics and sports medicine},
  author={Fayed, Aly M and Mansur, Nacime Salomao Barbachan and de Carvalho, Kepler Alencar and Behrens, Andrew and D’Hooghe, Pieter and de Cesar Netto, Cesar},
  journal={Journal of Experimental Orthopaedics},
  volume={10},
  number={1},
  pages={74},
  year={2023},
  publisher={Springer}
}

@article{app4ath_inj_mcbee2024assessing,
  title={Assessing ChatGPT’s competency in addressing interdisciplinary inquiries on Chatbot uses in sports rehabilitation: simulation study},
  author={McBee, Joseph C and Han, Daniel Y and Liu, Li and Ma, Leah and Adjeroh, Donald A and Xu, Dong and Hu, Gangqing},
  journal={JMIR Medical Education},
  volume={10},
  number={1},
  pages={e51157},
  year={2024},
  publisher={JMIR Publications Inc., Toronto, Canada}
}

@incollection{app4ath_inj_lotfi2024evaluating,
  title={Evaluating the Qualitative and Quantitative Performance of Generative AI on Knowledge in Sports Medicine: The Case of GPT},
  author={Lotfi, Nizar and Madani, Mohamed},
  booktitle={General Aspects of Applying Generative AI in Higher Education: Opportunities and Challenges},
  pages={103--119},
  year={2024},
  publisher={Springer}
}

@inproceedings{app4ath_inj_brogly2025evaluation,
  title={Evaluation of the phi-3-mini SLM for identification of texts related to medicine, health, and sports injuries},
  author={Brogly, Chris and Rjaibi, Saif and Liang, Charlotte and Lam, Erica and Wang, Edward and Paleczny, Sarah and Levitan, Adam and Cusimano, Michael D},
  booktitle={2025 IEEE 4th International Conference on Computing and Machine Intelligence (ICMI)},
  pages={1--5},
  year={2025},
  organization={IEEE}
}

@article{app4ath_inj_musat2024diagnostic,
  title={Diagnostic applications of AI in sports: a comprehensive review of injury risk prediction methods},
  author={Musat, Carmina Liana and Mereuta, Claudiu and Nechita, Aurel and Tutunaru, Dana and Voipan, Andreea Elena and Voipan, Daniel and Mereuta, Elena and Gurau, Tudor Vladimir and Gur{\u{a}}u, Gabriela and Nechita, Luiza Camelia},
  journal={Diagnostics},
  volume={14},
  number={22},
  pages={2516},
  year={2024},
  publisher={MDPI}
}

@article{app4ath_inj_saglam2025comparative,
  title={Comparative evaluation of artificial intelligence models GPT-4 and GPT-3.5 in clinical decision-making in sports surgery and physiotherapy: a cross-sectional study},
  author={Saglam, S{\"o}nmez and Uludag, Veysel and Karaduman, Zekeriya Okan and Ar{\i}can, Mehmet and Y{\"u}cel, M{\"u}cahid Osman and Dalaslan, Ra{\c{s}}it Emin},
  journal={BMC Medical Informatics and Decision Making},
  volume={25},
  number={1},
  pages={163},
  year={2025},
  publisher={Springer}
}

@article{app4ath_inj_ahsan2023chatbot,
  title={Chatbot Generative Pre-Trained Transformer and artificial intelligence in sports physical therapy and rehabilitation},
  author={Ahsan, Mohammad},
  journal={Saudi Journal of Sports Medicine},
  volume={23},
  number={2},
  pages={61--62},
  year={2023},
  publisher={Medknow}
}

@article{app4ath_psy_szabo2023chatgpt,
  title={ChatGPT is a breakthrough in science and education but fails a test in sports and exercise psychology},
  author={Szabo, Attila},
  journal={Baltic Journal of Sport and Health Sciences},
  volume={1},
  number={128},
  pages={25--40},
  year={2023}
}

@article{app4ath_psy_oliver2025generative,
  title={Generative AI in sport and exercise psychology: exploring opportunities and overcoming challenges},
  author={Oliver, Alex and Guiller, Jane},
  journal={Sport and Exercise Psychology Review},
  volume={19},
  number={2},
  pages={36--45},
  year={2025},
  publisher={British Psychological Society}
}

@article{app4ath_psy_masur2025assessment,
  title={Assessment of Recommendations Provided to Athletes Regarding Sleep Education by GPT-4o and Google Gemini: Comparative Evaluation Study},
  author={Masur, Lukas and Driller, Matthew and Suppiah, Haresh and Matzka, Manuel and Sperlich, Billy and D{\"u}king, Peter},
  journal={JMIR Formative Research},
  volume={9},
  number={1},
  pages={e71358},
  year={2025},
  publisher={JMIR Publications Inc., Toronto, Canada}
}

@article{app4ath_psy_vandelanotte2023increasing,
  title={Increasing physical activity using an just-in-time adaptive digital assistant supported by machine learning: a novel approach for hyper-personalised mHealth interventions},
  author={Vandelanotte, Corneel and Trost, Stewart and Hodgetts, Danya and Imam, Tasadduq and Rashid, Mamunur and To, Quyen G and Maher, Carol},
  journal={Journal of Biomedical Informatics},
  volume={144},
  pages={104435},
  year={2023},
  publisher={Elsevier}
}

@article{app4ath_psy_zuccolotto11,
  title={11 th MathSport International Conference 4-6 June 2025},
  author={Zuccolotto, Paola},
year={2025}
}

@article{app4ath_psy_merrill2024transforming,
  title={Transforming wearable data into personal health insights using large language model agents},
  author={Merrill, Mike A and Paruchuri, Akshay and Rezaei, Naghmeh and Kovacs, Geza and Perez, Javier and Liu, Yun and Schenck, Erik and Hammerquist, Nova and Sunshine, Jake and Tailor, Shyam and others},
  journal={Nature Communications},
  year={2026},
  publisher={Nature Publishing Group UK London}
}

@article{app4ath_psy_ferrara2024large,
  title={Large language models for wearable sensor-based human activity recognition, health monitoring, and behavioral modeling: A survey of early trends, datasets, and challenges},
  author={Ferrara, Emilio},
  journal={Sensors},
  volume={24},
  number={15},
  pages={5045},
  year={2024},
  publisher={MDPI}
}

@article{app4ath_psy_song2025investigating,
  title={Investigating the Relationship Between Physical Activity and Tailored Behavior Change Messaging: Connecting Contextual Bandit with Large Language Models},
  author={Song, Haochen and Hofer, Dominik and Islambouli, Rania and Hawkins, Laura and Bhattacharjee, Ananya and Franklin, Meredith and Williams, Joseph Jay},
  journal={arXiv preprint arXiv:2506.07275},
  year={2025}
}

@article{app4ath_psy_imran2024llasa,
  title={Llasa: Large multimodal agent for human activity analysis through wearable sensors},
  author={Imran, Sheikh Asif and Khan, Mohammad Nur Hossain and Biswas, Subrata and Islam, Bashima},
  journal={arXiv preprint arXiv:2406.14498},
  volume={3},
  number={4},
  year={2024}
}

@inproceedings{app4ath_psy_ji2024hargpt,
  title={Hargpt: Are llms zero-shot human activity recognizers?},
  author={Ji, Sijie and Zheng, Xinzhe and Wu, Chenshu},
  booktitle={2024 IEEE International Workshop on Foundation Models for Cyber-Physical Systems \& Internet of Things (FMSys)},
  pages={38--43},
  year={2024},
  organization={IEEE}
}

@article{app4res_methnani2023chatgpt,
  title={ChatGPT for sample-size calculation in sports medicine and exercise sciences: A cautionary note},
  author={Methnani, Jabeur and Latiri, Imed and Dergaa, Ismail and Chamari, Karim and Saad, Helmi Ben},
  journal={International Journal of Sports Physiology and Performance},
  volume={18},
  number={10},
  pages={1219--1223},
  year={2023},
  publisher={Human Kinetics}
}

@misc{app4res_anderson2023ai,
  title={AI did not write this manuscript, or did it? Can we trick the AI text detector into generated texts? The potential future of ChatGPT and AI in Sports \& Exercise Medicine manuscript generation},
  author={Anderson, Nash and Belavy, Daniel L and Perle, Stephen M and Hendricks, Sharief and Hespanhol, Luiz and Verhagen, Evert and Memon, Aamir R},
  journal={BMJ open sport \& exercise medicine},
  volume={9},
  number={1},
  pages={e001568},
  year={2023},
  publisher={BMJ Specialist Journals}
}

@article{app4res_dergaa2023human,
  title={From human writing to artificial intelligence generated text: examining the prospects and potential threats of ChatGPT in academic writing},
  author={Dergaa, Ismail and Chamari, Karim and Zmijewski, Piotr and Saad, Helmi Ben},
  journal={Biology of sport},
  volume={40},
  number={2},
  pages={615--622},
  year={2023},
  publisher={Termedia}
}

@incollection{app4res_latzel2024artificial,
  title={Artificial Intelligence in Sport Scientific Creation and Writing Process},
  author={Latzel, Richard and Glauner, Patrick},
  booktitle={Artificial Intelligence in Sports, Movement, and Health},
  pages={15--29},
  year={2024},
  publisher={Springer}
}

@article{app4res_hakam2024human,
  title={Human-written vs AI-generated texts in orthopedic academic literature: comparative qualitative analysis},
  author={Hakam, Hassan Tarek and Prill, Robert and Korte, Lisa and Lovrekovi{\'c}, Bruno and Ostoji{\'c}, Marko and Ramadanov, Nikolai and Muehlensiepen, Felix},
  journal={JMIR formative research},
  volume={8},
  pages={e52164},
  year={2024},
  publisher={JMIR Publications Toronto, Canada}
}

@article{app4ind_man_haghparast2025foresight,
  title={Foresight in Sports Businesses: Exploring Emerging Scenarios Based on AI-Language Models and Financial Management Strategies},
  author={Haghparast, Mehran and Soltan Hoseini, Mohammad and Nasr Esfahani, Davood},
  journal={Sports Business Journal},
  year={2025},
  publisher={Alzahra University}
}

@article{app4ind_man_haghparast2024financial,
  title={A Financial Management Maturity Model in Sports Organizations: A Novel Approach Using Artificial Intelligence},
  author={Haghparast, Mehran and Hoseini, Mohamad Soltan and Esfahani, Davood Nasr},
  journal={Journal of New Studies in Sport Management},
  year={2024}
}

@article{app4ind_man_merilehto2024pdfs,
  title={From PDFs to Structured Data: Utilizing LLM Analysis in Sports Database Management},
  author={Merilehto, Juhani},
  journal={arXiv preprint arXiv:2410.17619},
  year={2024}
}

@article{app4ind_man_salimi2025comprehensive,
  title={Comprehensive Site Selection Model for Sports Facilities in Iran: Leveraging AI Language Models},
  author={Salimi Beni, Elham and Mostahfezian, Mina and Khorvash, Majid and Nasr Esfahani, Davood},
  journal={Sport Management Journal},
  year={2025},
  publisher={University of Tehran}
}

@article{app4ind_tou_yenisoy2025investigating,
  title={Investigating esports tourism research using artificial intelligence applications: ChatGPT versus ZekAI},
  author={Yenisoy, Caner and Silik, Cemal Ersin},
  journal={Tourism and Recreation},
  volume={7},
  number={1},
  pages={54--68},
  year={2025},
  publisher={Turizm Rekreasyon ve Gastronomi Ara{\c{s}}t{\i}rmalar{\i} Derne{\u{g}}i (TURGADER)}
}

@article{app4ind_tou_memon2025ai,
  title={AI-Powered ChatGPT in Sports Tourism: Benefits, Challenges, and Future Prospects},
  author={Memon, Salman Bashir and Qureshi, Jawaid Ahmed and Shah, Samar Batool},
  journal={Redefining Tourism With AI and the Metaverse},
  pages={163--188},
  year={2025},
  publisher={IGI Global Scientific Publishing}
}

@inproceedings{app4ind_tal_martire2025leveraging,
  title={Leveraging LLMs and RAG for Enhanced Football Talent Scouting},
  author={Martire, Felice Antonio and Ragazzi, Davide},
  booktitle={International Conference on Advanced Information Systems Engineering},
  pages={298--309},
  year={2025},
  organization={Springer}
}

@article{app4ind_tal_raskar2025footyintel,
  title={Footyintel: Creating An AI Scout For Better Talent Recognition},
  author={Raskar, Sandeep and Thosar, Manas and Dandge, Atharva and Fale, Pranav},
  journal={International Journal of Environmental Sciences},
  pages={99--106},
  year={2025}
}

@article{app4ind_tal_mateus2024empowering,
  title={Empowering the sports scientist with artificial intelligence in training, performance, and health management},
  author={Mateus, Nuno and Abade, Eduardo and Coutinho, Diogo and G{\'o}mez, Miguel-{\'A}ngel and Pe{\~n}as, Carlos Lago and Sampaio, Jaime},
  journal={Sensors},
  volume={25},
  number={1},
  pages={139},
  year={2024},
  publisher={MDPI}
}

@inproceedings{app4fans_com_you2025timesoccer,
  title={Timesoccer: An end-to-end multimodal large language model for soccer commentary generation},
  author={You, Ling and Huang, Wenxuan and Xie, Xinni and Wei, Xiangyi and Li, Bangyan and Lin, Shaohui and Li, Yang and Wang, Changbo},
  booktitle={Proceedings of the 33rd ACM International Conference on Multimedia},
  pages={3418--3427},
  year={2025}
}

@inproceedings{app4fans_com_rao2024matchtime,
  title={Matchtime: Towards automatic soccer game commentary generation},
  author={Rao, Jiayuan and Wu, Haoning and Liu, Chang and Wang, Yanfeng and Xie, Weidi},
  booktitle={Proceedings of the 2024 Conference on Empirical Methods in Natural Language Processing},
  pages={1671--1685},
  year={2024}
}

@inproceedings{app4fans_com_wang2024commentary,
  title={Commentary Generation from Data Records of Multiplayer Strategy Esports Game},
  author={Wang, Zihan and Yoshinaga, Naoki},
  booktitle={Proceedings of the 2024 Conference of the North American Chapter of the Association for Computational Linguistics: Human Language Technologies (Volume 4: Student Research Workshop)},
  pages={263--271},
  year={2024}
}

@inproceedings{app4fans_com_lin2024personalized,
  title={Personalized Video Comment Generation},
  author={Lin, Xudong and Zare, Ali and Huang, Shiyuan and Yang, Ming-Hsuan and Chang, Shih-Fu and Zhang, Li},
  booktitle={Findings of the Association for Computational Linguistics: EMNLP 2024},
  pages={16806--16820},
  year={2024}
}

@inproceedings{app4fans_com_jiang2025domain,
  title={Domain adaptation of VLM for soccer video understanding},
  author={Jiang, Tiancheng and Wang, Henry and Salekin, Md Sirajus and Atighehchian, Parmida and Zhang, Shinan},
  booktitle={Proceedings of the Computer Vision and Pattern Recognition Conference},
  pages={6111--6121},
  year={2025}
}

@inproceedings{app4fans_com_mori2025live,
  title={Live Football Commentary System Providing Background Information},
  author={Mori, Yuichiro and Tanaka, Chikara and Maekawa, Aru and Kosugi, Satoshi and Ishigaki, Tatsuya and Funakoshi, Kotaro and Takamura, Hiroya and Okumura, Manabu},
  booktitle={Proceedings of the 63rd Annual Meeting of the Association for Computational Linguistics (Volume 3: System Demonstrations)},
  pages={394--404},
  year={2025}
}

@article{app4fans_com_vijayakumar2025player,
  title={Player Tracking-Integrated Soccer Game Commentary Generation},
  author={Vijayakumar, Adarsh and Toms, Amal and Vadivu, Senthil},
  journal={IJSAT-International Journal on Science and Technology},
  volume={16},
  number={2},
  year={2025},
  publisher={International Research Publication and Journals}
}

@inproceedings{app4fans_com_andrews2024aicommentator,
  title={AiCommentator: A multimodal conversational agent for embedded visualization in football viewing},
  author={Andrews, Peter and Nordberg, Oda Elise and Zubicueta Portales, Stephanie and Borch, Nj{\aa}l and Guribye, Frode and Fujita, Kazuyuki and Fjeld, Morten},
  booktitle={Proceedings of the 29th International Conference on Intelligent User Interfaces},
  pages={14--34},
  year={2024}
}

@article{app4fans_com_ge2024scbench,
  title={SCBench: A Sports Commentary Benchmark for Video LLMs},
  author={Ge, Kuangzhi and Chen, Lingjun and Zhang, Kevin and Luo, Yulin and Shi, Tianyu and Fan, Liaoyuan and Li, Xiang and Wang, Guanqun and Zhang, Shanghang},
  journal={arXiv preprint arXiv:2412.17637},
  year={2024}
}

@inproceedings{app4fans_com_andrews2024designing,
  title={Designing for automated sports commentary systems},
  author={Andrews, Peter and Nordberg, Oda Elise and Borch, Nj{\aa}l and Guribye, Frode and Fjeld, Morten},
  booktitle={Proceedings of the 2024 ACM International Conference on Interactive Media Experiences},
  pages={75--93},
  year={2024}
}

@inproceedings{app4fans_com_gautam2022soccer,
  title={Soccer game summarization using audio commentary, metadata, and captions},
  author={Gautam, Sushant and Midoglu, Cise and Shafiee Sabet, Saeed and Kshatri, Dinesh Baniya and Halvorsen, P{\aa}l},
  booktitle={Proceedings of the 1st Workshop on User-centric Narrative Summarization of Long Videos},
  pages={13--22},
  year={2022}
}

@inproceedings{app4fans_com_chen2025livecc,
  title={Livecc: Learning video llm with streaming speech transcription at scale},
  author={Chen, Joya and Zeng, Ziyun and Lin, Yiqi and Li, Wei and Ma, Zejun and Shou, Mike Zheng},
  booktitle={Proceedings of the Computer Vision and Pattern Recognition Conference},
  pages={29083--29095},
  year={2025}
}

@inproceedings{app4fans_com_li2025multi,
  title={Multi-Modal Large Language Model with RAG Strategies in Soccer Commentary Generation},
  author={Li, Xiang and He, Yangfan and Zu, Shuaishuai and Li, Zhengyang and Shi, Tianyu and Xie, Yiting and Zhang, Kevin},
  booktitle={2025 IEEE/CVF Winter Conference on Applications of Computer Vision (WACV)},
  pages={6197--6206},
  year={2025},
  organization={IEEE}
}

@article{app4fans_com_cook2024llm,
  title={LLM-Commentator: Novel fine-tuning strategies of large language models for automatic commentary generation using football event data},
  author={Cook, Alec and Karaku{\c{s}}, Oktay},
  journal={Knowledge-Based Systems},
  volume={300},
  pages={112219},
  year={2024},
  publisher={Elsevier}
}

@inproceedings{app4fans_com_xi2025player,
  title={Player-centric multimodal prompt generation for large language model based identity-aware basketball video captioning},
  author={Xi, Zeyu and Sun, Haoying and Wu, Yaofei and Yan, Junchi and Zhang, Haoran and Wu, Lifang and Wang, Liang and Chen, Changwen},
  booktitle={Proceedings of the IEEE/CVF International Conference on Computer Vision},
  pages={24330--24339},
  year={2025}
}

@inproceedings{app4fans_com_baughman2024large,
  title={Large scale generative AI text applied to sports and music},
  author={Baughman, Aaron and Morales, Eduardo and Agarwal, Rahul and Akay, Gozde and Feris, Rogerio and Johnson, Tony and Hammer, Stephen and Karlinsky, Leonid},
  booktitle={Proceedings of the 30th ACM SIGKDD Conference on Knowledge Discovery and Data Mining},
  pages={4784--4792},
  year={2024}
}

@inproceedings{app4fans_com_pavlovich2023soccer,
  title={Soccer Artificial Intelligence Commentary Service on the Base of Video Analytic and Large Language Models},
  author={Pavlovich, Roman V and Tsybulko, Evgeniya A and Zhigunov, Konstantin N and Khelvas, Aleksandr V and Gilya-Zetinov, Aleksandr A and Tykhonov, Illya V},
  booktitle={2023 31st Telecommunications Forum (TELFOR)},
  pages={1--4},
  year={2023},
  organization={IEEE}
}

@inproceedings{app4fans_com_sameer2025enhanced,
  title={Enhanced Cricket Commentary Using AI Vision and Multilingual Translation},
  author={Sameer, N Md and Jayavardhan, K and Iyyappan, Oviya Ramalakshmi},
  booktitle={2025 IEEE International Conference on Emerging Technologies and Applications (MPSec ICETA)},
  pages={1--6},
  year={2025},
  organization={IEEE}
}

@inproceedings{app4fans_com_mkhallati2023soccernet,
  title={SoccerNet-caption: Dense video captioning for soccer broadcasts commentaries},
  author={Mkhallati, Hassan and Cioppa, Anthony and Giancola, Silvio and Ghanem, Bernard and Van Droogenbroeck, Marc},
  booktitle={Proceedings of the IEEE/CVF Conference on Computer Vision and Pattern Recognition},
  pages={5074--5085},
  year={2023}
}

@inproceedings{app4fans_com_sarkhoosh2024soccersum,
  title={The SoccerSum Dataset for Automated Detection, Segmentation, and Tracking of Objects on the Soccer Pitch},
  author={Sarkhoosh, Mehdi Houshmand and Gautam, Sushant and Midoglu, Cise and Sabet, Saeed Shafiee and Torjusen, Thomas and Halvorsen, P{\aa}l},
  booktitle={Proceedings of the 15th ACM Multimedia Systems Conference},
  pages={353--359},
  year={2024}
}

@inproceedings{app4fans_com_zhang2024descriptive,
  title={A descriptive basketball highlight dataset for automatic commentary generation},
  author={Zhang, Benhui and Gao, Junyu and Yuan, Yuan},
  booktitle={Proceedings of the 32nd ACM international conference on multimedia},
  pages={10316--10325},
  year={2024}
}

@article{app4fans_hig_banu2025survey,
  title={Survey Paper On AI Based Sports Highlight Generation For Social Media},
  author={Banu, Sameena and others},
  journal={Journal of Scientific Research and Technology},
  pages={30--38},
  year={2025}
}

@inproceedings{app4fans_hig_midoglu2024ai,
  title={Ai-based sports highlight generation for social media},
  author={Midoglu, Cise and Sabet, Saeed Shafiee and Sarkhoosh, Mehdi Houshmand and Majidi, Mohammad and Gautam, Sushant and Solberg, H{\aa}kon Maric and Kupka, Tomas and Halvorsen, P{\aa}l},
  booktitle={Proceedings of the 3rd Mile-High Video Conference},
  pages={7--13},
  year={2024}
}

@inproceedings{app4fans_hig_sattar2023multi,
  title={Multi-Modal Architecture for Cricket Highlights Generation: Using Computer Vision and Large Language Model},
  author={Sattar, Husnain and Umar, Muhammad Shamil and Ijaz, Eeman and Arshad, Muhammad Umair},
  booktitle={2023 17th International Conference on Open Source Systems and Technologies (ICOSST)},
  pages={1--6},
  year={2023},
  organization={IEEE}
}

@inproceedings{app4fans_hig_kang2025diamond,
  title={DIAMOND: An LLM-Driven Agent for Context-Aware Baseball Highlight Summarization},
  author={Kang, Jeonghun and Kwon, Soonmok and Lee, Joonseok and Kim, Byung-Hak},
  booktitle={Proceedings of the 1st Workshop for Research on Agent Language Models (REALM 2025)},
  pages={386--400},
  year={2025}
}

@inproceedings{app4fans_hig_leehippo,
  title={HIPPO-VIDEO: Simulating Watch Histories with Large Language Models for History-Driven Video Highlighting},
  author={Lee, Jeongeun and Yu, Youngjae and Lee, Dongha},
  booktitle={Second Conference on Language Modeling},
  year={2025}
}

@article{app4fans_hig_davids2025sportsummarizer,
  title={SportSummarizer: A Unified Multimodal Fusion Transformer for Context-Aware Sports Video Summarization},
  author={Davids, D Minola and Raj, A Arul Edwin and Christopher, C Seldev},
  journal={Neurocomputing},
  pages={131011},
  year={2025},
  publisher={Elsevier}
}

@inproceedings{app4fans_new_wang2022knowledge,
  title={Knowledge enhanced sports game summarization},
  author={Wang, Jiaan and Li, Zhixu and Zhang, Tingyi and Zheng, Duo and Qu, Jianfeng and Liu, An and Zhao, Lei and Chen, Zhigang},
  booktitle={Proceedings of the Fifteenth ACM International Conference on Web Search and Data Mining},
  pages={1045--1053},
  year={2022}
}

@article{app4fans_new_cheng2024snil,
  title={SNIL: generating sports news from insights with large language models},
  author={Cheng, Liqi and Deng, Dazhen and Xie, Xiao and Qiu, Rihong and Xu, Mingliang and Wu, Yingcai},
  journal={IEEE Transactions on Visualization and Computer Graphics},
  year={2024},
  publisher={IEEE}
}

@inproceedings{app4fans_new_sarkar2024advancing,
  title={Advancing cricket narratives: AI-enhanced advanced journaling in the IPL using language models},
  author={Sarkar, Soham and Yashwanth, Tadisetty Sai and Giri, Animesh},
  booktitle={2024 IEEE International Conference on Electronics, Computing and Communication Technologies (CONECCT)},
  pages={1--6},
  year={2024},
  organization={IEEE}
}

@inproceedings{app4fans_new_chiang2025tree,
  title={Tree-of-Report: Table-to-Text Generation for Sports Game Reports with Tree-Structured Prompting},
  author={Chiang, Shang-Hsuan and Yang, Tsan-Tsung and Wang, Kuang-Da and Wang, Wei-Yao and Yen, An-Zi and Peng, Wen-Chih},
  booktitle={ACL 2025 Student Research Workshop},
year={2025}
}

@article{app4fans_new_chiang2024badge,
  title={BADGE: BADminton report Generation and Evaluation with LLM},
  author={Chiang, Shang-Hsuan and Chao, Lin-Wei and Wang, Kuang-Da and Wang, Chih-Chuan and Peng, Wen-Chih},
  journal={arXiv preprint arXiv:2406.18116},
  year={2024}
}

@inproceedings{app4fans_new_wang2021sportssum2,
  title={Sportssum2. 0: Generating high-quality sports news from live text commentary},
  author={Wang, Jiaan and Li, Zhixu and Yang, Qiang and Qu, Jianfeng and Chen, Zhigang and Liu, Qingsheng and Hu, Guoping},
  booktitle={Proceedings of the 30th ACM International Conference on Information \& Knowledge Management},
  pages={3463--3467},
  year={2021}
}

@article{app4fans_nar_lee2024sportify,
  title={Sportify: question answering with embedded visualizations and personified narratives for sports video},
  author={Lee, Chunggi and Lin, Tica and Pfister, Hanspeter and Zhu-Tian, Chen},
  journal={IEEE Transactions on Visualization and Computer Graphics},
  year={2024},
  publisher={IEEE}
}

@inproceedings{app4fans_nar_sarfati2023generating,
  title={Generating factually consistent sport highlights narrations},
  author={Sarfati, Noah and Yerushalmy, Ido and Chertok, Michael and Keller, Yosi},
  booktitle={Proceedings of the 6th International Workshop on Multimedia Content Analysis in Sports},
  pages={15--22},
  year={2023}
}

@inproceedings{app4fans_nar_lin2025sportsbuddy,
  title={SportsBuddy: Designing and Evaluating an AI-Powered Sports Video Storytelling Tool Through Real-World Deployment},
  author={Lin, Tica and Xiang, Ruxun and Liu, Gardenia and Tiwari, Divyanshu and Chiang, Meng-Chia and Ye, Chenjiayi and Pfister, Hanspeter and Zhu-Tian, Chen},
  booktitle={2025 IEEE 18th Pacific Visualization Conference (PacificVis)},
  pages={214--223},
  year={2025},
  organization={IEEE}
}

@inproceedings{app4fans_nar_sarkhoosh2024multimodal,
  title={Multimodal AI-based summarization and storytelling for soccer on social media},
  author={Sarkhoosh, Mehdi Houshmand and Gautam, Sushant and Midoglu, Cise and Sabet, Saeed Shafiee and Halvorsen, P{\aa}l},
  booktitle={Proceedings of the 15th ACM multimedia systems conference},
  pages={485--491},
  year={2024}
}

@inproceedings{app4fans_opi_qian2024esports,
  title={Esports' Debut as a Medal Event at 2023 Asian Games: Exploring Public Perceptions with BERTopic and GPT-4 Topic Fine-Tuning},
  author={Qian, Tyreal Yizhou and Yu, Bo and Li, Weizhe and Xu, Chenglong},
  booktitle={58th Hawaii International Conference on System Sciences, HICSS 2025},
  pages={4303--4312},
  year={2025},
  organization={IEEE Computer Society}
}

@article{app4fans_opi_argan23investigating,
  title={INVESTIGATING THE FACTORS INFLUENCING ADOPTION INTENTIONS OF CHATGPT FOR SPORT EVENTS},
  author={Argan, Metin and Din{\c{c}}, Halime},
  journal={SPORMETRE Beden E{\u{g}}itimi ve Spor Bilimleri Dergisi},
  volume={23},
  number={2},
  pages={77--97},
  publisher={Ankara University},
  year={2025},
}

@article{app4fans_opi_qian2025experience,
  title={Experience is all you need: a large language model application of fine-tuned GPT-3.5 and RoBERTa for aspect-based sentiment analysis of college football stadium reviews},
  author={Qian, Tyreal Yizhou and Li, Weizhe and Gong, Hua and Seifried, Chad and Xu, Chenglong},
  journal={Sport Management Review},
  volume={28},
  number={1},
  pages={1--25},
  year={2025},
  publisher={Taylor \& Francis}
}

@inproceedings{app4fans_opi_rauchegger2024onelove,
  title={OneLove beyond the field-A few-shot pipeline for topic and sentiment analysis during the FIFA World Cup in Qatar},
  author={Rauchegger, Christoph and Wang, Sonja Mei and Delobelle, Pieter},
  booktitle={Proceedings of the 20th Conference on Natural Language Processing (KONVENS 2024)},
  pages={349--357},
  year={2024}
}

@inproceedings{app4fans_mod_priya2024megan,
  title={Megan-A Sports Chatbot using OpenAI APIs and Django Framework with Python},
  author={Priya, M Yagnasri and Kamble, Samrudhi S and Shendre, Sanjivani P and Sridhar, Swetha and others},
  booktitle={2024 IEEE 9th International Conference for Convergence in Technology (I2CT)},
  pages={1--8},
  year={2024},
  organization={IEEE}
}

@inproceedings{app4fans_mod_kim2025bleacherbot,
  title={BleacherBot: AI Agent as a Sports Co-Viewing Partner},
  author={Kim, Kyusik and Song, Hyungwoo and Ryu, Jeongwoo and Oh, Changhoon and Suh, Bongwon},
  booktitle={Proceedings of the 2025 CHI Conference on Human Factors in Computing Systems},
  pages={1--31},
  year={2025}
}

@inproceedings{app4fans_mod_karat2025system,
  title={A System for Triggering Sports Instant Answers on Search Engines},
  author={Karat, Ankith and Tibrewal, Atishay and Kotian, Nishka and Dang, Manan and Valluri, Ravindra and Ravi Teja Marineni, Antony and Sahni, Sarthak and Sundaresan, Rhea and Kumar, Ankit and Mehndiratta, Aditya and others},
  booktitle={Proceedings of the 48th International ACM SIGIR Conference on Research and Development in Information Retrieval},
  pages={4304--4308},
  year={2025}
}

@inproceedings{app4fans_mod_schilling2024querying,
  title={Querying Football Matches for Event Data: Towards Using Large Language Models},
  author={Schilling, Alexander and Anurathan, James and M{\"u}hlberger, Johannes and Gerschner, Felix and R{\"o}ssle, Manfred and Theissler, Andreas and Klaiber, Marco},
  booktitle={International Sports Analytics Conference and Exhibition},
  pages={216--227},
  year={2024},
  organization={Springer}
}

@article{app4fans_mod_song2025korean,
  title={Korean football in-game conversation state tracking dataset for dialogue and turn level evaluation},
  author={Song, Sangmin and Park, Juhyoung and Choi, Juhwan and Lee, Junho and Jin, Kyohoon and Kim, YoungBin},
  journal={Engineering Applications of Artificial Intelligence},
  volume={139},
  pages={109572},
  year={2025},
  publisher={Elsevier}
}

@inproceedings{app4fans_mod_strand2024soccer,
  title={Soccer Information Retrieval via Natural Queries using SoccerRAG},
  author={Strand, Aleksander Theo and Gautam, Sushant and Midoglu, Cise and Halvorsen, P{\aa}l},
  booktitle={2024 International Conference on Content-Based Multimedia Indexing (CBMI)},
  pages={1--5},
  year={2024},
  organization={IEEE}
}

@inproceedings{app4fans_mod_strand2024soccerrag,
  title={Soccerrag: Multimodal soccer information retrieval via natural queries},
  author={Strand, Aleksander Theo and Gautam, Sushant and Midoglu, Cise and Halvorsen, P{\aa}l},
  booktitle={2024 International Conference on Content-Based Multimedia Indexing (CBMI)},
  pages={1--7},
  year={2024},
  organization={IEEE}
}

@article{app4fans_mod_sepasdar2024enhancing,
  title={Enhancing structured-data retrieval with graphrag: Soccer data case study},
  author={Sepasdar, Zahra and Gautam, Sushant and Midoglu, Cise and Riegler, Michael A and Halvorsen, P{\aa}l},
  journal={arXiv preprint arXiv:2409.17580},
  year={2024}
}

@inproceedings{app4fans_mod_sepasdar2024soccer,
  title={Soccer-graphrag: Applications of graphrag in soccer},
  author={Sepasdar, Zahra and Gautam, Sushant and Midoglu, Cise and Riegler, Michael A and Halvorsen, P{\aa}l},
  booktitle={International Workshop on Graph-Based Approaches in Information Retrieval},
  pages={1--10},
  year={2024},
  organization={Springer}
}

@article{app4fans_mod_wickramasinghe2025assessing,
  title={Assessing the accuracy of large language models in extracting latest cricket information},
  author={Wickramasinghe, Indika},
  journal={Scientific Journal of Sport and Performance},
  volume={4},
  number={2},
  pages={268--284},
  year={2025}
}

@article{app4fans_mod_wang2025agentic,
  title={Agentic generative AI for media content discovery at the national football league},
  author={Wang, Henry and Salekin, Sirajus and Lee, Jake and Claytor, Ross and Zhang, Shinan and Chi, Michael},
  year={2025}
}

@inproceedings{app4fans_mod_gautam2025soccerchat,
  title={Soccerchat: Integrating multimodal data for enhanced soccer game understanding},
  author={Gautam, Sushant and Midoglu, Cise and Thambawita, Vajira L and Riegler, Michael A and Halvorsen, Pal and Shah, Mubarak},
  booktitle={2025 International Conference on Content-Based Multimedia Indexing (CBMI)},
  pages={1--8},
  year={2025},
  organization={IEEE}
}

@inproceedings{app4fans_mod_rao2025multi,
  title={Multi-agent system for comprehensive soccer understanding},
  author={Rao, Jiayuan and Li, Zifeng and Wu, Haoning and Zhang, Ya and Wang, Yanfeng and Xie, Weidi},
  booktitle={Proceedings of the 33rd ACM International Conference on Multimedia},
  pages={3654--3663},
  year={2025}
}

@inproceedings{app4fans_mod_rao2025towards,
  title={Towards universal soccer video understanding},
  author={Rao, Jiayuan and Wu, Haoning and Jiang, Hao and Zhang, Ya and Wang, Yanfeng and Xie, Weidi},
  booktitle={Proceedings of the Computer Vision and Pattern Recognition Conference},
  pages={8384--8394},
  year={2025}
}

@article{app4fans_mod_unlu2023footgpt,
  title={FootGPT: A Large Language Model Development Experiment on a Minimal Setting},
  author={Unlu, Eren},
  journal={arXiv preprint arXiv:2308.08610},
  year={2023}
}

@inproceedings{app4fans_mod_gupta2025play,
  title={From Play to Replay: Composed Video Retrieval for Temporally Fine-Grained Videos},
  author={Gupta, Animesh and Parmar, Jay and Dave, Ishan Rajendrakumar and Shah, Mubarak},
  booktitle={The Thirty-ninth Annual Conference on Neural Information Processing Systems Datasets and Benchmarks Track},
  year={2025}
}

@article{app4coa_ana_kodathala2025sv3,
  title={SV3. 3B: A Sports Video Understanding Model for Action Recognition},
  author={Kodathala, Sai Varun and Vutukoori, Yashwanth Reddy and Vunnam, Rakesh},
  journal={arXiv preprint arXiv:2507.17844},
  year={2025}
}

@article{app4coa_ana_teo2025enhancing,
  title={Enhancing Sports Strategy with Video Analytics and Data Mining: Assessing the effectiveness of Multimodal LLMs in tennis video analysis},
  author={Teo, Charlton},
  journal={arXiv preprint arXiv:2507.02904},
  year={2025}
}

@article{app4coa_ana_shin2025soccer,
  title={Soccer-CLIP: Vision Language Model for Soccer Action Spotting},
  author={Shin, Yoonho and Park, Sanghoon and Han, Youngsub and Jeon, Byoung-Ki and Lee, Soonyoung and Kang, Byung Jun},
  journal={IEEE Access},
  volume={13},
  pages={44354--44365},
  year={2025},
  publisher={IEEE}
}

@inproceedings{app4coa_ana_nonaka2024rugby,
  title={Rugby scene classification enhanced by vision language model},
  author={Nonaka, Naoki and Fujihira, Ryo and Koshiba, Toshiki and Maeda, Akira and Seita, Jun},
  booktitle={Proceedings of the IEEE/CVF Conference on Computer Vision and Pattern Recognition},
  pages={3256--3266},
  year={2024}
}

@article{app4coa_ana_salehi2024actionatlas,
  title={Actionatlas: A videoqa benchmark for domain-specialized action recognition},
  author={Salehi, Mohammadreza Reza and Park, Jae Sung and Kusupati, Aditya and Krishna, Ranjay and Choi, Yejin and Hajishirzi, Hanna and Farhadi, Ali},
  journal={Advances in Neural Information Processing Systems},
  volume={37},
  pages={137372--137402},
  year={2024}
}

@inproceedings{app4coa_ana_liu2025f,
  title={{F}$^{3}$ Set: Towards Analyzing Fast, Frequent, and Fine-grained Events from Videos},
  author={Liu, Zhaoyu and Jiang, Kan and Ma, Murong and Hou, Zhe and Lin, Yun and Dong, Jin Song},
  booktitle={The Thirteenth International Conference on Learning Representations},
  year={2025}
}

@inproceedings{app4coa_ana_li2025improving,
  title={Improving LLM Video Understanding with 16 Frames Per Second},
  author={Li, Yixuan and Tang, Changli and Zhuang, Jimin and Yang, Yudong and Sun, Guangzhi and Li, Wei and MA, Zejun and Zhang, Chao},
  booktitle={Forty-second International Conference on Machine Learning},
  year={2025}
}

@inproceedings{app4coa_ana_chakraborty2025we,
  title={Do We Need Large VLMs for Spotting Soccer Actions?},
  author={Chakraborty, Ritabrata and Chakraborty, Rajatsubhra and Dasgupta, Avijit and Chaurasia, Sandeep},
  booktitle={The 14th International Joint Conference on Natural Language Processing and The 4th Conference of the Asia-Pacific Chapter of the Association for Computational Linguistics},
  pages={59--65},
  year={2025}
}

@inproceedings{app4coa_tac_caron2023tacticalgpt,
  title={TacticalGPT: uncovering the potential of LLMs for predicting tactical decisions in professional football},
  author={Caron, Matthew and M{\"u}ller, Oliver},
  booktitle={StatsBomb Conference},
  pages={1--11},
  year={2023}
}

@article{app4coa_tac_liu2024smartboard,
  title={Smartboard: Visual Exploration of Team Tactics with LLM Agent},
  author={Liu, Ziao and Xie, Xiao and He, Moqi and Zhao, Wenshuo and Wu, Yihong and Cheng, Liqi and Zhang, Hui and Wu, Yingcai},
  journal={IEEE Transactions on Visualization and Computer Graphics},
  year={2024},
  publisher={IEEE}
}

@article{app4coa_tac_lingrui2025tacticexpert,
  title={TacticExpert: Spatial-Temporal Graph Language Model for Basketball Tactics},
  author={Lingrui, Xu and Mandi, Liu and Lei, Zhang},
  journal={arXiv preprint arXiv:2503.10722},
  year={2025}
}

@article{app4coa_tac_michielssen2024using,
  title={Using large language models to generate baseball spray charts in the absence of numerical data},
  author={Michielssen, Senne and Maloof, Adam and Haumacher, Joe and Dreger, Alexander and Bonicki, Kyle and Hallgren, Karl},
  journal={Proceedings of the Institution of Mechanical Engineers, Part P: Journal of Sports Engineering and Technology},
  pages={17543371241257734},
  year={2024},
  publisher={SAGE Publications Sage UK: London, England}
}

@article{app4coa_tac_zhang2025chatmatch,
  title={ChatMatch: Exploring the potential of hybrid vision--language deep learning approach for the intelligent analysis and inference of racket sports},
  author={Zhang, Jiawen and Han, Dongliang and Han, Shuai and Li, Heng and Lam, Wing-Kai and Zhang, Mingyu},
  journal={Computer Speech \& Language},
  volume={89},
  pages={101694},
  year={2025},
  publisher={Elsevier}
}

@article{app4coa_tac_hu2024can,
  title={Can large language models do analytical reasoning?},
  author={Hu, Yebowen and Song, Kaiqiang and Cho, Sangwoo and Wang, Xiaoyang and Foroosh, Hassan and Yu, Dong and Liu, Fei},
  journal={arXiv preprint arXiv:2403.04031},
  year={2024}
}

@inproceedings{app4coa_tac_janssens2024large,
  title={Large Language Models on Race Commentary: Towards Granular Data in Cycling Analytics},
  author={Janssens, Bram and Bogaert, Matthias and Verstockt, Steven},
  booktitle={International Workshop on Machine Learning and Data Mining for Sports Analytics},
  pages={14--25},
  year={2024},
  organization={Springer}
}

@inproceedings{app4fans_nar_hu2024reasoning,
  title={When reasoning meets information aggregation: A case study with sports narratives},
  author={Hu, Yebowen and Song, Kaiqiang and Cho, Sangwoo and Wang, Xiaoyang and Yao, Wenlin and Foroosh, Hassan and Yu, Dong and Liu, Fei},
  booktitle={Proceedings of the 2024 conference on empirical methods in natural language processing},
  pages={4293--4308},
  year={2024}
}

@inproceedings{app4coa_tac_hu-etal-2024-sportsmetrics,
    title = "{S}ports{M}etrics: Blending Text and Numerical Data to Understand Information Fusion in {LLM}s",
    author = "Hu, Yebowen  and
      Song, Kaiqiang  and
      Cho, Sangwoo  and
      Wang, Xiaoyang  and
      Foroosh, Hassan  and
      Yu, Dong  and
      Liu, Fei",
    editor = "Ku, Lun-Wei  and
      Martins, Andre  and
      Srikumar, Vivek",
    booktitle = "Proceedings of the 62nd Annual Meeting of the Association for Computational Linguistics (Volume 1: Long Papers)",
    month = aug,
    year = "2024",
    address = "Bangkok, Thailand",
    publisher = "Association for Computational Linguistics",
    url = "https://aclanthology.org/2024.acl-long.17/",
    doi = "10.18653/v1/2024.acl-long.17",
    pages = "267--278",
}

@inproceedings{app4coa_pre_ibh2024stroke,
  title={A stroke of genius: Predicting the next move in badminton},
  author={Ibh, Magnus and Gra{\ss}hof, Stella and Hansen, Dan Witzner},
  booktitle={Proceedings of the IEEE/CVF Conference on Computer Vision and Pattern Recognition},
  pages={3376--3385},
  year={2024}
}

@article{app4coa_pre_oved2020predicting,
  title={Predicting in-game actions from interviews of NBA players},
  author={Oved, Nadav and Feder, Amir and Reichart, Roi},
  journal={Computational Linguistics},
  volume={46},
  number={3},
  pages={667--712},
  year={2020},
  publisher={MIT Press One Rogers Street, Cambridge, MA 02142-1209, USA journals-info~…}
}

@article{app4coa_pre_felice2024ai,
  title={AI for Handball: predicting and explaining the 2024 Olympic Games tournament with Deep Learning and Large Language Models},
  author={Felice, Florian},
  journal={arXiv preprint arXiv:2407.15987},
  year={2024}
}

@article{app4coa_pre_sprint2024social,
  title={Social networks and large language models for Division I basketball game winner prediction},
  author={Sprint, Gina},
  journal={IEEE Access},
  volume={12},
  pages={84774--84784},
  year={2024},
  publisher={IEEE}
}

@article{app4coa_pre_bhatnagar2025analyzing,
  title={Analyzing key factors influencing IPL cricket scores using explainability and multimodal data},
  author={Bhatnagar, Mohit and Bhatnagar, Manya},
  journal={Journal of Quantitative Analysis in Sports},
  volume={21},
  number={3},
  pages={253--267},
  year={2025},
  publisher={De Gruyter}
}

@article{zhou2024comprehensive,
  title={A comprehensive survey of action quality assessment: Method and benchmark},
  author={Zhou, Kanglei and Cai, Ruizhi and Wang, Liyuan and Shum, Hubert PH and Liang, Xiaohui},
  journal={arXiv preprint arXiv:2412.11149},
  year={2024}
}

@inproceedings{app4coa_aqa_wang2025beats,
  title={From Beats to Scores: A Multi-Modal Framework for Comprehensive Figure Skating Assessment},
  author={Wang, Fengshun and Wang, Qiurui and Chen, Dan},
  booktitle={Proceedings of the Computer Vision and Pattern Recognition Conference},
  pages={5905--5914},
  year={2025}
}

@inproceedings{app4coa_aqa_dibenedetto2025fine,
  title={Fine-Tuning Large Multimodal Models for Fitness Action Quality Assessment},
  author={Dibenedetto, Gaetano and Musacchio, Elio and Polignano, Marco and Lops, Pasquale},
  booktitle={Adjunct Proceedings of the 33rd ACM Conference on User Modeling, Adaptation and Personalization},
  pages={39--44},
  year={2025}
}

@inproceedings{app4coa_aqa_tang2025fitnessagent,
  title={FitnessAgent: A Unified Agent Framework for Open-Set and Personalized Fitness Evaluation},
  author={Tang, Zhenhui and Li, Jiahao and Guo, Ping and Tian, Bowen and Xing, Qingjun and Xing, Xuyang and Wang, Peng},
  booktitle={2025 IEEE International Conference on Robotics and Automation (ICRA)},
  pages={12437--12444},
  year={2025},
  organization={IEEE}
}

@article{app4coa_aqa_xing2025llm,
  title={LLM-FMS: A fine-grained dataset for functional movement screen action quality assessment},
  author={Xing, Qingjun and Xing, Xuyang and Guo, Ping and Tang, Zhenhui and Shen, Yanfei},
  journal={PloS one},
  volume={20},
  number={3},
  pages={e0313707},
  year={2025},
  publisher={Public Library of Science San Francisco, CA USA}
}

@article{app4coa_edu_kauppinen2024proactive,
  title={Proactive Autonomous Assignments as Pedagogical Responses to the Rise of Artificial Intelligence Solutions in Sport Management Teaching Practice},
  author={Kauppinen, Antti},
  journal={Sport Management Education Journal},
  volume={19},
  number={1},
  pages={54--58},
  year={2024},
  publisher={Human Kinetics}
}

@article{app4coa_edu_keiper2023artificial,
  title={Artificial intelligence in sport management education: Playing the AI game with ChatGPT},
  author={Keiper, Margaret C and Fried, Gil and Lupinek, Joshua and Nordstrom, Heidi},
  journal={Journal of Hospitality, Leisure, Sport \& Tourism Education},
  volume={33},
  pages={100456},
  year={2023},
  publisher={Elsevier}
}

@article{app4coa_edu_gencc2023artificial,
  title={Artificial intelligence in physical education and sports: New horizons with ChatGPT},
  author={Gen{\c{c}}, Ne{\c{s}}e},
  journal={Akdeniz Spor Bilimleri Dergisi},
  volume={6},
  number={1-Cumhuriyet'in 100. Y{\i}l{\i} {\"O}zel Say{\i}s{\i}},
  pages={17--32},
  year={2023},
  publisher={Hasan {\c{S}}AHAN}
}

@misc{app4coa_edu_fazackerley2025harnessing,
  title={Harnessing generative AI in exercise and sports science education: enhancing real-world learning and overcoming traditional barriers in data analysis},
  author={Fazackerley, Lewis A and Perrin, Dimitri and Minett, Geoffrey M},
  journal={Advances in Physiology Education},
  volume={49},
  number={2},
  pages={496--502},
  year={2025},
  publisher={American Physiological Society Rockville, MD}
}

@article{app4coa_edu_cui2025innovating,
  title={Innovating physical education with artificial intelligence: a potential approach},
  author={Cui, Bin and Jiao, Wei and Gui, Shuying and Li, Yang and Fang, Qun},
  journal={Frontiers in Psychology},
  volume={16},
  pages={1490966},
  year={2025},
  publisher={Frontiers Media SA}
}

@article{app4coa_edu_chang2025exploring,
  title={Exploring opportunities and challenges toward ChatGPT for inclusion in sport education},
  author={Chang, Shu-Hao and Chen, Su-Yen and Chang, Chin-Han},
  journal={Journal of Hospitality, Leisure, Sport \& Tourism Education},
  volume={37},
  pages={100572},
  year={2025},
  publisher={Elsevier}
}

@article{app4coa_edu_zhang2024using,
  title={Using ChatGPT to promote college students’ participation in physical activities and its effect on mental health},
  author={Zhang, Yi-Fan and Liu, Xin-Qiao},
  journal={World Journal of Psychiatry},
  volume={14},
  number={2},
  pages={330},
  year={2024}
}

@article{app4coa_edu_gao2025motion,
  title={From motion signals to insights: A unified framework for student behavior analysis and feedback in physical education classes},
  author={Gao, Xian and Ruan, Jiacheng and Gao, Jingsheng and Xie, Mingye and Zhang, Zongyun and Liu, Ting and Fu, Yuzhuo},
  journal={arXiv preprint arXiv:2503.06525},
  year={2025}
}

@inproceedings{app4ref_ref_app4held2024x,
  title={X-vars: Introducing explainability in football refereeing with multi-modal large language models},
  author={Held, Jan and Itani, Hani and Cioppa, Anthony and Giancola, Silvio and Ghanem, Bernard and Van Droogenbroeck, Marc},
  booktitle={Proceedings of the IEEE/CVF Conference on Computer Vision and Pattern Recognition},
  pages={3267--3279},
  year={2024}
}

@inproceedings{app4ref_ref_held2025enhancing,
  title={Enhancing Football Refereeing with $\{$AI$\}$:$\{$VARS$\}$ and $\{$X-VARS$\}$ for Assisted Decision-Making},
  author={Held, Jan and Cioppa, Anthony and Giancola, Silvio and Almahmoud, Elaf and Collins, Katherine M and Bhatt, Umang and Ghanem, Bernard and Van Droogenbroeck, Marc},
  booktitle={MathSport Conference},
  year={2025}
}

@inproceedings{gao2025fsbench,
  title={Fsbench: A figure skating benchmark for advancing artistic sports understanding},
  author={Gao, Rong and Liu, Xin and Hu, Zhuozhao and Xing, Bohao and Xia, Baiqiang and Yu, Zitong and K{\"a}lvi{\"a}inen, Heikki},
  booktitle={Proceedings of the Computer Vision and Pattern Recognition Conference},
  pages={13595--13605},
  year={2025}
}

@inproceedings{xiasportu,
  title={SPORTU: A Comprehensive Sports Understanding Benchmark for Multimodal Large Language Models},
  author={Xia, Haotian and Yang, Zhengbang and Zou, Junbo and Tracy, Rhys and Wang, Yuqing and Lu, Chi and Lai, Christopher and He, Yanjun and Shao, Xun and Xie, Zhuoqing and others},
  booktitle={The Thirteenth International Conference on Learning Representations},
year={2025}
}

@inproceedings{xia2024sportqa,
  title={SportQA: A Benchmark for Sports Understanding in Large Language Models},
  author={Xia, Haotian and Yang, Zhengbang and Wang, Yuqing and Tracy, Rhys and Zhao, Yun and Huang, Dongdong and Chen, Zezhi and Zhu, Yan and Wang, Yuan-Fang and Shen, Weining},
  booktitle={Proceedings of the 2024 Conference of the North American Chapter of the Association for Computational Linguistics: Human Language Technologies (Volume 1: Long Papers)},
  pages={5061--5081},
  year={2024}
}

@inproceedings{jardim2023qasports,
  title={Qasports: A question answering dataset about sports},
  author={Jardim, Pedro Calciolari and Moraes, Leonardo Mauro Pereira and Aguiar, Cristina Dutra},
  booktitle={Dataset Showcase Workshop (DSW)},
  pages={1--12},
  year={2023},
  organization={SBC}
}

@inproceedings{he2025finebadminton,
  title={Finebadminton: A multi-level dataset for fine-grained badminton video understanding},
  author={He, Xusheng and Liu, Wei and Ma, Shanshan and Liu, Qian and Ma, Chenghao and Wu, Jianlong},
  booktitle={Proceedings of the 33rd ACM International Conference on Multimedia},
  pages={12776--12783},
  year={2025}
}

@inproceedings{chen2025finequest,
  title={FineQuest: Adaptive Knowledge-Assisted Sports Video Understanding via Agent-of-Thoughts Reasoning},
  author={Chen, Haodong and Huang, Haojian and Yin, Xinxiang and Shao, Dian},
  booktitle={Proceedings of the 33rd ACM International Conference on Multimedia},
  pages={2909--2918},
  year={2025}
}

@article{li2024sports_qa,
  title={Sports-qa: A large-scale video question answering benchmark for complex and professional sports},
  author={Li, Haopeng and Deng, Andong and Liu, Jun and Rahmani, Hossein and Guo, Yulan and Schiele, Bernt and Bennamoun, Mohammed and Ke, Qiuhong},
  journal={International Journal of Computer Vision},
  volume={134},
  number={5},
  pages={196},
  year={2026},
  publisher={Springer}
}

@article{srivastava2023bigbench,
  title={Beyond the imitation game: Quantifying and extrapolating the capabilities of language models},
  author={Srivastava, Aarohi and Rastogi, Abhinav and Rao, Abhishek and Shoeb, Abu Awal and Abid, Abubakar and Fisch, Adam and Brown, Adam R and Santoro, Adam and Gupta, Aditya and Garriga-Alonso, Adri and others},
  journal={Transactions on machine learning research},
  year={2023}
}

@inproceedings{shao2020finegym,
  title={Finegym: A hierarchical video dataset for fine-grained action understanding},
  author={Shao, Dian and Zhao, Yue and Dai, Bo and Lin, Dahua},
  booktitle={Proceedings of the IEEE/CVF conference on computer vision and pattern recognition},
  pages={2616--2625},
  year={2020}
}

@inproceedings{xu2022finediving,
  title={Finediving: A fine-grained dataset for procedure-aware action quality assessment},
  author={Xu, Jinglin and Rao, Yongming and Yu, Xumin and Chen, Guangyi and Zhou, Jie and Lu, Jiwen},
  booktitle={Proceedings of the IEEE/CVF conference on computer vision and pattern recognition},
  pages={2949--2958},
  year={2022}
}

@inproceedings{wang2024lvbench,
  title={Lvbench: An extreme long video understanding benchmark},
  author={Wang, Weihan and He, Zehai and Hong, Wenyi and Cheng, Yean and Zhang, Xiaohan and Qi, Ji and Ding, Ming and Gu, Xiaotao and Huang, Shiyu and Xu, Bin and others},
  booktitle={Proceedings of the IEEE/CVF International Conference on Computer Vision},
  pages={22958--22967},
  year={2025}
}

@inproceedings{fu2025videomme,
  title={Video-mme: The first-ever comprehensive evaluation benchmark of multi-modal llms in video analysis},
  author={Fu, Chaoyou and Dai, Yuhan and Luo, Yongdong and Li, Lei and Ren, Shuhuai and Zhang, Renrui and Wang, Zihan and Zhou, Chenyu and Shen, Yunhang and Zhang, Mengdan and others},
  booktitle={Proceedings of the Computer Vision and Pattern Recognition Conference},
  pages={24108--24118},
  year={2025}
}

@inproceedings{wanginternvid,
  title={InternVid: A Large-scale Video-Text Dataset for Multimodal Understanding and Generation},
  author={Wang, Yi and He, Yinan and Li, Yizhuo and Li, Kunchang and Yu, Jiashuo and Ma, Xin and Li, Xinhao and Chen, Guo and Chen, Xinyuan and Wang, Yaohui and others},
  booktitle={The Twelfth International Conference on Learning Representations},
    year={2024}
}

@article{liu2024ETbench,
  title={Et bench: Towards open-ended event-level video-language understanding},
  author={Liu, Ye and Ma, Zongyang and Qi, Zhongang and Wu, Yang and Shan, Ying and Chen, Chang W},
  journal={Advances in Neural Information Processing Systems},
  volume={37},
  pages={32076--32110},
  year={2024}
}

@inproceedings{grauman2024egoexo4d,
  title={Ego-exo4d: Understanding skilled human activity from first-and third-person perspectives},
  author={Grauman, Kristen and Westbury, Andrew and Torresani, Lorenzo and Kitani, Kris and Malik, Jitendra and Afouras, Triantafyllos and Ashutosh, Kumar and Baiyya, Vijay and Bansal, Siddhant and Boote, Bikram and others},
  booktitle={Proceedings of the IEEE/CVF Conference on Computer Vision and Pattern Recognition},
  pages={19383--19400},
  year={2024}
}

@inproceedings{zhou2025mlvu,
  title={Mlvu: Benchmarking multi-task long video understanding},
  author={Zhou, Junjie and Shu, Yan and Zhao, Bo and Wu, Boya and Liang, Zhengyang and Xiao, Shitao and Qin, Minghao and Yang, Xi and Xiong, Yongping and Zhang, Bo and others},
  booktitle={Proceedings of the Computer Vision and Pattern Recognition Conference},
  pages={13691--13701},
  year={2025}
}

@article{tang2024videosalmonn2_1,
  title={Enhancing multimodal LLM for detailed and accurate video captioning using multi-round preference optimization},
  author={Tang, Changli and Li, Yixuan and Yang, Yudong and Zhuang, Jimin and Sun, Guangzhi and Li, Wei and Ma, Zujun and Zhang, Chao},
  journal={arXiv preprint arXiv:2410.06682},
  year={2024}
}

@article{hu2024fiova,
  title={FIOVA: A Multi-Annotator Benchmark for Human-Aligned Video Captioning},
  author={Hu, Shiyu and Li, Xuchen and Li, Xuzhao and Zhang, Jing and Wang, Yipei and Zhao, Xin and Cheong, Kang Hao},
  journal={arXiv preprint arXiv:2410.15270},
  year={2024}
}

@article{nagrani2024neptune,
  title={Neptune: The long orbit to benchmarking long video understanding},
  author={Nagrani, Arsha and Zhang, Mingda and Mehran, Ramin and Hornung, Rachel and Gundavarapu, Nitesh Bharadwaj and Jha, Nilpa and Myers, Austin and Zhou, Xingyi and Gong, Boqing and Schmid, Cordelia and others},
  journal={arXiv preprint arXiv:2412.09582},
  year={2024}
}

@inproceedings{hong2025motionbench,
  title={Motionbench: Benchmarking and improving fine-grained video motion understanding for vision language models},
  author={Hong, Wenyi and Cheng, Yean and Yang, Zhuoyi and Wang, Weihan and Wang, Lefan and Gu, Xiaotao and Huang, Shiyu and Dong, Yuxiao and Tang, Jie},
  booktitle={Proceedings of the Computer Vision and Pattern Recognition Conference},
  pages={8450--8460},
  year={2025}
}

@inproceedings{niu2025ovobench,
  title={OVO-Bench: How Far is Your Video-LLMs from Real-World Online Video Understanding?},
  author={Niu, Junbo and Li, Yifei and Miao, Ziyang and Ge, Chunjiang and Zhou, Yuanhang and He, Qihao and Dong, Xiaoyi and Duan, Haodong and Ding, Shuangrui and Qian, Rui and others},
  booktitle={Proceedings of the Computer Vision and Pattern Recognition Conference},
  pages={18902--18913},
  year={2025}
}

@inproceedings{ren2025vista_HRVideoBench,
  title={Vista: Enhancing long-duration and high-resolution video understanding by video spatiotemporal augmentation},
  author={Ren, Weiming and Yang, Huan and Min, Jie and Wei, Cong and Chen, Wenhu},
  booktitle={Proceedings of the Computer Vision and Pattern Recognition Conference},
  pages={3804--3814},
  year={2025}
}

@inproceedings{
    hong2025worldsense,
    title={WorldSense: Evaluating Real-world Omnimodal Understanding for Multimodal {LLM}s},
    author={Jack Hong and Shilin Yan and Jiayin Cai and Xiaolong Jiang and Yao Hu and Weidi Xie},
    booktitle={The Fourteenth International Conference on Learning Representations},
    year={2026},
    url={https://openreview.net/forum?id=YxsfxAvJv4}
}

@inproceedings{zhou2025harmonyset,
  title={Harmonyset: A comprehensive dataset for understanding video-music semantic alignment and temporal synchronization},
  author={Zhou, Zitang and Mei, Ke and Lu, Yu and Wang, Tianyi and Rao, Fengyun},
  booktitle={Proceedings of the Computer Vision and Pattern Recognition Conference},
  pages={3152--3162},
  year={2025}
}

@article{cheng2025vstar,
  title={V-star: Benchmarking video-llms on video spatio-temporal reasoning},
  author={Cheng, Zixu and Hu, Jian and Liu, Ziquan and Si, Chenyang and Li, Wei and Gong, Shaogang},
  journal={arXiv preprint arXiv:2503.11495},
  year={2025}
}

@inproceedings{nagrani2025minerva,
  title={Minerva: Evaluating complex video reasoning},
  author={Nagrani, Arsha and Menon, Sachit and Iscen, Ahmet and Buch, Shyamal and Mehran, Ramin and Jha, Nilpa and Hauth, Anja and Zhu, Yukun and Vondrick, Carl and Sirotenko, Mikhail and others},
  booktitle={Proceedings of the IEEE/CVF International Conference on Computer Vision},
  pages={23968--23978},
  year={2025}
}

@inproceedings{xun2025rtvbench,
  title={RTV-Bench: Benchmarking MLLM Continuous Perception, Understanding and Reasoning through Real-Time Video},
  author={Xun, ShuHang and Tao, Sicheng and Li, Jungang and Shi, Yibo and Lin, Zhixin and Zhu, Zhanhui and Yan, Yibo and Li, Hanqian and Zhang, LingHao and Wang, Shikang and others},
  booktitle={The Thirty-ninth Annual Conference on Neural Information Processing Systems Datasets and Benchmarks Track},
  year={2025}
}

@article{yang2025vidtext,
  title={VidText: Towards Comprehensive Evaluation for Video Text Understanding},
  author={Yang, Zhoufaran and Shu, Yan and Yang, Zhifei and Zhang, Yan and Li, Yu and Lu, Keyang and Zeng, Gangyan and Liu, Shaohui and Zhou, Yu and Sebe, Nicu},
  journal={arXiv preprint arXiv:2505.22810},
  year={2025}
}

@inproceedings{he2025egoexobench,
  title={EgoExoBench: A Benchmark for First-and Third-person View Video Understanding in MLLMs},
  author={He, Yuping and Huang, Yifei and Chen, Guo and Pei, Baoqi and Xu, Jilan and Lu, Tong and Pang, Jiangmiao},
  booktitle={The Thirty-ninth Annual Conference on Neural Information Processing Systems Datasets and Benchmarks Track},
  year={2025}
}

@inproceedings{yi2025exact,
  title={ExAct: A Video-Language Benchmark for Expert Action Analysis},
  author={Yi, Han and Pan, Yulu and He, Feihong and Liu, Xinyu and Zhang, Benjamin and Oguntola, Oluwatumininu and Bertasius, Gedas},
  booktitle={The Thirty-ninth Annual Conference on Neural Information Processing Systems Datasets and Benchmarks Track},
  year={2025}
}

@inproceedings{kong2025tuna,
    title = "{TUNA}: Comprehensive Fine-grained Temporal Understanding Evaluation on Dense Dynamic Videos",
    author = "Kong, Fanheng  and
      Zhang, Jingyuan  and
      Zhang, Hongzhi  and
      Feng, Shi  and
      Wang, Daling  and
      Yu, Linhao  and
      Ji, Xingguang  and
      Tian, Yu  and
      W., V.  and
      Zhang, Fuzheng",
    editor = "Che, Wanxiang  and
      Nabende, Joyce  and
      Shutova, Ekaterina  and
      Pilehvar, Mohammad Taher",
    booktitle = "Proceedings of the 63rd Annual Meeting of the Association for Computational Linguistics (Volume 1: Long Papers)",
    month = jul,
    year = "2025",
    address = "Vienna, Austria",
    publisher = "Association for Computational Linguistics",
    url = "https://aclanthology.org/2025.acl-long.91/",
    doi = "10.18653/v1/2025.acl-long.91",
    pages = "1810--1839",
    ISBN = "979-8-89176-251-0"
}

@inproceedings{li2025videoa11y,
  title={Videoa11y: Method and dataset for accessible video description},
  author={Li, Chaoyu and Padmanabhuni, Sid and Cheema, Maryam S and Seifi, Hasti and Fazli, Pooyan},
  booktitle={Proceedings of the 2025 CHI Conference on Human Factors in Computing Systems},
  pages={1--29},
  year={2025}
}

@inproceedings{hemmworld,
  title={MMWorld: Towards Multi-discipline Multi-faceted World Model Evaluation in Videos},
  author={He, Xuehai and Feng, Weixi and Zheng, Kaizhi and Lu, Yujie and Zhu, Wanrong and Li, Jiachen and Fan, Yue and Wang, Jianfeng and Li, Linjie and Yang, Zhengyuan and others},
  booktitle={The Thirteenth International Conference on Learning Representations},
year={2025}
}

@article{xie2025maverix,
  title={MAVERIX: Multimodal Audio-Visual Evaluation Reasoning IndeX},
  author={Xie, Liuyue and Wei, George Z and Kuthiala, Avik and Zheng, Ce and Bal, Ananya and Dabhi, Mosam and Wen, Liting and Rustagi, Taru and Lai, Ethan and Khyalia, Sushil and others},
  journal={arXiv preprint arXiv:2503.21699},
  year={2025}
}

@article{yang2025wildvideo,
  title={WildVideo: Benchmarking LMMs for Understanding Video-Language Interaction},
  author={Yang, Songyuan and Yu, Weijiang and Yang, Wenjing and Liu, Xinwang and Tan, Huibin and Lan, Long and Xiao, Nong},
  journal={IEEE Transactions on Pattern Analysis and Machine Intelligence},
  year={2025},
  publisher={IEEE}
}

@article{li2024videovista,
  title={Videovista: A versatile benchmark for video understanding and reasoning},
  author={Li, Yunxin and Chen, Xinyu and Hu, Baotian and Wang, Longyue and Shi, Haoyuan and Zhang, Min},
  journal={arXiv preprint arXiv:2406.11303},
  year={2024}
}

@inproceedings{chenlongvila,
  title={LongVILA: Scaling Long-Context Visual Language Models for Long Videos},
  author={Chen, Yukang and Xue, Fuzhao and Li, Dacheng and Hu, Qinghao and Zhu, Ligeng and Li, Xiuyu and Fang, Yunhao and Tang, Haotian and Yang, Shang and Liu, Zhijian and others},
  booktitle={The Thirteenth International Conference on Learning Representations},
year={2025}
}

@inproceedings{yu2025vrbench,
  title={Vrbench: A benchmark for multi-step reasoning in long narrative videos},
  author={Yu, Jiashuo and Wu, Yue and Chu, Meng and Ren, Zhifei and Huang, Zizheng and Chu, Pei and Zhang, Ruijie and He, Yinan and Li, Qirui and Li, Songze and others},
  booktitle={Proceedings of the IEEE/CVF International Conference on Computer Vision},
  pages={21655--21666},
  year={2025}
}

@inproceedings{wang2025Trust_videoLLMs,
  title={Benchmarking Trustworthiness in Multimodal LLMs for Video Understanding},
  author={Wang, Youze and Chen, Zijun and Chen, Ruoyu and Gu, Shishen and Hu, Wenbo and Liu, Jiayang and Dong, Yinpeng and Su, Hang and Zhu, Jun and Wang, Meng and others},
  booktitle={Proceedings of the AAAI Conference on Artificial Intelligence},
  volume={40},
  number={44},
  pages={37979--37987},
  year={2026}
}

@inproceedings{li2025causalstep,
  title={Causalstep: A benchmark for explicit stepwise causal reasoning in videos},
  author={Li, Xuchen and Li, Xuzhao and Hu, Shiyu and Huang, Kaiqi and Zhang, Wentao},
  booktitle={Proceedings of the AAAI Conference on Artificial Intelligence},
  volume={40},
  number={8},
  pages={6530--6538},
  year={2026}
}

@article{tang2025videoSALMONN2_2,
  title={video-SALMONN 2: Captioning-Enhanced Audio-Visual Large Language Models},
  author={Tang, Changli and Li, Yixuan and Yang, Yudong and Zhuang, Jimin and Sun, Guangzhi and Li, Wei and Ma, Zejun and Zhang, Chao},
  journal={arXiv preprint arXiv:2506.15220},
  year={2025}
}

@article{kong2025sivbench,
  title={SIV-Bench: A Video Benchmark for Social Interaction Understanding and Reasoning},
  author={Kong, Fanqi and Zu, Weiqin and Chen, Xinyu and Yang, Yaodong and Zhu, Song-Chun and Feng, Xue},
  journal={arXiv preprint arXiv:2506.05425},
  year={2025}
}

@inproceedings{jiangmmsearch,
  title={Mmsearch: Unveiling the potential of large models as multi-modal search engines},
  author={Jiang, Dongzhi and Zhang, Renrui and Guo, Ziyu and Wu, Yanmin and Qiu, Pengshuo and Lu, Pan and Chen, Zehui and Song, Guanglu and Gao, Peng and Liu, Yu and others},
  booktitle={The Thirteenth International Conference on Learning Representations},
year={2025}
}

@inproceedings{yuan2025momentseeker,
  title={MomentSeeker: A Task-Oriented Benchmark For Long-Video Moment Retrieval},
  author={Yuan, Huaying and Liu, Zheng and Wang, Yueze and Zhou, Junjie and Liang, Zhengyang and Zhao, Bo and Cao, Zhao and Wen, Ji-Rong and Dou, Zhicheng and others},
  booktitle={The Thirty-ninth Annual Conference on Neural Information Processing Systems Datasets and Benchmarks Track},
  year={2025}
}

@article{zhang2024mdibench,
  title={Multi-Dimensional Insights: Benchmarking Real-World Personalization in Large Multimodal Models},
  author={Zhang, YiFan and Lei, Shanglin and Qiao, Runqi and GongQue, Zhuoma and Song, Xiaoshuai and Dong, Guanting and Tan, Qiuna and Wei, Zhe and Yang, Peiqing and Tian, Ye and others},
  journal={arXiv preprint arXiv:2412.12606},
  year={2024}
}

@inproceedings{madan2025mipgaf,
  title={MIP-GAF: A MLLM-annotated Benchmark for Most Important Person Localization and Group Context Understanding},
  author={Madan, Surbhi and Ghosh, Shreya and Sookha, Lownish Rai and Ganaie, MA and Subramanian, Ramanathan and Dhall, Abhinav and Gedeon, Tom},
  booktitle={2025 IEEE/CVF Winter Conference on Applications of Computer Vision (WACV)},
  pages={1467--1476},
  year={2025},
  organization={IEEE}
}

@article{zou2024seconds,
  title={From seconds to hours: Reviewing multimodal large language models on comprehensive long video understanding},
  author={Zou, Heqing and Luo, Tianze and Xie, Guiyang and Lv, Fengmao and Wang, Guangcong and Chen, Junyang and Wang, Zhuochen and Zhang, Hansheng and Zhang, Huaijian and others},
  journal={arXiv preprint arXiv:2409.18938},
  year={2024}
}

@inproceedings{biester2025sports,
  title={Sports and Women’s Sports: Gender Bias in Text Generation with Olympic Data},
  author={Biester, Laura},
  booktitle={Proceedings of the 2025 Conference of the Nations of the Americas Chapter of the Association for Computational Linguistics: Human Language Technologies (Volume 2: Short Papers)},
  pages={195--205},
  year={2025}
}

@article{papini2025balancing,
  title={Balancing Act: Generative AI Tools and Scope of Practice in Health Coaching},
  author={Papini, Natalie M and Meyer, Megan and Squires, Nikole D and Clifford, Dawn},
  journal={American Journal of Health Promotion},
  pages={08901171251340383},
  year={2025},
  publisher={SAGE Publications Sage CA: Los Angeles, CA}
}

@misc{NSCA_CSCS_Exam,
  author       = {{National Strength and Conditioning Association (NSCA)}},
  title        = {Certified Strength and Conditioning Specialist (CSCS) Exam Description},
  year         = {2025},
  howpublished = {\url{https://www.nsca.com/certification/cscs/certified-strength-and-conditioning-specialist-exam-description}},
  note         = {Accessed: October 5, 2025}
}

@article{chen2024yourskatingcoach,
  title={YourSkatingCoach: A Figure Skating Video Benchmark for Fine-Grained Element Analysis},
  author={Chen, Wei-Yi and Lin, Yi-Ling and Su, Yu-An and Yeh, Wei-Hsin and Ku, Lun-Wei},
  journal={arXiv preprint arXiv:2410.20427},
  year={2024}
}

@inproceedings{vardhan2022PACE,
  title={Walking with pace-personalized and automated coaching engine},
  author={Vardhan, Madhurima and Hegde, Narayan and Merugu, Srujana and Prabhat, Shantanu and Nathani, Deepak and Seneviratne, Martin and Muhammad, Nur and Reddy, Pranay and Lakshminarasimhan, Sriram and Singh, Rahul and others},
  booktitle={Proceedings of the 30th ACM Conference on User Modeling, Adaptation and Personalization},
  pages={57--68},
  year={2022}
}

@article{thoppilan2022lamda,
  title={Lamda: Language models for dialog applications},
  author={Thoppilan, Romal and De Freitas, Daniel and Hall, Jamie and Shazeer, Noam and Kulshreshtha, Apoorv and Cheng, Heng-Tze and Jin, Alicia and Bos, Taylor and Baker, Leslie and Du, Yu and others},
  journal={arXiv preprint arXiv:2201.08239},
  year={2022}
}

@article{dubey2024llama3,
  title={The llama 3 herd of models},
  author={Grattafiori, Aaron and Dubey, Abhimanyu and Jauhri, Abhinav and Pandey, Abhinav and Kadian, Abhishek and Al-Dahle, Ahmad and Letman, Aiesha and Mathur, Akhil and Schelten, Alan and Vaughan, Alex and others},
  journal={arXiv preprint arXiv:2407.21783},
  year={2024}
}

@inproceedings{devlin2019bert,
  title={Bert: Pre-training of deep bidirectional transformers for language understanding},
  author={Devlin, Jacob and Chang, Ming-Wei and Lee, Kenton and Toutanova, Kristina},
  booktitle={Proceedings of the 2019 conference of the North American chapter of the association for computational linguistics: human language technologies, volume 1 (long and short papers)},
  pages={4171--4186},
  year={2019}
}

@article{hurst2024gpt4o,
  title={Gpt-4o system card},
  author={Hurst, Aaron and Lerer, Adam and Goucher, Adam P and Perelman, Adam and Ramesh, Aditya and Clark, Aidan and Ostrow, AJ and Welihinda, Akila and Hayes, Alan and Radford, Alec and others},
  journal={arXiv preprint arXiv:2410.21276},
  year={2024}
}

@article{raffel2020T5,
  title={Exploring the limits of transfer learning with a unified text-to-text transformer},
  author={Raffel, Colin and Shazeer, Noam and Roberts, Adam and Lee, Katherine and Narang, Sharan and Matena, Michael and Zhou, Yanqi and Li, Wei and Liu, Peter J},
  journal={Journal of machine learning research},
  volume={21},
  number={140},
  pages={1--67},
  year={2020}
}

@article{zhang2018cmedqa2,
  title={Multi-scale attentive interaction networks for chinese medical question answer selection},
  author={Zhang, Sheng and Zhang, Xin and Wang, Hui and Guo, Lixiang and Liu, Shanshan},
  journal={IEEE Access},
  volume={6},
  pages={74061--74071},
  year={2018},
  publisher={IEEE}
}

@article{team2024qwen2,
  title={Qwen2 technical report},
  author={Team, Qwen and others},
  journal={arXiv preprint arXiv:2407.10671},
  volume={2},
  pages={3},
  year={2024}
}

@article{abdin2024phi3,
  title={Phi-3 Technical Report: A Highly Capable Language Model Locally on Your Phone},
  author={Abdin, Marah and Aneja, Jyoti and Awadalla, Hany and Awadallah, Ahmed and Awan, Ammar Ahmad and Bach, Nguyen and Bahree, Amit and Bakhtiari, Arash and Bao, Jianmin and Behl, Harkirat and others},
  journal={arXiv preprint arXiv:2404.14219},
  year={2024}
}

@article{chan2021capture,
  title={Capture-24: Activity tracker dataset for human activity recognition},
  author={Chan Chang, S and Walmsley, R and Gershuny, J and Harms, T and Thomas, E and Milton, K and Kelly, P and Foster, C and Wong, A and Gray, N and others},
  year={2021},
  publisher={University of Oxford}
}

@article{chiang2023vicuna,
  title={Vicuna: An open-source chatbot impressing gpt-4 with 90\%* chatgpt quality},
  author={Chiang, Wei-Lin and Li, Zhuohan and Lin, Ziqing and Sheng, Ying and Wu, Zhanghao and Zhang, Hao and Zheng, Lianmin and Zhuang, Siyuan and Zhuang, Yonghao and Gonzalez, Joseph E and others},
  journal={See https://vicuna. lmsys. org (accessed 14 April 2023)},
  volume={2},
  number={3},
  pages={6},
  year={2023}
}

@misc{sportsvision2025nsva_subset,
  author       = {Sportsvision},
  title        = {NSVA Subset: Basketball Video-Text Dataset},
  howpublished = {\url{https://huggingface.co/datasets/sportsvision/nsva_subset}},
  year         = {2025},
  note         = {Hugging Face},
}

@article{cheng2024videollama2,
  title={Videollama 2: Advancing spatial-temporal modeling and audio understanding in video-llms},
  author={Cheng, Zesen and Leng, Sicong and Zhang, Hang and Xin, Yifei and Li, Xin and Chen, Guanzheng and Zhu, Yongxin and Zhang, Wenqi and Luo, Ziyang and Zhao, Deli and others},
  journal={arXiv preprint arXiv:2406.07476},
  year={2024}
}

@inproceedings{deliege2021soccernetV2,
  title={Soccernet-v2: A dataset and benchmarks for holistic understanding of broadcast soccer videos},
  author={Deliege, Adrien and Cioppa, Anthony and Giancola, Silvio and Seikavandi, Meisam J and Dueholm, Jacob V and Nasrollahi, Kamal and Ghanem, Bernard and Moeslund, Thomas B and Van Droogenbroeck, Marc},
  booktitle={Proceedings of the IEEE/CVF conference on computer vision and pattern recognition},
  pages={4508--4519},
  year={2021}
}

@inproceedings{dosovitskiy_vit,
  title={An Image is Worth 16x16 Words: Transformers for Image Recognition at Scale},
  author={Dosovitskiy, Alexey and Beyer, Lucas and Kolesnikov, Alexander and Weissenborn, Dirk and Zhai, Xiaohua and Unterthiner, Thomas and Dehghani, Mostafa and Minderer, Matthias and Heigold, Georg and Gelly, Sylvain and others},
  booktitle={International Conference on Learning Representations},
year={2021}
}

@article{zhang2021tennis7,
  title={Vid2player: Controllable video sprites that behave and appear like professional tennis players},
  author={Zhang, Haotian and Sciutto, Cristobal and Agrawala, Maneesh and Fatahalian, Kayvon},
  journal={ACM Transactions on Graphics (TOG)},
  volume={40},
  number={3},
  pages={1--16},
  year={2021},
  publisher={ACM New York, NY}
}

@article{liu2023llava,
  title={Visual instruction tuning},
  author={Liu, Haotian and Li, Chunyuan and Wu, Qingyang and Lee, Yong Jae},
  journal={Advances in neural information processing systems},
  volume={36},
  pages={34892--34916},
  year={2023}
}

@article{lillavaonevision,
  title={LLaVA-OneVision: Easy Visual Task Transfer},
  author={Li, Bo and Zhang, Yuanhan and Guo, Dong and Zhang, Renrui and Li, Feng and Zhang, Hao and Zhang, Kaichen and Zhang, Peiyuan and Li, Yanwei and Liu, Ziwei and others},
  journal={Transactions on Machine Learning Research},
year={2025}
}

@inproceedings{li2024llavanext,
  title={Llava-interleave: Tackling multi-image, video, and 3d in large multimodal models},
  author={Li, Feng and Zhang, Renrui and Zhang, Hao and Zhang, Yuanhan and Li, Bo and Li, Wei and Ma, Zejun and Li, Chunyuan},
  booktitle={The Thirteenth International Conference on Learning Representations},
  year={2024}
}

@misc{hudl2024wyscout,
  author       = {Hudl},
  title        = {Wyscout: Football Data and Analytics Platform},
  year         = {2024},
  howpublished = {\url{https://wyscout.com}},
  note         = {Accessed: October 5, 2025},
}

@misc{vanzandycke2024deepsport,
  author       = {Gabriel Van Zandycke},
  title        = {DeepSport Dataset},
  year         = {2024},
  howpublished = {\url{https://www.kaggle.com/datasets/gabrielvanzandycke/deepsport-dataset}},
  note         = {Accessed: October 5, 2025},
}

@misc{petti2021baseballr,
  author       = {B. Petti and S. Gilani},
  title        = {baseballr: The Sports Dataverse’s R Package for Baseball Data},
  year         = {2021},
  howpublished = {\url{https://billpetti.github.io/baseballr/}},
  note         = {Accessed: October 5, 2025},
}

@article{brown2020gpt3,
  title={Language models are few-shot learners},
  author={Brown, Tom and Mann, Benjamin and Ryder, Nick and Subbiah, Melanie and Kaplan, Jared D and Dhariwal, Prafulla and Neelakantan, Arvind and Shyam, Pranav and Sastry, Girish and Askell, Amanda and others},
  journal={Advances in neural information processing systems},
  volume={33},
  pages={1877--1901},
  year={2020}
}

@inproceedings{black2022gptneox,
  title={GPT-NeoX-20B: An Open-Source Autoregressive Language Model},
  author={Black, Sidney and Biderman, Stella and Hallahan, Eric and Anthony, Quentin and Gao, Leo and Golding, Laurence and He, Horace and Leahy, Connor and McDonell, Kyle and Phang, Jason and others},
  booktitle={Proceedings of BigScience Episode\# 5--Workshop on Challenges \& Perspectives in Creating Large Language Models},
  pages={95--136},
  year={2022}
}

@misc{sportvu2025,
  author       = {S. LLC},
  title        = {SportVU},
  year         = {2025},
  howpublished = {\url{https://www.stats.com/sportvu-basketball/}},
  note         = {Accessed: October 5, 2025},
}

@misc{openai2024gpt4v,
  author       = {{OpenAI}},
  title        = {GPT-4V(ision) System Card},
  year         = {2024},
  howpublished = {\url{https://cdn.openai.com/papers/GPTV_System_Card.pdf?utm_source=chatgpt.com}},
  note         = {Accessed: October 6, 2025},
}

@article{ouyang2022gpt3_5,
  title={Training language models to follow instructions with human feedback},
  author={Ouyang, Long and Wu, Jeffrey and Jiang, Xu and Almeida, Diogo and Wainwright, Carroll and Mishkin, Pamela and Zhang, Chong and Agarwal, Sandhini and Slama, Katarina and Ray, Alex and others},
  journal={Advances in neural information processing systems},
  volume={35},
  pages={27730--27744},
  year={2022}
}

@misc{sportdevs2025handball,
  author       = {SportDevs},
  title        = {Handball API},
  year         = {2025},
  howpublished = {\url{https://sportdevs.com/handball}},
  note         = {Accessed: October 5, 2025},
}

@inproceedings{wang2023shuttleset,
  title={Shuttleset: A human-annotated stroke-level singles dataset for badminton tactical analysis},
  author={Wang, Wei-Yao and Huang, Yung-Chang and Ik, Tsi-Ui and Peng, Wen-Chih},
  booktitle={Proceedings of the 29th ACM SIGKDD Conference on Knowledge Discovery and Data Mining},
  pages={5126--5136},
  year={2023}
}

@inproceedings{ban2022badmintondb,
  title={Badmintondb: A badminton dataset for player-specific match analysis and prediction},
  author={Ban, Kar-Weng and See, John and Abdullah, Junaidi and Loh, Yuen Peng},
  booktitle={Proceedings of the 5th international ACM workshop on multimedia content analysis in sports},
  pages={47--54},
  year={2022}
}

@misc{jiang2023mistral7b,
      title={Mistral 7B}, 
      author={Albert Q. Jiang and Alexandre Sablayrolles and Arthur Mensch and Chris Bamford and Devendra Singh Chaplot and Diego de las Casas and Florian Bressand and Gianna Lengyel and Guillaume Lample and Lucile Saulnier and Lélio Renard Lavaud and Marie-Anne Lachaux and Pierre Stock and Teven Le Scao and Thibaut Lavril and Thomas Wang and Timothée Lacroix and William El Sayed},
      year={2023},
      eprint={2310.06825},
      archivePrefix={arXiv},
      primaryClass={cs.CL},
      url={https://arxiv.org/abs/2310.06825}, 
}

@article{xu2019fisv,
  title={Learning to score figure skating sport videos},
  author={Xu, Chengming and Fu, Yanwei and Zhang, Bing and Chen, Zitian and Jiang, Yu-Gang and Xue, Xiangyang},
  journal={IEEE transactions on circuits and systems for video technology},
  volume={30},
  number={12},
  pages={4578--4590},
  year={2019},
  publisher={IEEE}
}

@inproceedings{xia2023fs1000,
  title={Skating-mixer: Long-term sport audio-visual modeling with mlps},
  author={Xia, Jingfei and Zhuge, Mingchen and Geng, Tiantian and Fan, Shun and Wei, Yuantai and He, Zhenyu and Zheng, Feng},
  booktitle={Proceedings of the AAAI Conference on Artificial Intelligence},
  volume={37},
  number={3},
  pages={2901--2909},
  year={2023}
}

@inproceedings{ji2023finefs,
  title={Localization-assisted uncertainty score disentanglement network for action quality assessment},
  author={Ji, Yanli and Ye, Lingfeng and Huang, Huili and Mao, Lijing and Zhou, Yang and Gao, Lingling},
  booktitle={Proceedings of the 31st ACM International Conference on Multimedia},
  pages={8590--8597},
  year={2023}
}

@inproceedings{chen2024internvl_2,
  title={Internvl: Scaling up vision foundation models and aligning for generic visual-linguistic tasks},
  author={Chen, Zhe and Wu, Jiannan and Wang, Wenhai and Su, Weijie and Chen, Guo and Xing, Sen and Zhong, Muyan and Zhang, Qinglong and Zhu, Xizhou and Lu, Lewei and others},
  booktitle={Proceedings of the IEEE/CVF conference on computer vision and pattern recognition},
  pages={24185--24198},
  year={2024}
}

@inproceedings{parmar2022fitnessaqa,
  title={Domain knowledge-informed self-supervised representations for workout form assessment},
  author={Parmar, Paritosh and Gharat, Amol and Rhodin, Helge},
  booktitle={European Conference on Computer Vision},
  pages={105--123},
  year={2022},
  organization={Springer}
}

@article{zhang2025llavavideo,
  title={LLaVA-video: Video instruction tuning with synthetic data},
  author={Zhang, Yuanhan and Wu, Jinming and Li, Wei and Li, Bo and Ma, Zejun and Liu, Ziwei and Li, Chunyuan},
  journal={Transactions on Machine Learning Research},
  year={2025}
}

@article{glm2024chatglm,
  title={Chatglm: A family of large language models from glm-130b to glm-4 all tools},
  author={GLM, Team and Zeng, Aohan and Xu, Bin and Wang, Bowen and Zhang, Chenhui and Yin, Da and Zhang, Dan and Rojas, Diego and Feng, Guanyu and Zhao, Hanlin and others},
  journal={arXiv preprint arXiv:2406.12793},
  year={2024}
}

@inproceedings{anguita2013public,
  title={A public domain dataset for human activity recognition using smartphones.},
  author={Anguita, Davide and Ghio, Alessandro and Oneto, Luca and Parra, Xavier and Reyes-Ortiz, Jorge Luis and others},
  booktitle={Esann},
  volume={3},
  number={1},
  pages={3--4},
  year={2013}
}

@inproceedings{maaz2024videochatgpt,
  title={Video-ChatGPT: Towards Detailed Video Understanding via Large Vision and Language Models},
  author={Maaz, Muhammad and Rasheed, Hanoona and Khan, Salman and Khan, Fahad},
  booktitle={Proceedings of the 62nd Annual Meeting of the Association for Computational Linguistics (Volume 1: Long Papers)},
  pages={12585--12602},
  year={2024}
}

@article{touvron2023llama2,
  title={Llama 2: Open foundation and fine-tuned chat models},
  author={Touvron, Hugo and Martin, Louis and Stone, Kevin and Albert, Peter and Almahairi, Amjad and Babaei, Yasmine and Bashlykov, Nikolay and Batra, Soumya and Bhargava, Prajjwal and Bhosale, Shruti and others},
  journal={arXiv preprint arXiv:2307.09288},
  year={2023}
}

@misc{anthropic2025claude35sonnet,
  title        = {Claude 3.5 Sonnet},
  author       = {Anthropic},
  year         = {2025},
  howpublished = {\url{https://www.anthropic.com/news/claude-3-5-sonnet}},
  note         = {Accessed: October 5, 2025}
}

@article{wang2024qwen2vl,
  title={Qwen2-vl: Enhancing vision-language model's perception of the world at any resolution},
  author={Wang, Peng and Bai, Shuai and Tan, Sinan and Wang, Shijie and Fan, Zhihao and Bai, Jinze and Chen, Keqin and Liu, Xuejing and Wang, Jialin and Ge, Wenbin and others},
  journal={arXiv preprint arXiv:2409.12191},
  year={2024}
}

@misc{love2015bbc,
  author       = {Love, C.},
  title        = {BBC Football Commentary Data -- Webscraping},
  year         = {2015},
  howpublished = {\url{https://sciolisticramblings.wordpress.com/2015/08/24/bbc-football-commentary-data-webscraping/}},
  note         = {Accessed: October 5, 2025}
}

@misc{secareanu2023football,
  author       = {Secareanu, A.},
  title        = {Football Events},
  year         = {2023},
  howpublished = {\url{https://www.kaggle.com/datasets/secareanualin/football-events}},
  note         = {Accessed: October 5, 2025}
}

@article{touvron2023llama,
  title={LLaMA: Open and Efficient Foundation Language Models},
  author={Touvron, Hugo and Lavril, Thibaut and Izacard, Gautier and Martinet, Xavier and Lachaux, Marie-Anne and Lacroix, Timoth{\'e}e and Rozi{\`e}re, Baptiste and Goyal, Naman and Hambro, Eric and Azhar, Faisal and others},
  journal={arXiv preprint arXiv:2302.13971},
  year={2023}
}

@article{xi2025simple,
  title={A simple yet effective knowledge guided method for entity-aware video captioning on a basketball benchmark},
  author={Xi, Zeyu and Shi, Ge and Li, Xuefen and Yan, Junchi and Li, Zun and Wu, Lifang and Liu, Zilin and Wang, Liang},
  journal={Neurocomputing},
  volume={619},
  pages={129177},
  year={2025},
  publisher={Elsevier}
}

@inproceedings{giancola2018soccernetV1,
  title={Soccernet: A scalable dataset for action spotting in soccer videos},
  author={Giancola, Silvio and Amine, Mohieddine and Dghaily, Tarek and Ghanem, Bernard},
  booktitle={Proceedings of the IEEE conference on computer vision and pattern recognition workshops},
  pages={1711--1721},
  year={2018}
}

@article{yu2025temporally,
  title={Temporally-Grounded Language Generation: A Benchmark for Real-Time Vision-Language Models},
  author={Yu, Keunwoo Peter and Chai, Joyce},
  journal={arXiv preprint arXiv:2505.11326},
  year={2025}
}

@inproceedings{chen2024videollmonline,
  title={Videollm-online: Online video large language model for streaming video},
  author={Chen, Joya and Lv, Zhaoyang and Wu, Shiwei and Lin, Kevin Qinghong and Song, Chenan and Gao, Difei and Liu, Jia-Wei and Gao, Ziteng and Mao, Dongxing and Shou, Mike Zheng},
  booktitle={Proceedings of the IEEE/CVF Conference on Computer Vision and Pattern Recognition},
  pages={18407--18418},
  year={2024}
}

@inproceedings{zhang2023videollama,
  title={Video-LLaMA: An Instruction-tuned Audio-Visual Language Model for Video Understanding},
  author={Zhang, Hang and Li, Xin and Bing, Lidong},
  booktitle={Proceedings of the 2023 Conference on Empirical Methods in Natural Language Processing: System Demonstrations},
  pages={543--553},
  year={2023}
}

@article{sanh2019distilbert,
  title={DistilBERT, a distilled version of BERT: smaller, faster, cheaper and lighter},
  author={Sanh, Victor and Debut, Lysandre and Chaumond, Julien and Wolf, Thomas},
  journal={arXiv preprint arXiv:1910.01108},
  year={2019}
}

@misc{mistral_large_instruct_2411,
  author       = {Mistral AI Team},
  title        = {Mistral-Large-Instruct-2411: A 123B Parameter Dense Language Model},
  year         = {2024},
  howpublished = {\url{https://huggingface.co/mistralai/Mistral-Large-Instruct-2411}},
  note         = {Accessed: October 5, 2025}
}

@article{liu2019roberta,
  title={Roberta: A robustly optimized bert pretraining approach},
  author={Liu, Yinhan and Ott, Myle and Goyal, Naman and Du, Jingfei and Joshi, Mandar and Chen, Danqi and Levy, Omer and Lewis, Mike and Zettlemoyer, Luke and Stoyanov, Veselin},
  journal={arXiv preprint arXiv:1907.11692},
  year={2019}
}

@inproceedings{xue2021mt5,
  title={mT5: A Massively Multilingual Pre-trained Text-to-Text Transformer},
  author={Xue, Linting and Constant, Noah and Roberts, Adam and Kale, Mihir and Al-Rfou, Rami and Siddhant, Aditya and Barua, Aditya and Raffel, Colin},
  booktitle={Proceedings of the 2021 Conference of the North American Chapter of the Association for Computational Linguistics: Human Language Technologies},
  pages={483--498},
  year={2021}
}

@misc{patel2022nbaapi,
  author       = {Patel, S.},
  title        = {An API client package to access the APIs for NBA.com},
  year         = {2022},
  month        = {Mar},
  howpublished = {\url{https://github.com/swar/nba_api}},
  note         = {Accessed: October 5, 2023}
}

@inproceedings{wiseman2017ROTO,
  title={Challenges in Data-to-Document Generation},
  author={Wiseman, Sam and Shieber, Stuart M and Rush, Alexander M},
  booktitle={Proceedings of the 2017 Conference on Empirical Methods in Natural Language Processing},
  pages={2253--2263},
  year={2017}
}

@inproceedings{puduppully2019MLB,
  title={Data-to-text Generation with Entity Modeling},
  author={Puduppully, Ratish and Dong, Li and Lapata, Mirella},
  booktitle={Proceedings of the 57th Annual Meeting of the Association for Computational Linguistics},
  pages={2023--2035},
  year={2019}
}

@article{chen2021copilot,
  title={Evaluating large language models trained on code},
  author={Chen, Mark and Tworek, Jerry and Jun, Heewoo and Yuan, Qiming and Pinto, Henrique Ponde De Oliveira and Kaplan, Jared and Edwards, Harri and Burda, Yuri and Joseph, Nicholas and Brockman, Greg and others},
  journal={arXiv preprint arXiv:2107.03374},
  year={2021}
}

@inproceedings{li2022blip,
  title={Blip: Bootstrapping language-image pre-training for unified vision-language understanding and generation},
  author={Li, Junnan and Li, Dongxu and Xiong, Caiming and Hoi, Steven},
  booktitle={International conference on machine learning},
  pages={12888--12900},
  year={2022},
  organization={PMLR}
}

@article{liu2024deepseekv3,
  title={Deepseek-v3 technical report},
  author={Liu, Aixin and Feng, Bei and Xue, Bing and Wang, Bingxuan and Wu, Bochao and Lu, Chengda and Zhao, Chenggang and Deng, Chengqi and Zhang, Chenyu and Ruan, Chong and others},
  journal={arXiv preprint arXiv:2412.19437},
  year={2024}
}

@inproceedings{held2023vars,
  title={VARS: Video assistant referee system for automated soccer decision making from multiple views},
  author={Held, Jan and Cioppa, Anthony and Giancola, Silvio and Hamdi, Abdullah and Ghanem, Bernard and Van Droogenbroeck, Marc},
  booktitle={Proceedings of the IEEE/CVF Conference on Computer Vision and Pattern Recognition},
  pages={5086--5097},
  year={2023}
}

@inproceedings{lin2024videollava,
  title={Video-LLaVA: Learning United Visual Representation by Alignment Before Projection},
  author={Lin, Bin and Ye, Yang and Zhu, Bin and Cui, Jiaxi and Ning, Munan and Jin, Peng and Yuan, Li},
  booktitle={Proceedings of the 2024 Conference on Empirical Methods in Natural Language Processing},
  pages={5971--5984},
  year={2024}
}

@misc{anthropic2024claude3opus,
  author       = {Anthropic},
  title        = {Claude 3 Opus Model Card},
  year         = {2024},
  howpublished = {\url{https://www.anthropic.com/claude-3-model-card}},
  note         = {Accessed: October 5, 2025}
}

@article{xu2024assessment,
  title={Assessment of Personalized Exercise Prescriptions Issued by ChatGPT 4.0 and Intelligent Health Promotion Systems for Patients with Hypertension Comorbidities Based on the Transtheoretical Model: A Comparative Analysis},
  author={Xu, Yang and Liu, Qiankun and Pang, Jiaxue and Zeng, Chunlu and Ma, Xiaoqing and Li, Pengyao and Ma, Li and Huang, Juju and Xie, Hui},
  journal={Journal of Multidisciplinary Healthcare},
  pages={5063--5078},
  year={2024},
  publisher={Taylor \& Francis}
}

@article{wang2024artificial,
  title={Artificial intelligence in physical education: comprehensive review and future teacher training strategies},
  author={Wang, Yuping and Wang, Xinyan},
  journal={Frontiers in public health},
  volume={12},
  pages={1484848},
  year={2024},
  publisher={Frontiers Media SA}
}

@article{pajo202510,
  title={A 10-WEEK LARGE LANGUAGE MODEL-GENERATED (LLM) VERSUS HUMAN-MADE VOLLEYBALL TRAINING PROGRAM ON THE JUMPING PERFORMANCE OF COLLEGIATE VOLLEYBALL ATHLETES},
  author={Pajo, Leonard Sydrick and Rabuya, Rey and Andacao, Arvin and Tuano, Arianne Michael Sim and Lobo, Joseph},
  journal={Journal of Physical Education},
  volume={36},
  pages={e3611},
  year={2025},
  publisher={SciELO Brasil}
}

\clearpage
\twocolumn[\vspace*{2em}] 

\appendix

\begin{figure}[t]
    \centering
  \includegraphics[width=0.95\columnwidth]{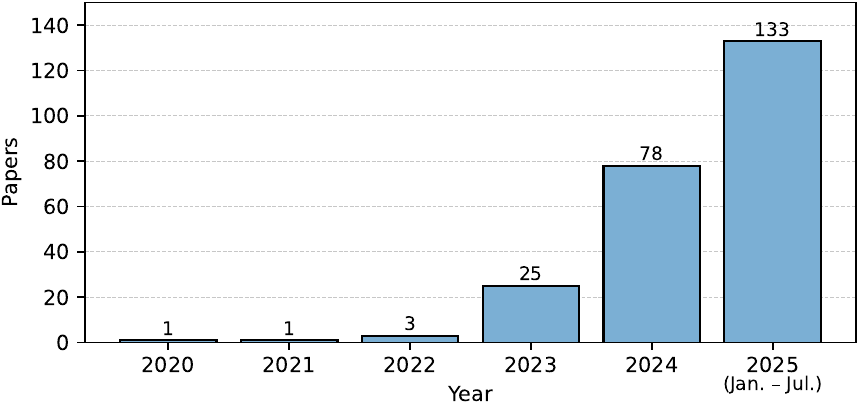}
  \caption{Papers on large models in sports over the years (data for 2025 is up to July).}
  \label{fig:year}
  \vspace{-0.5cm}
\end{figure}

\section{Methodology for Literature Selection}
\label{app_sec:appendix_prisma}

In this section, we detail the systematic methodology employed for literature identification, screening, and selection.
To capture the fragmented and rapidly evolving landscape of large models in sports, we adopted a \textbf{systematic snowballing methodology}~\cite{snowballing}, adhering to the reporting standards of the \textbf{Preferred Reporting Items for Systematic Reviews and Meta-Analysis} (\textbf{PRISMA}) statement~\cite{page2021prisma}.
This dual-direction strategy (leveraging both reference lists and citation networks) is particularly effective for interdisciplinary fields such as large models in sports, ensuring high relevance by tracing semantic connections rather than relying solely on keyword indexing.

\subsection{Construction of the Start Set}

The effectiveness of snowballing relies heavily on the quality of the initial start set. Instead of a broad, potentially noisy keyword search, we established our foundation by identifying \textbf{3} highly relevant and comprehensive survey papers based on domain expertise~\citep{xia2024language, zhou2025artificial, zhao2025survey}. These papers serve as our "seeds" for initiating the iterative snowballing process. These papers were chosen for their:

\noindent\textbf{Comprehensiveness.} Collectively, they cover the entire spectrum from traditional deep learning to modern large models.

\noindent\textbf{Recency.} All selected seeds were published in 2024--2025, ensuring the survey is anchored in the most current research landscape.

\noindent\textbf{Academic Standing.} The set combines rigorous articles from premier journals with pioneering preprints that address the rapid evolution of large models before formal publication cycles.

\noindent\textbf{Connectivity.} They serve as central hubs in the citation network, linking to a wide range of task-specific studies.

\subsection{Iterative Snowballing Procedure}

Starting from these seed papers, we performed iterative forward and backward snowballing to expand our corpus. To manage the scale of the literature and ensure precision, we applied a specific Boolean query as a filtering mechanism during the forward pass.

\noindent\textbf{Backward Snowballing.} We scrutinized the reference lists of the included papers to uncover relevant prior studies and foundational works.

\noindent\textbf{Forward Snowballing.} We leveraged Google Scholar's ``Cited by'' feature to access the citation list of each paper. To efficiently filter out out-of-domain works from the large volume of citations, we enabled the ``Search within citing articles'' option and applied the following Boolean search string:

\begin{querybox}
    "Sports" AND ("Large Language Model" OR "LLM" OR "GPT" OR "BERT" OR "T5")
\end{querybox}
    
This step allowed us to strictly identify studies that integrate large models within sports contexts, capturing the latest research developments up to July 2025.

\noindent\textbf{Iteration \& Saturation.} Newly identified papers that met the inclusion criteria were added to the set and treated as new seeds. This cycle was repeated until theoretical saturation was reached (i.e., the filtered search yielded no new relevant papers).

\subsection{Inclusion and Exclusion Criteria}

To isolate relevant studies from the retrieved pool, we applied the following rigorous filters across four dimensions:

\noindent\textbf{Research Topic.}
We included studies that target tasks within the sports domain or involve sports data analysis, provided that they utilize large \mbox{models} (e.g., LLMs, MLLMs) as a core methodological component. To maintain the survey's specific focus on the era of large models, we excluded studies that rely solely on traditional deep learning architectures (e.g., CNNs, LSTMs) without the integration of large models.

\noindent\textbf{Publication Type.}
To ensure technical depth, scientific rigor, and mitigate the risk of low-quality evidence, we restricted our selection to full-length academic contributions. Included works comprise peer-reviewed conference and journal papers, as well as cutting-edge preprints that represent the latest advancements in the field. We excluded non-technical documents such as editorials, posters, extended abstracts, opinion pieces, and short papers that lacked sufficient implementation details or experimental validation.

\noindent\textbf{Time Window.}
We defined a specific temporal scope to align with the emergence and proliferation of large models. The search and inclusion window was strictly defined from January 1, 2020, to July 31, 2025. 
Although the final search and screening process was executed on October 4, 2025, we enforced this cutoff date to ensure a consistent timeframe for data analysis.

\noindent\textbf{Language.}
To ensure accessibility and consistent analysis, we included only articles written in English. Studies published in other languages were excluded.

\subsection{Selection Results}

Throughout the iterative forward and backward snowballing process, we examined a cumulative total of approximately \textbf{2,200} candidate records. After rigorously applying the inclusion and exclusion criteria to these candidates, a final set of \textbf{241} core academic papers was selected for this survey. The rapid growth trend and temporal distribution of these included works are illustrated in Figure~\ref{fig:year}.
\section{More Details on Large Model Applications in Sports}
\label{app_sec:appendix_app}


This section serves as a comprehensive supplement to Section~\ref{sec:app}, offering a detailed literature review of specific studies and methodologies. 
Given that task definitions, analyses of large model–related technologies, and common evaluation metrics have been elaborated in the main text, this section will focus on \textbf{systematically listing} the specific contributions and relevant content of each research work.
The organization follows the taxonomy illustrated in Figure~\ref{app_and_task_figure}, detailing applications across the \textbf{6 stakeholder groups} and \textbf{19 specific tasks}.

\subsection{Applications for Athletes and Trainers}
\label{append_app:ath}

\noindent\textbf{Exercise and Training Plans.}
Recent AI coaches powered by LLMs have significantly streamlined the generation of effective training plans. 
Many works have used LLMs to generate exercise prescriptions for various health conditions and fitness goals~\citep{app4ath_pla_cavazzotto2024chatgpt, app4ath_pla_cosentinotowards, app4ath_pla_puce2025harnessing, app4ath_pla_papini2025balancing, masagca2025ai, lederman2025promises}, including weight management~\citep{app4ath_pla_saracc2025evaluating}, resistance and jump training~\citep{app4ath_pla_washif2024artificial, app4ath_pla_havers2025reproducibility, pajo202510}, upper body and core training~\citep{app4ath_pla_canuzakov2025digital, app4ath_pla_erol2024does}, and nutritional strategies for ultra-endurance sports~\citep{app4ath_pla_puce2024optimizing, app4ath_pla_solomon2025sports}.
LLMs can also help trainers develop fitness programs for specific patient populations, including obese people~\citep{app4ath_pla_li2025gpt, philuek2025effects}, those with chronic diseases~\citep{xu2024assessment,app4ath_pla_onan2025examining, app4ath_pla_akrimi2025chatgpt}, and those with epilepsy~\citep{app4ath_pla_rocha2024potential, app4ath_pla_rocha2025can}.
In terms of methods, some studies employ digital twins with multimodal outputs~\citep{app4ath_pla_vahdati2025multi}, behavioral science theories~\citep{app4ath_pla_hegde2024infusing, app4ath_pla_jorke2025gptcoach, app4ath_pla_dindorf2025characteristics}, and RAG technology~\citep{app4ath_pla_zhang2025rag, app4ath_pla_ko2025legolas}; in terms of effectiveness, some emphasize the importance of personalization and contextual understanding~\citep{app4ath_pla_dergaa2024using, app4ath_pla_zhu2024could, app4ath_pla_han2025intent}, focus on acceptance, trust, and quality~\citep{app4ath_pla_duking2024chatgpt, app4ath_pla_wachholz2025acceptance}, and foster user self-reflection~\citep{app4ath_pla_li2025visualizing}.
In addition, some studies have designed AI coaches tailored to the specific requirements of individual sports, such as boxing~\citep{app4ath_pla_bullard2025enhancing} and table tennis~\citep{app4ath_pla_ma2025t3set, app4ath_pla_ma2025table}.
Within these sport-specific domains, two prominent tasks have emerged to provide professional-grade feedback.
One is \textit{expert commentary generation}, which utilizes MLLMs to provide evaluative insights and skill-level-aware feedback for basketball~\citep{app4ath_pla_seino2025expert} and soccer~\citep{app4ath_pla_ashutosh2025expertaf} based on video demonstrations.
The other is \textit{motion instruction generation}, where frameworks like MAAIG~\citep{app4ath_pla_yeh2023maaig} and the reference-based CoachMe~\citep{app4ath_pla_yeh2025coachme} automatically derive technical corrective guidance from 3D skeletal data to assist athletes in figure skating and boxing.


\noindent\textbf{Sports Injury and Rehabilitation.}
Diagnosing and treating sports injuries necessitates extensive interdisciplinary knowledge, and LLMs have demonstrated a broad understanding of this domain~\citep{app4ath_inj_hasnain2023role, app4ath_inj_lotfi2024evaluating}, encompassing orthopedics~\citep{app4ath_inj_fayed2023artificial} and sports rehabilitation~\citep{app4ath_inj_mcbee2023interdisciplinary, app4ath_inj_mcbee2024assessing}. 
Specifically, these models assist in providing preventive advice~\citep{app4ath_inj_zhu2025full}, identifying and labeling medical information~\citep{app4ath_inj_brogly2025evaluation}, and supporting diagnostic imaging~\citep{app4ath_inj_lotfi2024evaluating} and data processing~\citep{app4ath_inj_musat2024diagnostic}. 
Furthermore, they play a crucial role in clinical decision-making~\citep{app4ath_inj_saglam2025comparative}, surgical treatment planning~\citep{app4ath_inj_cheng2023artificial}, and enabling patient outcome prediction~\citep{app4ath_inj_ahsan2023chatbot} and medical oversight.


\noindent\textbf{Sports Psychology and Behavior.}
LLMs have demonstrated initial potential in this field, capable of answering sports-related questions~\citep{app4ath_psy_vandelanotte2023increasing}, assessing cognitive abilities~\citep{app4ath_psy_zuccolotto11}, and summarizing psychological theories~\citep{app4ath_psy_oliver2025generative}. 
A significant line of research integrates these models with wearable technology for real-time monitoring and behavioral modeling~\citep{ app4ath_psy_ferrara2024large, app4ath_psy_imran2024llasa, app4ath_psy_ji2024hargpt, app4ath_psy_merrill2024transforming}. 
Furthermore, LLMs provide assistance in specialized areas such as managing exercise addiction~\citep{app4ath_psy_szabo2023chatgpt} and facilitate behavioral interventions by enhancing motivation for sports participation~\citep{app4ath_psy_song2025investigating} and delivering sleep education~\citep{app4ath_psy_masur2025assessment}.

\subsection{Applications for Coaches and Educators}
\label{append_app:coa}



\noindent\textbf{Action Spotting and Recognition.}
In this task, the majority of approaches employ MLLMs to facilitate direct action spotting and recognition. 
Specific methodologies include keyframe sampling~\citep{app4coa_ana_kodathala2025sv3}, contrastive pretraining~\citep{app4coa_ana_shin2025soccer}, and domain adaptation~\citep{app4fans_com_jiang2025domain}, primarily focusing on soccer. 
Applications extend to other sports, with studies fine-tuning MLLMs for rally-sequence recognition in tennis~\citep{app4coa_ana_teo2025enhancing}, utilizing high-frame-rate modeling for gymnastics and diving~\citep{app4coa_ana_li2025improving}, and performing scene-level classification in rugby~\citep{app4coa_ana_nonaka2024rugby}. 
Beyond visual-centric approaches, textual signals such as commentary have also been leveraged for spotting tasks~\citep{app4coa_ana_chakraborty2025we}. 
Additionally, benchmarks like ActionAtlas~\citep{app4coa_ana_salehi2024actionatlas} and F³Set~\citep{app4coa_ana_liu2025f} provide platforms for evaluating fine-grained recognition capabilities.



\noindent\textbf{Sports Action Quality Assessment.}
Recent endeavors in this domain focus on fine-tuning large multimodal models to facilitate personalized fitness evaluation~\citep{app4coa_aqa_dibenedetto2025fine}. 
Researchers have also proposed unified agent frameworks tailored for open-set and user-specific assessments~\citep{app4coa_aqa_tang2025fitnessagent}, and established fine-grained datasets incorporating LLM-based evaluations for functional movement screening~\citep{app4coa_aqa_xing2025llm}. 
In specific sports such as figure skating, MLLMs have been used to quantify technical and program scores, providing critical support for both athlete training and referee judging~\citep{app4coa_aqa_wang2025beats}.



\noindent\textbf{Sports Tactics and Strategies.}
Early work in this field was largely text-centric, converting structured data into natural language for tactical modeling, such as fine-tuning LLMs on event sequences~\citep{app4coa_tac_caron2023tacticalgpt}, transforming cycling commentary into graph representations~\citep{app4coa_tac_janssens2024large}, or parsing play-by-play logs into spatial spray charts~\citep{app4coa_tac_michielssen2024using}. 
Subsequent studies have examined the analytical reasoning of LLMs, including computing team scores from play-by-play data~\citep{app4coa_tac_hu-etal-2024-sportsmetrics} and aggregating narratives for score inference~\citep{app4fans_nar_hu2024reasoning}. 
More recent approaches integrate diverse structural and spatial information for richer tactical reasoning, utilizing sketch-based LLM agents for interactive tactic design~\citep{app4coa_tac_liu2024smartboard}, graph LLMs for zero-shot generalization~\citep{app4coa_tac_lingrui2025tacticexpert}, and multi-agent systems that combine video detection with statistical inference~\citep{app4coa_tac_zhang2025chatmatch}.



\noindent\textbf{Game and Player Performance Prediction.}
Early research for this task predominantly utilized BERT-based models~\citep{devlin2019bert} to predict specific player actions or traits, such as forecasting badminton strokes from skeleton poses~\citep{app4coa_pre_ibh2024stroke} and analyzing NBA players' performance deviations based on pre-game interview transcripts~\citep{app4coa_pre_oved2020predicting}. 
More recent advancements leverage LLMs to synthesize diverse data sources for broader game outcome predictions. 
Specific applications include predicting basketball results via in-context learning on social media data~\citep{app4coa_pre_sprint2024social}, explaining handball match outcomes through feature attribution summarization~\citep{app4coa_pre_felice2024ai}, and fusing features from multimodal pre-match reports to enhance cricket score predictions~\citep{app4coa_pre_bhatnagar2025analyzing}.



\noindent\textbf{Sports Education.}
In this domain, LLMs are extensively applied to assist educators with lesson planning~\citep{app4coa_edu_gencc2023artificial, wang2024artificial}, designing interactive activities~\citep{app4coa_edu_cui2025innovating}, providing formative feedback~\citep{app4coa_edu_keiper2023artificial}, and automating assignment creation~\citep{app4coa_edu_kauppinen2024proactive}. 
Research also emphasizes curricular support through automated visualization, synthetic dataset creation~\citep{app4coa_edu_fazackerley2025harnessing}, and sensing-driven feedback mechanisms~\citep{app4coa_edu_gao2025motion}. 
Furthermore, student-centered applications focus on personalized exercise planning and mental health support~\citep{app4coa_edu_zhang2024using}, while other studies investigate the inclusion, trust, and acceptance of tools like ChatGPT in educational practice~\citep{app4coa_edu_chang2025exploring}.

\subsection{Applications for Referees}
\label{append_app:ref}



\noindent\textbf{Sports Refereeing.}
In this task, large models are leveraged to enhance fairness and transparency in decision-making. 
A prominent example is X-VARS~\citep{app4ref_ref_app4held2024x}, which introduces an explainable Video Assistant Referee system. 
By fine-tuning MLLMs on expert-annotated foul data, this system provides textual rationales alongside decisions, thereby demonstrating clear benefits in improving decision accuracy, consistency, and trust among referees~\citep{app4ref_ref_held2025enhancing}.

\subsection{Applications for Fans and Social Media}
\label{append_app:fan}

\noindent\textbf{Sports Commentary Generation.}
Recent studies leverage LLMs to automatically produce commentary, offering fans an enhanced viewing experience. 
Most approaches adopt an agentic framework, where LLMs are prompted with extracted match information such as detected key events ~\citep{app4fans_com_pavlovich2023soccer, app4fans_com_andrews2024aicommentator,app4fans_com_andrews2024designing,app4fans_com_sameer2025enhanced}, player and ball tracking data~\citep{app4fans_com_andrews2024aicommentator,app4fans_com_andrews2024designing,app4fans_com_vijayakumar2025player}, player background information~\citep{app4fans_com_mori2025live,app4fans_com_xi2025player}, audio signals~\citep{app4fans_com_gautam2022soccer}, and external knowledge~\citep{app4fans_com_li2025multi}. 
Beyond agentic frameworks, recent work explores end-to-end training to improve quality, either by fine-tuning MLLMs~\citep{app4fans_com_wang2024commentary, app4fans_com_cook2024llm,app4fans_com_baughman2024large,app4fans_com_jiang2025domain} or designing novel architectures for better temporal alignment~\citep{app4fans_com_zhang2024descriptive,app4fans_com_rao2024matchtime,app4fans_com_you2025timesoccer}. 
Additionally, efforts target complementary directions like constructing benchmarks~\citep{app4fans_com_chen2025livecc, app4fans_com_ge2024scbench}, enabling real-time streaming~\citep{app4fans_com_chen2025livecc, ding2025streammind, yu2025temporally}, supporting multilingual commentary~\citep{app4fans_com_sameer2025enhanced}, generating personalized narratives~\citep{app4fans_com_andrews2024designing}, and advancing commercial applications~\citep{app4fans_com_baughman2024large}.



\noindent\textbf{Sports Highlight Generation.}
Most research in this domain employs MLLMs to facilitate highlight generation via key event detection, incorporating techniques such as action spotting~\citep{app4fans_hig_banu2025survey} and multimodal fusion with textual encoding~\citep{app4fans_hig_davids2025sportsummarizer}. 
Some approaches explicitly leverage commentary transcripts or role-play prompting to enhance event classification accuracy~\citep{app4fans_hig_sattar2023multi,app4fans_hig_kang2025diamond}. 
Other works use MLLMs for summary and caption generation to support social media highlights~\citep{app4fans_hig_midoglu2024ai}, or for personalized highlight generation via simulated watch histories and preference descriptions~\citep{app4fans_hig_leehippo}.



\noindent\textbf{Sports News Generation.}
Early work primarily focused on summarizing unstructured text commentary, utilizing LLMs to select and rewrite key segments~\citep{app4fans_new_wang2021sportssum2, app4fans_new_wang2022knowledge} or to extract salient events~\citep{app4fans_new_sarkar2024advancing}.
Recent advancements have extended these capabilities to process structured data, employing chain-of-thought prompting to interpret statistical tables~\citep{app4fans_new_chiang2025tree} or leveraging CSV inputs to generate comprehensive game reports~\citep{app4fans_new_chiang2024badge}.
Beyond summarization, \citet{app4fans_new_cheng2024snil} propose an insight-driven approach where high-level user queries guide LLMs to construct narrative episodes enriched with data visualizations.



\noindent\textbf{Sports Narratives and Storytelling.}
Recent research in this field focuses on generating factually consistent highlight narrations through advanced prompt engineering techniques~\citep{app4fans_nar_sarfati2023generating}. 
Significant advancements have also been made in leveraging multimodal embedded visualizations and personalized narratives to facilitate tactical understanding for general audiences~\citep{app4fans_nar_lee2024sportify,app4fans_nar_lin2025sportsbuddy}. 
Furthermore, other works adapt narrative generation pipelines to platform-specific contexts, enabling the production of personalized reports and posts designed for large-scale fan interaction~\citep{app4fans_nar_sarkhoosh2024multimodal, app4fans_com_baughman2024large}.



\noindent\textbf{Public Opinion Analysis in Sports.}
In this field, LLMs have been applied to identify key discussion themes within large-scale social media datasets~\citep{app4fans_opi_qian2024esports}. 
Significant progress has been made in performing fine-grained sentiment and stance detection via aspect-based analysis~\citep{app4fans_opi_qian2025experience} and utilizing few-shot prompting to analyze controversial topics~\citep{app4fans_opi_rauchegger2024onelove}. 
Additionally, researchers employ social science frameworks combined with LLMs to examine user acceptance and perceptions of emerging AI tools within the sports community~\citep{app4fans_opi_argan23investigating}.

\noindent\textbf{Sports Models and Systems.}
LLMs have been extensively applied to build sports chatbots for interactive dialogue~\citep{app4fans_mod_priya2024megan} or co-viewing experiences~\citep{app4fans_mod_kim2025bleacherbot}, often incorporating dialogue state tracking for sports-specific contexts~\citep{app4fans_mod_song2025korean}. 
General sports models, particularly for soccer, are developed using diverse techniques including fine-tuning~\citep{app4fans_mod_unlu2023footgpt, app4fans_mod_gautam2025soccerchat, app4fans_mod_rao2025towards}, knowledge graph integration~\citep{chen2025finequest}, and multi-agent LLM architectures~\citep{app4fans_mod_rao2025multi}. 
In the area of search and retrieval, interactive agents combine LLMs with offline query understanding and online decision-making~\citep{app4fans_mod_karat2025system}. 
Furthermore, RAG systems are utilized to query sports knowledge from natural language sources~\citep{app4fans_mod_schilling2024querying,app4fans_mod_strand2024soccer,app4fans_mod_strand2024soccerrag,app4fans_mod_sepasdar2024enhancing,app4fans_mod_sepasdar2024soccer,app4fans_mod_wang2025agentic}. 
Extended applications also include online information retrieval for cricket~\citep{app4fans_mod_wickramasinghe2025assessing} and fine-grained video retrieval for sports such as gymnastics and diving~\citep{app4fans_mod_gupta2025play}.

\subsection{Applications for Researchers}
\label{append_app:res}



\noindent\textbf{Sports Academic Writing.}
In fields such as sports science and medicine, LLMs like ChatGPT are increasingly utilized to generate outlines, draft abstracts, and provide grammar and style suggestions~\citep{app4res_latzel2024artificial, app4res_hakam2024human}. 
However, the literature emphasizes the need for caution due to inherent risks in content accuracy~\citep{app4res_dergaa2023human}, the reliability of generated references~\citep{app4res_anderson2023ai}, calculation precision~\citep{app4res_methnani2023chatgpt}, and originality.

\subsection{Applications for the Sports Industry}
\label{append_app:ind}



\noindent\textbf{Sports Management.}
Large models are increasingly applied to streamline diverse functions within sports organizations. 
In financial management, research demonstrates their ability to conduct interviews, extract key themes, and develop tailored organizational strategies~\citep{app4ind_man_haghparast2024financial}. 
For database management, LLMs are used to structure and analyze complex club data to enhance operational efficiency~\citep{app4ind_man_merilehto2024pdfs}. 
In facility management, these models support human-computer dialogue to facilitate site selection and knowledge acquisition~\citep{app4ind_man_salimi2025comprehensive}. 
Furthermore, they are employed to simulate future industry scenarios and provide robust support for data-driven strategic decisions~\citep{app4ind_man_haghparast2025foresight}.



\noindent\textbf{Sports Talent Scouting.}
Large models enhance this domain by introducing more objective and data-driven methodologies for athlete evaluation~\citep{app4ind_tal_mateus2024empowering}. 
Recent research has deployed LLMs to analyze complex player datasets~\citep{app4ind_tal_raskar2025footyintel} and convert unstructured scouting reports into searchable, structured knowledge formats. 
Furthermore, some work combines large models with RAG strategies to optimize the integration of diverse information sources~\citep{app4ind_tal_martire2025leveraging}.



\noindent\textbf{Sports Tourism.}
Large models are increasingly leveraged in this domain to enhance intelligence and personalization, offering solutions for virtual guides, information assistants, and community building~\citep{app4ind_tou_memon2025ai}. 
Research also highlights the role of these models in improving operational efficiency and fan engagement during major events. 
Notably, LLMs also show strong potential in the specialized sector of esports tourism~\citep{app4ind_tou_yenisoy2025investigating}.
\section{More Details on Datasets for Large Models in Sports}
\label{app_sec:appendix_data}

In this section, we further provide a comprehensive introduction to the datasets for large models in sports, as an extension of the main discussion in Section~\ref{sec:data}. 

\subsection{Task-Specific Datasets}
\label{app:task_specific}

This subsection echoes Section~\ref{subsec:data_categorization} of the paper and provides a further overview of these task-specific datasets. Table~\ref{tab:appendix_task_table1} and Table~\ref{tab:appendix_task_table2} present the datasets related to specific tasks for 5 sports stakeholder groups: athletes and trainers, coaches and educators, referees, fans and social media, and the sports industry. The information covers dataset names, involved sports types, data modalities, methods used in the papers, corresponding large models, best achieved performance results with their respective evaluation metrics, and the availability of open-source links, which can be directly accessed by clicking in the table.

\noindent\textbf{These datasets are unevenly distributed across tasks.} From the perspective of target users, datasets for coaches and fans are relatively abundant, while those for referees, researchers, and the sports industry are relatively scarce. In terms of task types, early tasks in traditional computer vision and natural language processing, such as action spotting and recognition and sports commentary generation, have received more research attention and have richer datasets, whereas tasks like public opinion analysis in sports and sports talent scouting lack open-source data, reflecting an imbalance in scholarly focus across different tasks.

\noindent\textbf{These datasets are unevenly distributed across sports.} Popular sports such as soccer, basketball, and badminton receive more attention and have richer datasets, whereas niche sports like track and field, aquatics, and even esports are severely underrepresented. Nevertheless, these underexplored areas hold research value and warrant further expansion and investigation.

\noindent\textbf{These datasets are unevenly distributed across modalities.} Video and text are the most common modalities in sports datasets, while audio, sensor data (e.g., IMU), and skeletal data are relatively scarce. This reflects the current research focus on video in the sports domain and also highlights the untapped potential of other modalities.

\noindent\textbf{These datasets are generally used with pre-existing models rather than being used to train or fine-tune models.} Most researchers tend to rely on the inherent capabilities of large models, which explains the widespread use of powerful closed-source models such as GPT-4~\citep{achiam2023gpt4}. This trend reflects both the scarcity of sports data and the significant value of constructing dedicated sports datasets and models, emphasizing the need for more attention to the field of large models in sports.

\setlength{\tabcolsep}{2pt}
\begin{table*}[ht]
\centering
\fontsize{7pt}{9pt}\selectfont
\begin{tabular}{cccccccc}
\toprule
\textbf{Task}         & \textbf{Dataset} & \textbf{Sports}                   & \textbf{Modal}            & \textbf{Method}          & \textbf{Related Large Model} & \textbf{Performance}     & \textbf{Link} \\
\midrule
\multicolumn{8}{c}{\cellcolor{gray!10}\textit{\textbf{Athletes and Trainers}}}                                                                                                                                                \\
\midrule
\multirow{10}{*}{\textbf{PLA}} 
& YourSkatingCoach~\citeyearpar{chen2024yourskatingcoach} & Figure Skating                    & V, T                     & MAAIG~\citeyearpar{app4ath_pla_yeh2023maaig}                    & T5~\citeyearpar{raffel2020T5}                  & 22.08 (METEOR)           & \ding{55}             \\
& PACE~\citeyearpar{vardhan2022PACE}             & Fitness                 & T                            & \citet{app4ath_pla_hegde2024infusing}                        & LaMDA~\citeyearpar{thoppilan2022lamda}               & 3.78 ± 1.00 / 5.00 (Likert) &   \clickablecheckmark{https://github.com/fitllm/classifiers}            \\
& NSCA-CSCS~\citeyearpar{NSCA_CSCS_Exam}        & Fitness                           & T                            & PH-LLM~\citeyearpar{app4ath_pla_cosentinotowards}                   & Gemini Ultra 1.0~\citeyearpar{team2023gemini}    & 88.00 (Acc)               & \ding{55}             \\
                      & T3Set~\citeyearpar{app4ath_pla_ma2025t3set}            & Table Tennis                      & V, M, T                & SenseCoach~\citeyearpar{app4ath_pla_ma2025t3set}               & Llama 3.3-70B~\citeyearpar{dubey2024llama3}       & 51.64 (P@6-S1L)          &      \clickablecheckmark{https://github.com/jima-cs/T3Set}         \\
                      & SCD~\citeyearpar{app4ath_pla_han2025intent}              & Soccer                            & T                            & \citet{app4ath_pla_han2025intent}                        & BERT~\citeyearpar{devlin2019bert}    & 85.64 (BERTScore)        & \ding{55}             \\
                      & Custom Dataset~\citeyearpar{app4ath_pla_ma2025table}           & Table Tennis                      & V, I, S, T    & \citet{app4ath_pla_ma2025table}                        & GPT-4~\citeyearpar{achiam2023gpt4}               & 67.40 (Acc)               &     \clickablecheckmark{https://github.com/mwlsus/ttcs_by_llm}          \\
                      
                      & Ego-Exo4D~\citeyearpar{grauman2024egoexo4d}        & SC, BK, CL & V, T                     & ExpertAF~\citeyearpar{app4ath_pla_ashutosh2025expertaf}                 & Llama 3-8B~\citeyearpar{dubey2024llama3}          & 49.60 (METEOR)            &    \clickablecheckmark{https://github.com/thechargedneutron/ExpertAF}           \\
                      & Ego-Exo4D~\citeyearpar{grauman2024egoexo4d}        & Basketball                        & V, T                     & \citet{app4ath_pla_seino2025expert}                        & GPT-4o~\citeyearpar{hurst2024gpt4o}              & 25.60 (METEOR)            & \ding{55}             \\
                      
                      & FS~\citeyearpar{app4ath_pla_yeh2025coachme}               & Figure Skating                    & \multirow{2}{*}{S, T} & \multirow{2}{*}{CoachMe~\citeyearpar{app4ath_pla_yeh2025coachme}} & \multirow{2}{*}{T5~\citeyearpar{raffel2020T5} } & 26.5 (BERTScore)         &   \multirow{2}{*}{\clickablecheckmark{https://motionxperts.github.io/}}            \\
                      & BX~\citeyearpar{app4ath_pla_yeh2025coachme}               & Boxing                            &                                 &                          &                     & 36.9 (BERTScore)         &      \\ 
\specialrule{0.1pt}{1pt}{1pt} 
\multirow{3}{*}{\textbf{INJ}} & cMedQA2~\citeyearpar{zhang2018cmedqa2}        & - & T & \citet{app4ath_inj_zhu2025full} & Qwen2-0.5B~\citeyearpar{team2024qwen2} & 30.56 (BLEU-4)      & \ding{55} \\
                     & Custom Dataset~\citeyearpar{app4ath_inj_saglam2025comparative} & - & T & \citet{app4ath_inj_saglam2025comparative} & GPT-4~\citeyearpar{achiam2023gpt4}      & 47.80 (Cronbach’s $\alpha$) & \ding{55} \\
                     & Custom Dataset~\citeyearpar{app4ath_inj_brogly2025evaluation} & - & T & \citet{app4ath_inj_brogly2025evaluation} & phi-3-mini~\citeyearpar{abdin2024phi3} & 34.13 (Spearman's $\rho$)           & \ding{55} \\
\specialrule{0.1pt}{1pt}{1pt} 
\multirow{3}{*}{\textbf{PSY}} 
&Custom Dataset~\citeyearpar{app4ath_psy_merrill2024transforming} & Fitness & T    & PHIA~\citeyearpar{app4ath_psy_merrill2024transforming}   & Gemini 1.0 Ultra~\citeyearpar{team2023gemini} & 84.20 (Acc) & \clickablecheckmark{https://github.com/yahskapar/personal-health-insights-agent} \\
&Capture24~\citeyearpar{chan2021capture}      & Fitness & M    & HARGPT~\citeyearpar{app4ath_psy_ji2024hargpt} & GPT-4~\citeyearpar{achiam2023gpt4}            & 79.50 (F1) & \clickablecheckmark{https://github.com/aiot-lab/HARGPT} \\
&In-the-Wild~\citeyearpar{app4ath_psy_imran2024llasa}    & Fitness & M, T & LLaSA~\citeyearpar{app4ath_psy_imran2024llasa}  & Vicuna-7B~\citeyearpar{chiang2023vicuna}        & 79.95 (Acc)      & \clickablecheckmark{https://github.com/BASHLab/LLaSA} \\

\midrule
\multicolumn{8}{c}{\cellcolor{gray!10}\textit{\textbf{Coaches and Educators}}} \\
\midrule

\multirow{11}{*}{\textbf{ACT}}  
& Custom Dataset~\citeyearpar{app4coa_ana_nonaka2024rugby}   & Rugby                                    & I, T & \citet{app4coa_ana_nonaka2024rugby}                            & LLaVA-7B~\citeyearpar{liu2023llava}                          & 63.10 ± 2.20 (F1)        &  \ding{55}                 \\
                      & ActionAtlas v1.0~\citeyearpar{app4coa_ana_salehi2024actionatlas} & 56 Sports                                & V    & \citet{app4coa_ana_salehi2024actionatlas}                  & GPT-4o~\citeyearpar{hurst2024gpt4o}                            & 42.95 ± 2.91 (Acc)        &   \clickablecheckmark{https://github.com/mrsalehi/action-atlas?tab=readme-ov-file}                \\

& NSVA Subset~\citeyearpar{sportsvision2025nsva_subset}      & BK, AF            & V, T & SV3.3B~\citeyearpar{app4coa_ana_kodathala2025sv3}                       & Llama 3.2-3B~\citeyearpar{dubey2024llama3}                      & 85.60 ± 5.20 (BERT F1) &   \clickablecheckmark{https://huggingface.co/sportsvision/SV3.3B}                \\
                      & FineTennis~\citeyearpar{app4coa_ana_teo2025enhancing}       & Tennis                                   & V    & \citet{app4coa_ana_teo2025enhancing}                            & Video-LLaMA2-7B~\citeyearpar{cheng2024videollama2}                    & 76.00 (Edit Score)       &       \clickablecheckmark{https://github.com/bigcrushes/videollama2_tennis}            \\
                      & SoccerNet-v2~\citeyearpar{deliege2021soccernetV2}     & Soccer                                   & V    & \multirow{2}{*}{Soccer-CLIP~\citeyearpar{app4coa_ana_shin2025soccer}} & \multirow{2}{*}{ViT-B/32~\citeyearpar{dosovitskiy_vit}}         & 75.70 (t-AmAP)           & \multirow{2}{*}{\ding{55}} \\
                      & Tennis7~\citeyearpar{zhang2021tennis7}          & Tennis                                   & V    &                              &                                   & 93.80 (Acc)            &                   \\
                      
                      & F³Set~\citeyearpar{app4coa_ana_liu2025f}            & TN, BM, TT          & V, T & F³ED~\citeyearpar{app4coa_ana_liu2025f}                         & GPT-4~\citeyearpar{achiam2023gpt4}                             & 75.20 (F1$_{elm}$)            &      \clickablecheckmark{https://github.com/F3Set/F3Set}             \\
                      & Video-MME~\citeyearpar{fu2025videomme}        & SC,BK,GY,DV & V, T & F-16~\citeyearpar{app4coa_ana_li2025improving}                         & LLaVA-OV~\citeyearpar{lillavaonevision}              & 65.00 (Acc)              &   \clickablecheckmark{https://github.com/bytedance/F-16}                \\
                      & SoccerNet-v2~\citeyearpar{deliege2021soccernetV2}     & Soccer                                   & T    & \citet{app4coa_ana_chakraborty2025we}                            & Llama 3.1-8B~\citeyearpar{dubey2024llama3}                      & 64.50 (mAP)              &   \ding{55}                \\
                      & SoccerNet-v2~\citeyearpar{deliege2021soccernetV2}     & Soccer                                   & V, T & \citet{app4fans_com_jiang2025domain}           & LLaVA-NeXT-Video~\citeyearpar{li2024llavanext} & 63.50 (Acc)             & \ding{55} \\
    & UCI-HAR~\citeyearpar{anguita2013public} & Fitness & M & \citet{app4coa_edu_gao2025motion} & GPT-4~\citeyearpar{achiam2023gpt4} & 92.30 (Acc) & \ding{55} \\
& SoccerNet~\citeyearpar{giancola2018soccernetV1} & Soccer & V, A, T & \citet{app4fans_hig_banu2025survey} & Video-LLaMA~\citeyearpar{zhang2023videollama} & 87.00 (F1) & \ding{55} \\
\specialrule{0.1pt}{1pt}{1pt} 
\multirow{6}{*}{\textbf{AQA}} & Fis-V~\citeyearpar{xu2019fisv}       & \multirow{3}{*}{Figure Skating} & \multirow{3}{*}{V, A, T} & \multirow{3}{*}{\citet{app4coa_aqa_wang2025beats}} & \multirow{3}{*}{InternVL2~\citeyearpar{chen2024internvl_2}} & 84.00 (Spearman's $\rho$) &  \multirow{3}{*}{\clickablecheckmark{https://github.com/ycwfs/FigureSkating-Quality-Assessment}} \\
                     & FS1000~\citeyearpar{xia2023fs1000}      &                                 &                          &                    &                           & 90.00 (Spearman's $\rho$) &   \\
                     & FineFS~\citeyearpar{ji2023finefs}      &                                 &                          &                    &                           & 76.00 (Spearman's $\rho$) &   \\
                     & Fitness-AQA~\citeyearpar{parmar2022fitnessaqa} & Fitness                         & V, T                     & \citet{app4coa_aqa_dibenedetto2025fine}                  & LLaVA-Video-7B~\citeyearpar{zhang2025llavavideo}            & 22.82 (mAP)     & \clickablecheckmark{https://github.com/GaetanoDibenedetto/UMAP25}  \\
                     & FMS~\citeyearpar{app4coa_aqa_tang2025fitnessagent}         & Fitness                         & V, T                     & FitnessAgent~\citeyearpar{app4coa_aqa_tang2025fitnessagent}        & ChatGLM4~\citeyearpar{glm2024chatglm}                  & 39.34 (Acc)     & \ding{55} \\
                     & LLM-FMS~\citeyearpar{app4coa_aqa_xing2025llm}     & Fitness                         & V, T                     & \citet{app4coa_aqa_xing2025llm}      & -                         & 91.00 (Acc)        &  \\
\specialrule{0.1pt}{1pt}{1pt} 
                      & Custom Dataset~\citeyearpar{app4coa_tac_caron2023tacticalgpt}           & Soccer                        & T    & TacticalGPT~\citeyearpar{app4coa_tac_caron2023tacticalgpt}                             & GPT-NeoX-20B~\citeyearpar{black2022gptneox}   & 50.00 (Acc)          &\ding{55}  \\
                      & STATS SportVU~\citeyear{sportvu2025}       & Basketball                    & I, T & Smartboard~\citeyearpar{app4coa_tac_liu2024smartboard}                             & GPT-4V~\citeyearpar{openai2024gpt4v}         & -                 &  \ding{55}\\
                      & Custom Dataset~\citeyearpar{petti2021baseballr}      & Baseball                      & T    & \citet{app4coa_tac_michielssen2024using}                                      & Curie~\citeyearpar{brown2020gpt3}  & 97.00 (Acc)          &  \clickablecheckmark{https://github.com/tony-baseball/Hitting-Spray-Charts-with-ggplot-and-ggplotly}\\
                      & SportsMetrics~\citeyearpar{app4coa_tac_hu-etal-2024-sportsmetrics}       & BK, AF & T    & \citet{app4coa_tac_hu2024can,app4coa_tac_hu-etal-2024-sportsmetrics}                                      & Gemini-Pro~\citeyear{team2023gemini}     & 32.30 ($\Delta$GScore)    & \clickablecheckmark{https://github.com/YebowenHu/SportsMetrics} \\
                      & Custom Dataset~\citeyearpar{app4coa_tac_janssens2024large}      & Cycling                       & T    & \citet{app4coa_tac_janssens2024large}                                      & GPT-4o~\citeyearpar{hurst2024gpt4o}         & -                 & \ding{55} \\
                      & Custom Dataset~\citeyearpar{app4fans_nar_hu2024reasoning}      & Basketball                    & T    &  SportsGen~\citeyearpar{app4fans_nar_hu2024reasoning} & GPT-4o~\citeyearpar{hurst2024gpt4o}         & 98.41 (DnC-10)    & \clickablecheckmark{https://github.com/YebowenHu/SportsGen} \\
                      & Custom Dataset~\citeyearpar{app4coa_tac_zhang2025chatmatch}      & Badminton                     & V, T & ChatMatch~\citeyearpar{app4coa_tac_zhang2025chatmatch}                              & GPT-3.5-turbo~\citeyearpar{ouyang2022gpt3_5}  & 98.84 (Acc)       & \ding{55} \\
\multirow{-8}{*}{\textbf{TAC}} & Basketball-Instants~~\citeyear{vanzandycke2024deepsport} & Basketball                    & I, T & TacticExpert~\citeyearpar{app4coa_tac_lingrui2025tacticexpert}                           & Vicuna-7B-v1.5~\citeyearpar{chiang2023vicuna} & 83.33 (Macro F1) & \ding{55} \\

\specialrule{0.1pt}{1pt}{1pt} 

\multirow{6}{*}{\textbf{PRD}} & Custom Dataset~\citeyearpar{app4coa_pre_oved2020predicting} & Basketball & T                     & \citet{app4coa_pre_oved2020predicting}                             & BERT~\citeyearpar{devlin2019bert}                         & 58.50 (Acc)  & \ding{55} \\
                     & ShuttleSet~\citeyearpar{wang2023shuttleset}     & \multirow{2}{*}{Badminton}  & \multirow{2}{*}{V, T} & \multirow{2}{*}{RallyTemPose~\citeyearpar{app4coa_pre_ibh2024stroke}} & \multirow{2}{*}{BERT~\citeyearpar{devlin2019bert}}        & 54.30 (Acc)  & \multirow{2}{*}{\clickablecheckmark{https://github.com/MagnusPetersenTbh/RallyTempPose}}  \\
                     & BadmintonDB~\citeyearpar{ban2022badmintondb}    &   &                       &                               &                              & 62.80 (Acc)  &  \\
                     & Custom Dataset~\citeyearpar{app4coa_pre_sprint2024social} & Basketball & T                     & \citet{app4coa_pre_sprint2024social}                             & GPT-3.5-turbo~\citeyearpar{ouyang2022gpt3_5}                & 64.90 (Acc)  & \clickablecheckmark{https://github.com/gsprint23/DivisionIBasketballTwitter}  \\
                     & SportDevs~\citeyearpar{sportdevs2025handball}      & Handball   & T                     & \citet{app4coa_pre_felice2024ai}                             & Mistral-7B~\citeyearpar{jiang2023mistral7b}                   & 5.20 (RMSE) & \ding{55} \\
                     & Custom Dataset~\citeyearpar{app4coa_pre_bhatnagar2025analyzing} & Cricket    & V, T                  & \citet{app4coa_pre_bhatnagar2025analyzing}                             & GPT-4o mini~\citeyearpar{hurst2024gpt4o}, etc. & 86.30 (F1)   & \clickablecheckmark{https://bit.ly/iplscorepredictions} \\

\midrule
\multicolumn{8}{c}{\cellcolor{gray!10}\textit{\textbf{Referees}}} \\  
\midrule
\textbf{REF} & SoccerNet-XFoul~\citeyearpar{app4ref_ref_app4held2024x} & Soccer & V, T & X-VARS~\citeyearpar{app4ref_ref_app4held2024x, app4ref_ref_held2025enhancing} & Video-ChatGPT~\citeyearpar{maaz2024videochatgpt} & 3.80 / 5.00 (Likert) & \clickablecheckmark{https://github.com/heldJan/X-VARS} \\

\bottomrule
\end{tabular}

\caption{Summary of task-specific sports datasets related to large models, including athletes and trainers, coaches and educators, and referees. Task: PLA: exercise and training plans, INJ: sports injury and rehabilitation, PSY: sports psychology and behavior, ACT: action spotting and recognition, AQA: sports action quality assessment, TAC: sports tactics and strategies, PRD: game and player performance prediction, REF: sports refereeing. Sports: SC: soccer, BK: basketball, CL: sports climbing, AF: American football, TN: tennis, BM: badminton, TT: table tennis, GY: gymnastics, DV: diving. Modal: V: video, I: image, A: audio, S: skeleton data, M: IMU data, T: text.}

\vspace{-0.5cm}
\label{tab:appendix_task_table1}
\end{table*}

\setlength{\tabcolsep}{2pt}
\begin{table*}[ht]
\centering
\fontsize{6.97pt}{8.5pt}\selectfont
\begin{tabular}{cccccccc}
\toprule
\textbf{Task}         & \textbf{Dataset} & \textbf{Sports}                   & \textbf{Modal}            & \textbf{Method}          & \textbf{Related Large Model} & \textbf{Performance}     & \textbf{Link} \\
\midrule
\multicolumn{8}{c}{\cellcolor{gray!10}\textit{\textbf{Fans and Social Media}}}                                                                                                                                                \\
\midrule

\multirow{22}{*}{\textbf{CMT}} & Custom Dataset~\citeyearpar{app4fans_com_gautam2022soccer}                  & Soccer                      & V, A, T               & \citet{app4fans_com_gautam2022soccer}                         & GPT-3~\citeyearpar{brown2020gpt3}                                 & 0.31 (ROUGE-L)      & \clickablecheckmark{https://github.com/simula/soccer-summarization}                  \\
                      & SN-Caption-test-align~\citeyearpar{app4fans_com_rao2024matchtime}           & Soccer                      & V, T                  & MatchVoice~\citeyearpar{app4fans_com_rao2024matchtime}                & Llama 3~\citeyearpar{dubey2024llama3}                               & 42.00 (CIDEr)       & \clickablecheckmark{https://haoningwu3639.github.io/MatchTime/}                  \\
                      & LoL19~\citeyearpar{app4fans_com_wang2024commentary}                           & Esports                     & T                     & \citet{app4fans_com_wang2024commentary}                         & Llama 2 13B~\citeyearpar{touvron2023llama2}                            & -4.61 (BARTScore)   & \clickablecheckmark{https://github.com/ArnoZWang/esports-data-to-text}                  \\
                      & Custom Dataset~\citeyearpar{app4fans_com_andrews2024aicommentator}                  & Soccer                      & V, T                  & AiCommentator~\citeyearpar{app4fans_com_andrews2024aicommentator}             & GPT-3.5-turbo~\citeyearpar{ouyang2022gpt3_5}                         & 0.56 (Cohen’s d)    & \ding{55}                  \\
                      & CommentarySet~\citeyearpar{app4fans_com_ge2024scbench}                   & TF,SC,BK,GY,TT,TN      & V, T                  & \citet{app4fans_com_ge2024scbench}                         & InternVL-Chat-2~\citeyearpar{chen2024internvl_2}                       & 5.44 (SCORES)       & \ding{55}                  \\
                      & Custom Dataset~\citeyearpar{love2015bbc, secareanu2023football}                  & Soccer                      & T                     & LLM-Commentator~\citeyearpar{app4fans_com_cook2024llm}           & LLaMA 7B~\citeyearpar{touvron2023llama}                              & 92.00 (F1)          & \clickablecheckmark{https://github.com/paddelcourt/llm-sport-commentator}                  \\
                      & \multirow{3}{*}{Custom Dataset~\citeyearpar{app4fans_com_baughman2024large}} & Golf                        & \multirow{3}{*}{V, T} & \multirow{3}{*}{\citet{app4fans_com_baughman2024large}}        & Llama 2 7B~\citeyearpar{touvron2023llama2}                            & 99.12 (ROUGE-L)     & \multirow{3}{*}{\ding{55}} \\
                      &                                 & Tennis                      &                       &                           & Sandstone 3B~\citeyearpar{raffel2020T5}                          & 86.80 (ROUGE-L)     &                    \\
                      &                                 & American Football           &                       &                           & Llama 2 7B~\citeyearpar{touvron2023llama2}                            & 86.80 (ROUGE-L)     &                    \\
                      & BH-Commentary~\citeyearpar{app4fans_com_zhang2024descriptive}                   & Basketball                  & V, T                  & \citet{app4fans_com_zhang2024descriptive}                         & BERT~\citeyearpar{devlin2019bert}                                  & 12.19 (CIDEr)       & \clickablecheckmark{https://anonymous.4open.science/r/dataset-DC8E}                  \\
                      & SoccerNet-Caption~\citeyearpar{app4fans_com_mkhallati2023soccernet}               & Soccer                      & V, T                  & TimeSoccer~\citeyearpar{app4fans_com_you2025timesoccer}                & Llama 2 7B~\citeyearpar{touvron2023llama2}                            & 8.30 (CIDEr)        & \clickablecheckmark{https://vpx-ecnu.github.io/TimeSoccer-Website/}                  \\
                      & SoccerNet-V2~\citeyearpar{deliege2021soccernetV2}                    & \multirow{2}{*}{Soccer}     & \multirow{2}{*}{V, T}                  & \multirow{2}{*}{\citet{app4fans_com_jiang2025domain}}        & \multirow{2}{*}{Claude 3.5 Sonnet~\citeyearpar{anthropic2025claude35sonnet}} & 2.59 / 5.00 (Likert) & \multirow{2}{*}{\ding{55}} \\
                      & WyScout~\citeyearpar{hudl2024wyscout}                         &                             &                   &                           &                                       & 2.96 / 5.00 (Likert) &                    \\
                      & LFCBI~\citeyearpar{app4fans_com_mori2025live}                           & Soccer                      & V, T                  & \citet{app4fans_com_mori2025live}                         & GPT-4o~\citeyearpar{hurst2024gpt4o}                                & 15.50 (MSE)         & \clickablecheckmark{https://drive.google.com/drive/folders/1_EqBtLr9YCnRDlB4IS9p69PhnZffTmPx}                  \\
                      & SoccerTrack-Commentary~\citeyearpar{app4fans_com_vijayakumar2025player}          & Soccer                      & V, I, T               & \citet{app4fans_com_vijayakumar2025player}                         & GPT-3~\citeyearpar{brown2020gpt3}                                 & 33.84 (CIDEr)       & \ding{55}                  \\
                      & LiveSports-3K-CC~\citeyearpar{app4fans_com_chen2025livecc}                & 49 Sports                   & V, A, T               & LiveCC~\citeyearpar{app4fans_com_chen2025livecc}                    & Qwen2-VL-7B~\citeyearpar{wang2024qwen2vl}                           & 40.08 (Win Rate)    & \clickablecheckmark{https://showlab.github.io/livecc/}                  \\
                      & SoccerNet-v2~\citeyearpar{deliege2021soccernetV2}                    & Soccer                      & V, A, T               & SoccerComment~\citeyearpar{app4fans_com_li2025multi}             & Vicuna-7B-v1.5~\citeyearpar{chiang2023vicuna}                        & 36.58 (CIDEr)       & \ding{55}                  \\
                      & NBA-Identity~\citeyearpar{app4fans_com_xi2025player}                    & \multirow{2}{*}{Basketball} & \multirow{2}{*}{V, T} & \multirow{2}{*}{LLM-IAVC~\citeyearpar{app4fans_com_xi2025player}} & \multirow{2}{*}{Llama 3.2-3B~\citeyearpar{dubey2024llama3}}          & 105.30 (CIDEr)      & \multirow{2}{*}{\clickablecheckmark{https://github.com/Zeyu1226-mt/LLM-IAVC}} \\
                      & VC-NBA-2022~\citeyearpar{xi2025simple}                     &                             &                       &                           &                                       & 150.70 (CIDEr)      &                    \\
                      & Custom Dataset~\citeyearpar{app4fans_com_sameer2025enhanced}                  & Cricket                     & V, T                  & \citet{app4fans_com_sameer2025enhanced}                         & GPT-4o mini~\citeyearpar{hurst2024gpt4o}, etc.                     & 83.00 (BERT F1)     & \ding{55}                  \\
                      & SoccerNet-Caption~\citeyearpar{app4fans_com_mkhallati2023soccernet}               & Soccer                      & V, T                  & StreamMind~\citeyearpar{ding2025streammind}                & Video-LLaMA2-7B~\citeyearpar{cheng2024videollama2}                        & 82.04 (ROUGE-L)     & \clickablecheckmark{https://aka.ms/StreamMind}                  \\
                      & SoccerNet~\citeyearpar{giancola2018soccernetV1}                       & Soccer                      & V, T                  & VLM-TSI~\citeyearpar{yu2025temporally}                   & VideoLLM-Online~\citeyearpar{chen2024videollmonline}                       & 39.10 (TRACE)       &     \clickablecheckmark{https://github.com/yukw777/tglg}              \\
\specialrule{0.1pt}{1pt}{1pt} 
\multirow{6}{*}{\textbf{HLG}} & CricPulse~\citeyearpar{app4fans_hig_sattar2023multi}      & Cricket  & V, T                     & \citet{app4fans_hig_sattar2023multi}                                 & BERT~\citeyearpar{devlin2019bert}                        & 97.00 (F1)  & \ding{55}                  \\
& Custom Dataset~\citeyearpar{app4fans_hig_midoglu2024ai}  & Soccer   & V, A                     & SmartCrop~\citeyearpar{app4fans_hig_midoglu2024ai}                        & GPT-4~\citeyearpar{achiam2023gpt4}                       &   -        & \ding{55}                   \\
                     
                     & Custom Dataset~\citeyearpar{app4fans_hig_kang2025diamond} & Baseball & T                        & DIAMOND~\citeyearpar{app4fans_hig_kang2025diamond}                          & Mistral-Large~\citeyearpar{mistral_large_instruct_2411} & 76.50 (F1)  & \ding{55}                  \\
                     & HIPPO-Video~\citeyearpar{app4fans_hig_leehippo}    & -        & V, T                     & HiPHer~\citeyearpar{app4fans_hig_leehippo}                           & GPT-4~\citeyearpar{achiam2023gpt4}                       & 76.60 (mAP) & \clickablecheckmark{https://github.com/jeongeunnn-e/HIPPO-Video}                  \\
                     & Custom Dataset~\citeyearpar{app4fans_hig_davids2025sportsummarizer} & Cricket  & \multirow{2}{*}{V, A, T} & \multirow{2}{*}{SportSummarizer~\citeyearpar{app4fans_hig_davids2025sportsummarizer}} & \multirow{2}{*}{DistilBERT~\citeyearpar{sanh2019distilbert}} & 0.93 (HD)   & \multirow{2}{*}{\ding{55}} \\
                     & SoccerNet~\citeyearpar{giancola2018soccernetV1}      & Soccer   &                          &                                  &                             & 0.92 (HD)   &                   \\
\specialrule{0.1pt}{1pt}{1pt} 

\multirow{10}{*}{\textbf{NSG}} & SportsSum2.0~\citeyearpar{app4fans_new_wang2021sportssum2}   & Soccer     & \multirow{2}{*}{T} & \multirow{2}{*}{\citet{app4fans_new_wang2021sportssum2}}              & \multirow{2}{*}{RoBERTa~\citeyearpar{liu2019roberta}, etc.} & 47.78 (ROUGE-L)        & \multirow{2}{*}{\clickablecheckmark{https://github.com/krystalan/SportsSum2.0}} \\
                      & SportsSum~\citeyearpar{huang2020generating}      & Soccer     &                    &                                 &                                & 47.49 (ROUGE-L)        &                    \\
                  & K-SportsSum~\citeyearpar{app4fans_new_wang2022knowledge}    & Soccer     & \multirow{2}{*}{T} & \multirow{2}{*}{KES~\citeyearpar{app4fans_new_wang2022knowledge}}            & \multirow{2}{*}{mT5~\citeyearpar{xue2021mt5}}           & 47.17 (ROUGE-L)        & \multirow{2}{*}{\clickablecheckmark{https://github.com/krystalan/K-SportsSum}} \\
                      & SportsSum~\citeyearpar{huang2020generating}      & Soccer     &                    &                                 &                                & 47.79 (ROUGE-L)        &                    \\
                      & NBA API~\citeyearpar{patel2022nbaapi}        & Basketball & T                  & SNIL~\citeyearpar{app4fans_new_cheng2024snil}                            & GPT-3.5~\citeyearpar{ouyang2022gpt3_5}                        & 63.00 (Acc)          & \clickablecheckmark{https://github.com/13Lychee/SNIL}                  \\
                      & ShuttleSet~\citeyearpar{wang2023shuttleset}     & Badminton  & T                  & BADGE~\citeyearpar{app4fans_new_chiang2024badge}                           & GPT-4~\citeyearpar{achiam2023gpt4}                          & 8.63 / 10.00 (LLM) & \clickablecheckmark{https://github.com/AndyChiangSH/BADGE}                  \\
                      & Custom Dataset~\citeyearpar{app4fans_new_sarkar2024advancing} & Cricket    & T                  & \citet{app4fans_new_sarkar2024advancing}                               & Google Gemini~\citeyearpar{team2023gemini}                  & 9.20 / 10.00 (ACS)          & \ding{55}                  \\
                      & RotoWire~\citeyearpar{wiseman2017ROTO}       & Basketball & \multirow{3}{*}{T} & \multirow{3}{*}{Tree-of-Report~\citeyearpar{app4fans_new_chiang2025tree}} & \multirow{3}{*}{GPT-4o mini~\citeyearpar{hurst2024gpt4o}}   & 54.92 (CS F1)          & \multirow{3}{*}{\ding{55}} \\
                      & MLB~\citeyearpar{puduppully2019MLB}            & Baseball   &                    &                                 &                                & 62.99 (CS F1)          &                    \\
                      & ShuttleSet+~\citeyearpar{app4fans_new_chiang2025tree}    & Badminton  &                    &                                 &                                & 93.94 (CS F1)          &                   \\
\specialrule{0.1pt}{1pt}{1pt} 
\multirow{4}{*}{\textbf{NAR}} & Custom Dataset~\citeyearpar{app4fans_nar_sarfati2023generating} & Soccer     & T    & \citet{app4fans_nar_sarfati2023generating}           & T5-large~\citeyearpar{raffel2020T5}    & 49.04 (ROUGE-L) & \ding{55} \\
                     & SportsVU~\citeyearpar{app4fans_nar_lee2024sportify}       & Basketball & V, T & Sportify~\citeyearpar{app4fans_nar_lee2024sportify}    & -           & 72.22 (Acc)     & \clickablecheckmark{https://chungyi347.github.io/Sportify/} \\
                     & SoccerSum~\citeyearpar{app4fans_nar_sarkhoosh2024multimodal, app4fans_com_sarkhoosh2024soccersum}      & Soccer     & V, A & SoccerSum~\citeyearpar{app4fans_nar_sarkhoosh2024multimodal}   & GPT-4 Turbo~\citeyearpar{achiam2023gpt4} & -               & \clickablecheckmark{https://github.com/simula/soccersum} \\
                     & Custom Dataset~\citeyearpar{app4fans_nar_lin2025sportsbuddy} & Basketball & V    & SportsBuddy~\citeyearpar{app4fans_nar_lin2025sportsbuddy} & GPT-4o~\citeyearpar{hurst2024gpt4o}      & 90.80 (Acc)      & \ding{55} \\
\specialrule{0.1pt}{1pt}{1pt} 
\multirow{2}{*}{\textbf{OPI}} & Custom Dataset~\citeyearpar{app4fans_opi_rauchegger2024onelove} & Soccer & T & \citet{app4fans_opi_rauchegger2024onelove}    & GPT-4-turbo~\citeyearpar{achiam2023gpt4} & 70.30 (F1) & \ding{55} \\
& Custom Dataset~\citeyearpar{app4fans_opi_qian2025experience} & Soccer & T & ABSA~\citeyearpar{app4fans_opi_qian2025experience} & RoBERTa~\citeyearpar{liu2019roberta}     & 80.00 (F1)  & \clickablecheckmark{https://github.com/TyrealQ/Experience-is-all-you-need} \\

\specialrule{0.1pt}{1pt}{1pt} 
\multirow{17}{*}{\textbf{MOD}} & Custom Dataset~\citeyearpar{app4fans_mod_schilling2024querying}        & Soccer                  & T                     & \citet{app4fans_mod_schilling2024querying}                            & GPT-3.5~\citeyearpar{ouyang2022gpt3_5}                            & 71.40 (Acc)                  & \ding{55}                  \\
                      & SoccerNet~\citeyearpar{giancola2018soccernetV1}             & Soccer                  & V, A, I               & SoccerRAG~\citeyearpar{app4fans_mod_strand2024soccer, app4fans_mod_strand2024soccerrag}                    & GPT-4~\citeyearpar{achiam2023gpt4}, etc.                        & 80.00 (Acc)                  & \clickablecheckmark{https://github.com/simula/soccer-rag}                  \\
                      & Custom Dataset~\citeyearpar{app4fans_mod_karat2025system}        & Soccer                  & T                     & \citet{app4fans_mod_karat2025system}                            & GPT-4o~\citeyearpar{hurst2024gpt4o}                             & 88.25 (Precision)            & \ding{55}                  \\
                      & Custom Dataset~\citeyearpar{app4fans_mod_wickramasinghe2025assessing}        & Cricket                 & T                     & \citet{app4fans_mod_wickramasinghe2025assessing}                            & Copilot~\citeyearpar{chen2021copilot}                            & 100.00 (Acc)                 & \ding{55}                  \\
                      & TF-CoVR~\citeyearpar{app4fans_mod_gupta2025play}               & GY, DV      & V, T                  & TF-CoVR-Base~\citeyearpar{app4fans_mod_gupta2025play}                 & BLIP~\citeyearpar{li2022blip}                               & 23.02 (mAP@10)               & \clickablecheckmark{https://github.com/UCF-CRCV/TF-CoVR}                  \\
                      & KICK~\citeyearpar{app4fans_mod_song2025korean}                  & Soccer                  & T                     & \citet{app4fans_mod_song2025korean}                            & GPT-4o~\citeyearpar{hurst2024gpt4o}                             & 15.86 (JGA)                  & \clickablecheckmark{https://github.com/ezzy4me/KICK}                  \\
                      & SoccerNet-XFoul~\citeyearpar{app4ref_ref_app4held2024x}       & \multirow{2}{*}{Soccer} & \multirow{2}{*}{V, T} & \multirow{2}{*}{SoccerChat~\citeyearpar{app4fans_mod_gautam2025soccerchat}}  & \multirow{2}{*}{Qwen2-VL-7B~\citeyearpar{wang2024qwen2vl}}       & 6.81 / 10.00 (LLM)       & \multirow{2}{*}{\clickablecheckmark{https://github.com/simula/SoccerChat}} \\
                      & SoccerNet-v2~\citeyearpar{deliege2021soccernetV2}          &                         &                       &                              &                                    & 6.42 / 10.00 (LLM)       &                    \\
                      & SoccerBench~\citeyearpar{app4fans_mod_rao2025multi}           & Soccer                  & V, A, T               & SoccerAgent~\citeyearpar{app4fans_mod_rao2025multi}                  & DeepSeek-v3~\citeyearpar{liu2024deepseekv3}                        & 60.90 (Acc)                  & \clickablecheckmark{https://github.com/jyrao/SoccerAgent}                  \\
                      & SoccerNet-v2~\citeyearpar{deliege2021soccernetV2}          & \multirow{3}{*}{Soccer} & \multirow{3}{*}{V, T} & \multirow{3}{*}{MatchVision~\citeyearpar{app4fans_mod_rao2025towards}} & \multirow{3}{*}{Llama 3-8B~\citeyearpar{dubey2024llama3}}        & 80.10 (Acc)                  & \multirow{3}{*}{\clickablecheckmark{https://jyrao.github.io/UniSoccer/}} \\
                      & SN-Caption-test-align~\citeyearpar{app4fans_com_rao2024matchtime} &                         &                       &                              &                                    & 44.18 (CIDEr)                &                    \\
                      & MVFoul~\citeyearpar{held2023vars}                &                         &                       &                              &                                    & 44.00 (Acc)                  &                    \\
                      & SoccerNet-V2~\citeyearpar{deliege2021soccernetV2}          & \multirow{2}{*}{Soccer} & \multirow{2}{*}{V, T} & \multirow{2}{*}{\citet{app4fans_com_jiang2025domain}}           & \multirow{2}{*}{LLaVA-NeXT-Video~\citeyearpar{li2024llavanext}}  & 83.76 (Acc)                  & \multirow{2}{*}{\ding{55}} \\
                      & WyScout~\citeyearpar{hudl2024wyscout}               &                         &                       &                              &                                    & 81.83 (Acc)                  &                    \\
                      & Gym-QA~\citeyearpar{chen2025finequest}                & Gymnastics              & V, T                  & \multirow{3}{*}{FineQuest~\citeyearpar{chen2025finequest}}   & \multirow{3}{*}{Video-LLaVA~\citeyearpar{lin2024videollava}, etc.} & \multirow{2}{*}{57.00 (Acc)} & \multirow{3}{*}{\ding{55}} \\
                      & Diving-QA~\citeyearpar{chen2025finequest}             & Diving                  & V, T                  &                              &                                    &                              &                    \\
                      & SPORTU~\citeyearpar{xiasportu}                & 7 Sports                & V, T                  &                              &                                    & 73.20 (Acc)                  &                   \\
\midrule
\multicolumn{8}{c}{\cellcolor{gray!10}\textit{\textbf{The Sports Industry}}}                                                                                                                                                \\
\midrule
\textbf{MAN}                  & Custom Dataset~\citeyearpar{app4ind_man_merilehto2024pdfs} & -      & T    & \citet{app4ind_man_merilehto2024pdfs}          & Claude 3 Opus~\citeyearpar{anthropic2024claude3opus} & 90.28 (Acc) & \ding{55}         \\
\specialrule{0.1pt}{1pt}{1pt} 
\multirow{2}{*}{\textbf{TSC}} & Custom Dataset~\citeyearpar{app4ind_tal_martire2025leveraging} & Soccer & T    & \citet{app4ind_tal_martire2025leveraging}          & GPT-4o~\citeyearpar{hurst2024gpt4o}        & 3.80 / 5.00 (Likert) & \ding{55} \\
                     & Custom Dataset~\citeyearpar{app4ind_tal_raskar2025footyintel} & Soccer & I, T & Footyintel~\citeyearpar{app4ind_tal_raskar2025footyintel} & -             & - & \ding{55}                  \\
                     
\bottomrule
\end{tabular}


\caption{Summary of task-specific sports datasets related to large models, including fans and social media, and the sports industry. Task: CMT: sports commentary generation, HLG: sports highlight generation, NSG: sports news generation, NAR: sports narratives and storytelling, OPI: public opinion analysis in sports, MOD: sports models and systems, MAN: sports management, TSC: sports talent scouting. Sports: TF: track and field, SC: soccer, BK: basketball, TN: tennis, TT: table tennis, GY: gymnastics, DV: diving. Modal: V: video, I: image, A: audio, T: text.}

\vspace{-0.5cm}
\label{tab:appendix_task_table2}
\end{table*}
\subsection{Sports Understanding Datasets}
\label{app:sports_understanding}

This subsection echoes Section~\ref{subsec:data_categorization} and provides a more detailed overview of datasets related to sports understanding tasks for large models. We cover datasets specifically designed for sports understanding with large models (§\ref{app_subsec:spec}), general video understanding datasets that include sports content (§\ref{app_subsec:general}), and other general-purpose datasets containing sports-related data (§\ref{app_subsec:other}).
Table~\ref{tab:appendix_su_table} presents a comprehensive summary of the first two categories of datasets from multiple perspectives, including dataset names, covered sports types, data sources, annotation methods, benchmark availability, input modalities, the number and average duration of videos, the number of QA pairs, and open-source links, which are directly accessible by clicking.

\subsubsection{Specialized Sports Understanding Datasets}
\label{app_subsec:spec}

Recently, numerous datasets have been developed to evaluate and enhance the general sports understanding capabilities of large models.

For LLMs, QASports~\citep{jardim2023qasports} introduced the first large-scale sports question-answering dataset with rich contextual information and diverse questions for model training and evaluation. The sports understanding subtask in BIG-bench~\citep{srivastava2023bigbench} includes 986 binary-choice questions, primarily testing models’ general understanding of sports activities. SportQA~\citep{xia2024sportqa} comprises over 70,000 multiple-choice questions across three difficulty levels, enabling a comprehensive evaluation of LLMs’ performance in sports understanding. SPORTU-text~\citep{xiasportu} and FSBench-Text~\citep{gao2025fsbench} assess models’ understanding of rules, events, and scenarios in 5 major sports and figure skating, respectively.

For MLLMs, Sports-QA~\citep{li2024sports_qa} is the first dataset specifically designed for sports video question answering, advancing the evaluation of multimodal models in sports video understanding. SPORTU-video~\citep{xiasportu} covers 7 sports and provides systematic video understanding tasks across three difficulty levels, while Sports-3K-QA~\citep{app4fans_com_chen2025livecc} includes a broader range of 49 different sports.
FSAnno~\citep{gao2025fsbench} constructs a large-scale, multi-task, multimodal figure skating dataset, while FSBench-Motion~\citep{gao2025fsbench} extends it by adding motion data and QA pairs, supporting tasks ranging from single-action analysis to full-performance commentary. FineBadminton~\citep{he2025finebadminton} is a large-scale badminton video dataset with fine-grained annotations, on which FBBench~\citep{he2025finebadminton} evaluates models’ fine-grained sports video understanding. Gym-QA and Diving-QA~\citep{chen2025finequest}, built upon FineGym~\citep{shao2020finegym} and FineDiving~\citep{xu2022finediving}, respectively, offer new benchmarks for sports video question answering in gymnastics and diving.

\subsubsection{General Video Understanding Datasets Featuring Sports}
\label{app_subsec:general}


In addition to datasets specifically designed for sports understanding, many general video understanding datasets also include sports content, in which sports constitute an important component.

In addition to large-scale, multi-task, and comprehensive video understanding datasets~\citep{fu2025videomme, hemmworld, yang2025wildvideo}, some focus on specific capabilities. For example, InternVid~\citep{wanginternvid}, FIOVA~\citep{hu2024fiova}, and VidText~\citep{yang2025vidtext} are primarily used for video description or subtitle generation, while Ego-Exo4D~\citep{grauman2024egoexo4d} and EgoExoBench~\citep{he2025egoexobench} focus on video understanding from different viewpoints. LVBench~\citep{wang2024lvbench}, MLVU~\citep{zhou2025mlvu}, Neptune~\citep{nagrani2024neptune}, LongVILA\_sft~\citep{chenlongvila}, and VRBench~\citep{yu2025vrbench} are dedicated to long video understanding, while E.T. Bench~\citep{liu2024ETbench}, MotionBench~\citep{hong2025motionbench}, and ExAct~\citep{yi2025exact} are used for fine-grained action, skill, or motion understanding. TUNA~\citep{kong2025tuna} and VideoA11y-40K~\citep{li2025videoa11y} emphasize temporal information and dynamic video understanding, while VideoVista~\citep{li2024videovista}, V-STaR~\citep{cheng2025vstar}, MINERVA~\citep{nagrani2025minerva}, VRBench~\citep{yu2025vrbench}, and CausalStep~\citep{li2025causalstep} target video reasoning tasks such as temporal-spatial, multi-step, and causal reasoning.

Furthermore, video-SALMONN-2~\citep{tang2024videosalmonn2_1, tang2025videoSALMONN2_2}, WorldSense~\citep{hong2025worldsense}, HarmonySet~\citep{zhou2025harmonyset}, and MAVERIX~\citep{xie2025maverix} focus on joint understanding of audio and video, demonstrating multimodal capabilities; while OVO-Bench~\citep{niu2025ovobench} and RTV-Bench~\citep{xun2025rtvbench} examine the real-time processing capabilities of models. In terms of model capability evaluation, Trust-videoLLMs~\citep{wang2025Trust_videoLLMs} is used to evaluate the credibility of video understanding, while SIV-Bench~\citep{kong2025sivbench} studies the understanding of social interaction behaviors in videos.

\subsubsection{Other General Datasets Featuring Sports}
\label{app_subsec:other}

Moreover, some other types of general datasets also contain sports content. To evaluate the image understanding capabilities of large models, MDI-Benchmark~\citep{zhang2024mdibench} collected 514 real images and 1,298 question-answer pairs to test basic perception and complex reasoning, and designed sports-related questions for different age groups. MIP-GAF~\citep{madan2025mipgaf} constructed a large dataset to examine the understanding of key figures in images, which also includes sports scenes. Furthermore, to assess the ability of large models as multimodal search engines, MMSearch~\citep{jiangmmsearch} collected 300 unimodal and multimodal samples, and MomentSeeker~\citep{yuan2025momentseeker} constructed a dataset consisting of 268 long videos with an average length of over 1,200 seconds, all of which focus on sports scenes.

\setlength{\tabcolsep}{2pt}
\begin{table*}[ht]
\centering
\fontsize{7.1pt}{9.5pt}\selectfont

\begin{tabular}{lccccccccc}
\toprule
\textbf{Dataset}          & \textbf{Sports}            & \textbf{Source} & \textbf{Annotation}     & \textbf{Benchmark} & \textbf{Modal}                & \textbf{\# Video}     & \textbf{Avg. Length}        & \textbf{\# QA} & \textbf{Link}     \\
\midrule
\multicolumn{10}{c}{\cellcolor{gray!10}\textit{\textbf{Specialized Sports Understanding Datasets}}}                                                                                                                                                                               \\
\midrule
\textbf{QASports}~\citeyearpar{jardim2023qasports}         & SC, BK, AF                 & Fandom  & auto                     & \ding{55}                  & T                                & -                     & -                           & $\sim$1500K    & \clickablecheckmark{https://osf.io/n7r23/}                  \\
\textbf{BIG-bench-SU}~\citeyearpar{srivastava2023bigbench}     & SC, BK, AF, BB, IH         & program  & crowd                    & \ding{51}                  & T                                & -                     & -                           & 986            &      \clickablecheckmark{https://github.com/google/BIG-bench/tree/main/bigbench/benchmark_tasks/sports_understanding}                   \\
\textbf{SportQA-Level-1}~\citeyearpar{xia2024sportqa}       & -                          & existing (dataset) & manual                       & \ding{51}                  & T                                & -                     & -                           & 21385          & \multirow{3}{*}{\clickablecheckmark{https://github.com/haotianxia/SportQA}} \\
\textbf{SportQA-Level-2}~\citeyearpar{xia2024sportqa}       & 35 Sports                  & Wikipedia  & expert                  & \ding{51}                  & T                                & -                     & -                           & 45685          &                         \\
\textbf{SportQA-Level-3}~\citeyearpar{xia2024sportqa}       & SC, BK, TN, AF, TT, VB     & expertise    & expert                    & \ding{51}                  & T                                & -                     & -                           & 3522           &                         \\
\textbf{SPORTU-text}~\citeyearpar{xiasportu}      & SC, BK, TN, AF, VB         & existing    & expert                   & \ding{51}                  & T                                & -                     & -                           & 900            & \multirow{2}{*}{\clickablecheckmark{https://github.com/chili-lab/SPORTU}} \\
\textbf{SPORTU-video}~\citeyearpar{xiasportu}     & SC,BK,BM,TN,BB,VB,IH & competition     & expert             & \ding{51}                  & V, T                         & 1701                  & -                           & 12048          &                         \\
\textbf{Sports-3K-QA}~\citeyearpar{app4fans_com_chen2025livecc} & 49 Sports & YouTube & manual & \ding{51} & V, T  & 412 & -  & 1174 & 
\clickablecheckmark{https://showlab.github.io/livecc/} \\
\textbf{FSAnno}~\citeyearpar{gao2025fsbench}           & \multirow{3}{*}{FS}        & \multirow{3}{*}{competition} & \multirow{3}{*}{expert} & \ding{55}                  & V, A, T                  & 783                   & \multirow{3}{*}{$\sim$3.5m} & -              & \multirow{3}{*}{\ding{55}}                       \\
\textbf{FSBench-Text}~\citeyearpar{gao2025fsbench}     &                            &          &                    & \ding{51}                  & T                                & -                     &                             & 500            & \ding{55}                       \\
\textbf{FSBench-Motion}~\citeyearpar{gao2025fsbench}   &                            &       &                       & \ding{51}                  & V, T                         & 783                    &                             & 3500           &                        \\
\textbf{FineBadminton}~\citeyearpar{he2025finebadminton}           & \multirow{2}{*}{BM}        & \multirow{2}{*}{YouTube} & \multirow{2}{*}{manual}     & \ding{55}                  & V, T                         & 3215 & 12.4s      & -              & \multirow{2}{*}{\clickablecheckmark{https://github.com/FineBadminton/FineBadminton} }       \\
\textbf{FBBench}~\citeyearpar{he2025finebadminton}          &                            &       &                       & \ding{51}                  & V, T                         &          2563             &     -                        & 2563           &                        \\
\textbf{Gym-QA}~\citeyearpar{chen2025finequest}           & GY                         & existing  & manual                      & \ding{51}                  & V, T                         & 6031                  & -                           & 27469          & \ding{55}                       \\
\textbf{Diving-QA}~\citeyearpar{chen2025finequest}        & DV                         & existing   & manual                     & \ding{51}                  & V, T                         & $\sim$100             & -                           & 1055           & \ding{55}                       \\
\textbf{Sports-QA}~\citeyearpar{li2024sports_qa}        & SC, BK, GY, VB             & existing    & manual          & \ding{51}                  & V, T                         & 5967                  & 20.9s                       & 94073          & \clickablecheckmark{https://github.com/HopLee6/Sports-QA}                  \\
\midrule
\multicolumn{10}{c}{\cellcolor{gray!10}\textit{\textbf{General Video Understanding Datasets}}}                                                                                                                                                                        \\

\midrule
\textbf{InternVid}~\citeyearpar{wanginternvid}        & -                          & YouTube    & auto                  & \ding{55}                  & V, A, T                  & 7.1M                  & 6.4m                        & N/A            &            \clickablecheckmark{https://github.com/OpenGVLab/InternVideo/tree/main/Data/InternVid}             \\
\textbf{Ego-Exo4D}~\citeyearpar{grauman2024egoexo4d}        & SC, BK, CL                 & field   & expert                    & \ding{55}                  & V, A                        & 5035                  & 2.6m                        & N/A            &     \clickablecheckmark{https://ego-exo4d-data.org/}                    \\
\textbf{E.T. Bench}~\citeyearpar{liu2024ETbench}       & SC, BK, TN, CR, etc.       & existing    & manual                    & \ding{51}                  & V, T                         & 7002                  & 129s                        & 7289           & \clickablecheckmark{https://polyu-chenlab.github.io/etbench/}                  \\
\textbf{VideoVista}~\citeyearpar{li2024videovista}       & SC, etc.                   & existing   & auto                     & \ding{51}                  & V, A, T                  & 894                   & 131s                        & 24906          &   \clickablecheckmark{https://github.com/HITsz-TMG/VideoVista}                      \\
\textbf{FIOVA}~\citeyearpar{hu2024fiova}            & BB, etc.                   & -  & manual                            & \ding{51}                  & V                               & 3002                  & 33.6s                       & N/A            &    \clickablecheckmark{https://huggingface.co/datasets/huuuuusy/FIOVA}                     \\
\textbf{Neptune}~\citeyearpar{nagrani2024neptune}          & BK, etc.                   & existing   & manual                       & \ding{51}                  & V, A, T                  & 2405                  & 2.5m                        & 3268           &   \clickablecheckmark{https://github.com/google-deepmind/neptune}                      \\
\textbf{video-SALMONN2}~\citeyearpar{tang2025videoSALMONN2_2}  & BB, etc.                   & -        & manual                      & \ding{51}                  & V, A, T                  & 483                   & 51s                         & N/A            &         \clickablecheckmark{https://huggingface.co/datasets/videoSALMONN2/video-SALMONN_2_testset}                \\
\textbf{LVBench}~\citeyearpar{wang2024lvbench}          & BK, etc.                   & YouTube  & manual                      & \ding{51}                  & V, T                         & 103                   & 4101s                       & 1549           &    \clickablecheckmark{https://lvbench.github.io/}                     \\
\textbf{MMWorld}~\citeyearpar{hemmworld}          & SC, BK, GY, VB             & existing    & manual                      & \ding{51}                  & V, A, T                  & 1910                  & $\sim$105s                  & 6627           &   \clickablecheckmark{https://huggingface.co/datasets/Xuehai/MMWorld}                      \\
\textbf{LongVILA\_sft}~\citeyearpar{chenlongvila}    & -                          & existing     & manual                     & \ding{55}                  & V, T                         & 15292                 & -                           & 15292          &     \clickablecheckmark{https://huggingface.co/datasets/LongVILA/longvila_sft_dataset}                    \\
\textbf{MLVU}~\citeyearpar{zhou2025mlvu}             & SC, BK, BM, TT, VB         & -       & manual                       & \ding{51}                  & V, T                         & 3102                  & 930s                        & 3102           & \clickablecheckmark{https://github.com/JUNJIE99/MLVU}                  \\
\textbf{Video-MME}~\citeyearpar{fu2025videomme}        & SC, BK, etc.               & YouTube    & manual                    & \ding{51}                  & V, A, T                  & 900                   & 1017.9s                     & 2700           & \clickablecheckmark{https://video-mme.github.io}                  \\
\textbf{MotionBench}~\citeyearpar{hong2025motionbench}      & BB, etc.                   & existing, syn., web  & manual        & \ding{51}                  & V, T                         & 5385                  & \textless{}10s              & 8052           & \clickablecheckmark{https://huggingface.co/datasets/zai-org/MotionBench}                  \\
\textbf{OVO-Bench}~\citeyearpar{niu2025ovobench}        & -                          & existing, YouTube    & manual             & \ding{51}                  & V, T                         & 644                   & 428.89s                     & 2814           &      \clickablecheckmark{https://huggingface.co/datasets/JoeLeelyf/OVO-Bench}                   \\
\textbf{VISTA-400K}~\citeyearpar{ren2025vista_HRVideoBench}       & -                          & existing      & manual                    & \ding{55}                  & V, T                         & 403994                & 48.6s                       & $\sim$381K     &            \clickablecheckmark{https://huggingface.co/datasets/TIGER-Lab/VISTA-400K}             \\
\textbf{HRVideoBench}~\citeyearpar{ren2025vista_HRVideoBench}     & -                          & online    & manual                     & \ding{51}                  & V, T                         & 200                   & 5.4s                        & 200            &          \clickablecheckmark{https://huggingface.co/datasets/TIGER-Lab/HRVideoBench}               \\
\textbf{HarmonySet-train}~\citeyearpar{zhou2025harmonyset} & \multirow{2}{*}{-}         & \multirow{2}{*}{YouTube} & \multirow{2}{*}{manual}     & \ding{55}                  & \multirow{2}{*}{V, A, T} & 44470                 & \multirow{2}{*}{31.5s}      & 44470          & \multirow{2}{*}{\clickablecheckmark{https://huggingface.co/datasets/Zzitang/HarmonySet/tree/main}}       \\
\textbf{HarmonySet-MC}~\citeyearpar{zhou2025harmonyset}    &         &                   &                              & \ding{51}                  &                                     & 3858                  &                             & 3858           &                         \\
\textbf{VideoA11y-40K}~\citeyearpar{li2025videoa11y}    & -                          & online        & auto               & \ding{55}                  & V, A                        & 40000                 & -                           & N/A            &     \clickablecheckmark{https://huggingface.co/datasets/chaoyuli/VideoA11y-40K}                    \\
\textbf{TUNA}~\citeyearpar{kong2025tuna}             & SC, BK, etc.               & existing      & manual                  & \ding{51}                  & V, T                         & 1000                  & 14.5s                       & 2000           &         \clickablecheckmark{https://huggingface.co/datasets/friedrichor/TUNA-Bench}                \\

\textbf{V-STaR}~\citeyearpar{cheng2025vstar}           & -                          & existing, YouTube    & manual           & \ding{51}                  & V, T                         & 2094                  & 110.23s                     & -              &          \clickablecheckmark{https://huggingface.co/datasets/V-STaR-Bench/V-STaR}               \\
\textbf{MINERVA}~\citeyearpar{nagrani2025minerva}          & BK, TN, etc.               & YouTube     & manual                 & \ding{51}                  & V, T                         & 223                   & 12m                         & 1515           &         \clickablecheckmark{https://github.com/google-deepmind/neptune}                \\
\textbf{MAVERIX}~\citeyearpar{xie2025maverix}          & SC, BK, etc.               & existing     & manual                   & \ding{51}                  & V, A, T                  & 700                   & 5.7m                        & 2556           & \ding{55}                       \\
\textbf{RTV-Bench}~\citeyearpar{xun2025rtvbench}        & SC, BK, etc.               & existing, online      & manual          & \ding{51}                  & V, T                         & 552                   & 18.2m                       & 4631           & \clickablecheckmark{https://github.com/LJungang/RTV-Bench}                  \\
\textbf{VidText}~\citeyearpar{yang2025vidtext}          & SC, BK, BM, TT, SW         & existing, YouTube    & manual           & \ding{51}                  & V, A, T                  & 939                   & 108.2s                      & 2857           &          \clickablecheckmark{https://huggingface.co/datasets/sy1998/VidText}               \\
\textbf{SIV-Bench}~\citeyearpar{kong2025sivbench}        & SC, etc.                   & YouTube, TikTok   & manual           & \ding{51}                  & V, A, T                  & 2792                  & 32.49s                      & 8728           &     \clickablecheckmark{https://huggingface.co/datasets/Fancylalala/SIV-Bench}                    \\
\textbf{ExAct}~\citeyearpar{yi2025exact}            & SC, BK, CL                 & existing    & expert                    & \ding{51}                  & V, T                         & 3521                  & 105s                        & 3521           &          \clickablecheckmark{https://huggingface.co/datasets/Alexhimself/ExAct}               \\
\textbf{VRBench}~\citeyearpar{yu2025vrbench}          & SC, BK, VB, etc.           & YouTube    & manual                  & \ding{51}                  & V, A, T                  & 960                   & 1.6h                        & 8243           &            \clickablecheckmark{https://huggingface.co/datasets/OpenGVLab/VRBench}             \\
\textbf{EgoExoBench}~\citeyearpar{he2025egoexobench}      & BK, etc.                   & existing    & manual                    & \ding{51}                  & V, T                         & -                     & -                           & 7350           &    \clickablecheckmark{https://github.com/ayiyayi/EgoExoBench}                     \\

\textbf{WildVideo}~\citeyearpar{yang2025wildvideo}        & -                          & existing   & manual                     & \ding{51}                  & V, T                         & 1318                  & $\sim$30s                   & 17625          & \ding{55}                  \\   
\textbf{WorldSense}~\citeyearpar{hong2025worldsense}       & -                          & existing    & manual                    & \ding{51}                  & V, A, T                  & 1662                  & 141.1s                      & 3172           &         \clickablecheckmark{https://huggingface.co/datasets/honglyhly/WorldSense}                \\
\textbf{Trust-videoLLMs}~\citeyearpar{wang2025Trust_videoLLMs}  & -                          & existing, syn., YouTube & manual   & \ding{51}                  & V, A, T                  & 6955                  & -                           & -              &    \clickablecheckmark{https://github.com/wangyouze/Trust-videoLLMs}                     \\
\textbf{CausalStep}~\citeyearpar{li2025causalstep}       & -                          & existing    & manual                    & \ding{51}                  & V, T                         & 100                   & 430.5s                      & 1852           & \ding{55}                       \\
\bottomrule
\end{tabular}

\caption{Summary of large-model-related datasets specifically for sports understanding and general video understanding with sports content. Sports: SC: soccer, BK: basketball, BM: badminton, TN: tennis, GY: gymnastics, FS: figure skating, AF: American football, BB: baseball, CR: cricket, TT: table tennis, VB: volleyball, DV: diving, IH: ice hockey, CL: sports climbing, SW: swimming. Source: syn.: synthesis. Modal: V: video, A: audio, T: text.}
\vspace{-0.5cm}
\label{tab:appendix_su_table}
\end{table*}

\setlength{\tabcolsep}{10pt}
\begin{table*}[ht]
\centering
\fontsize{12pt}{15pt}\selectfont
\begin{tabular}{ll@{\hspace{20pt}}ll}
\toprule
\textbf{Abbr.} & \textbf{Sports} & \textbf{Abbr.} & \textbf{Tasks} \\
\midrule
\textbf{AF} & American Football & \textbf{ACT} & Action Spotting and Recognition \\
\textbf{BB} & Baseball & \textbf{AQA} & Sports Action Quality Assessment \\
\textbf{BK} & Basketball & \textbf{CMT} & Sports Commentary Generation \\
\textbf{BM} & Badminton & \textbf{EDU} & Sports Education \\
\textbf{BX} & Boxing & \textbf{HLG} & Sports Highlight Generation \\
\textbf{CL} & Sports Climbing & \textbf{INJ} & Sports Injury and Rehabilitation \\
\textbf{CR} & Cricket & \textbf{MAN} & Sports Management \\
\textbf{CY} & Cycling & \textbf{MOD} & Sports Models and Systems \\
\textbf{DV} & Diving & \textbf{NAR} & Sports Narratives and Storytelling \\
\textbf{FS} & Figure Skating & \textbf{NSG} & Sports News Generation \\
\textbf{GY} & Gymnastics & \textbf{OPI} & Public Opinion Analysis in Sports \\
\textbf{HB} & Handball & \textbf{PLA} & Exercise and Training Plans \\
\textbf{IH} & Ice Hockey & \textbf{PRD} & Game and Player Performance Prediction \\
\textbf{RG} & Rugby & \textbf{PSY} & Sports Psychology and Behavior \\
\textbf{SC} & Soccer & \textbf{REF} & Sports Refereeing \\
\textbf{TF} & Track and Field & \textbf{TAC} & Sports Tactics and Strategies \\
\textbf{TN} & Tennis & \textbf{TOU} & Sports Tourism \\
\textbf{TT} & Table Tennis & \textbf{TSC} & Sports Talent Scouting \\
\textbf{VB} & Volleyball & \textbf{WRI} & Sports Academic Writing \\
\bottomrule
\end{tabular}
\caption{Abbreviations for sports and tasks mentioned in this paper (sorted alphabetically by abbreviation).}
\label{abbr_tab}
\end{table*}

\end{document}